\documentclass[accepted]{uai2023} 

\usepackage[american]{babel}
\usepackage{xr}

\usepackage{natbib} 
\usepackage{multicol}
\usepackage[utf8]{inputenc} 
\usepackage[T1]{fontenc}    
\usepackage{hyperref}       
\usepackage{url}            
\usepackage{booktabs}       
\usepackage{amsfonts}       
\usepackage{nicefrac}       
\usepackage{microtype}      
\usepackage{xcolor}   
\usepackage{amsmath}
\usepackage{amssymb}
\usepackage{mathtools}
\usepackage{amsthm}
\usepackage{algorithm}
\usepackage{algorithmic}
\usepackage{natbib}
\usepackage{balance}
\newcommand{\twopartdef}[4]
{
	\left\{
		\begin{array}{ll}
			#1 & \mbox{if } #2 \\
			#3 & \mbox{if } #4
		\end{array}
	\right.
}
\usepackage{subcaption}
\usepackage{multirow}
\usepackage{graphicx}
\usepackage{microtype}

\newcommand\blfootnote[1]{%
  \begingroup
  \renewcommand\thefootnote{}\footnote{#1}%
  \addtocounter{footnote}{-1}%
  \endgroup
}

\theoremstyle{plain}
\newtheorem{theorem}{Theorem}[section]
\newtheorem{proposition}[theorem]{Proposition}
\newtheorem{lemma}[theorem]{Lemma}
\newtheorem{corollary}[theorem]{Corollary}
\theoremstyle{definition}
\newtheorem{definition}[theorem]{Definition}
\newtheorem{assumption}[theorem]{Assumption}
\theoremstyle{remark}
\newtheorem{remark}[theorem]{Remark}
\newtheorem{notation}[theorem]{Notation}
\allowdisplaybreaks

\title{Online Heavy-tailed Change-point detection}
\date{}

\author[1]{{Abishek Sankararaman\footnote{here}}}
{\author[1]{{Balakrishnan (Murali) Narayanaswamy}}}
\affil[1]{%
AWS AI Labs
}

\makeatletter
\newcommand*{\addFileDependency}[1]{
\typeout{(#1)}
\@addtofilelist{#1}
\IfFileExists{#1}{}{\typeout{No file #1.}}
}\makeatother

\newcommand*{\myexternaldocument}[1]{%
\externaldocument{#1}%
\addFileDependency{#1.tex}%
\addFileDependency{#1.aux}%
}
\myexternaldocument{sankararaman_560-supp}

\begin{document}

\maketitle

\begin{abstract}
       We study algorithms for online  change-point detection (OCPD), where samples that are potentially heavy-tailed, are presented one at a time and a change in the underlying mean must be detected as early as possible. We present an algorithm based on clipped Stochastic Gradient Descent (SGD), that works even if we only assume that the second moment of the data generating process is bounded. We derive guarantees on worst-case, finite-sample false-positive rate (FPR) over the family of all distributions with bounded second moment. Thus, our method is the first OCPD algorithm that guarantees finite-sample FPR, even if the data is high dimensional and  the underlying distributions are heavy-tailed. The technical contribution of our paper is to show that clipped-SGD can estimate the mean of a random vector and simultaneously provide confidence bounds at all confidence values. We combine this robust estimate with a union bound argument and construct a sequential change-point algorithm with finite-sample FPR guarantees. We show empirically that our algorithm works well in a variety of situations, whether the underlying data are heavy-tailed, light-tailed, high dimensional or discrete. No other algorithm achieves bounded FPR theoretically or empirically, over all settings we study simultaneously. 
\end{abstract}
\blfootnote{* Correspondence to Abishek Sankararaman : abisanka@amazon.com}
{\color{black}
\section{Introduction}

Online change-point detection (OCPD) is a fundamental problem in statistics where instantiations of a random variable are presented one after another and we want to detect if some parameter or statistic corresponding to the underlying data generating distribution has changed. This problem has been widely studied in machine learning, mathematical statistics  and information theory over the past century. In part, this is due to the wide-ranging applications of OCPD to computational biology \citep{muggeo2011efficient}, online advertising \citep{zhang2017online}, cyber-security \citep{osanaiye2016change, kurt2018bayesian, polunchenko2012nearly}, cloud-computing \citep{maghakian2019online}, finance \citep{lavielle2007adaptive}, medical diagnostics \citep{yang2006adaptive, gao2018automatic} and robotics \citep{konidaris2010constructing}. We refer interested readers to the recent surveys of \citep{aminikhanghahi2017survey} and \citep{xie2021sequential} for details of applications of OCPD. These surveys build upon the classical texts in change-point detection obtained over the last decade \citep{basseville1993detection,tartakovsky1991sequential,krichevsky1981performance}.

Classical results for OCPD have focused on algorithms that assume known distributions for either one or both of the pre- and post-change data \citep{wald1992sequential,page1954continuous,shiryaev2007optimal,lorden1971procedures,pollak1985optimal,ritov1990decision,moustakides1986optimal,tartakovsky1991sequential}. In recent years, algorithms have been developed for cases when the pre- and post- change distributions are unknown, but belong to a parametric class such as the exponential family \citep{lai2010sequential,fryzlewicz2014wild,frick2014multiscale,cho2016change}. Non-parametric algorithms have been developed in \citep{padilla2021optimal,madrid2021optimal} and the references therein, but they only give asymptotic guarantees. The algorithms of \citep{adams2007bayesian,lai2010sequential,maillard2019sequential,alami2020restarted} have finite-sample guarantees, but either rely on parametric assumptions such as an exponential family, or on tail assumptions such as sub-gaussian distribution families. The works of \citep{bhatt2022offline} and \citep{li2021adversarially}  build upon the work in \citep{niu2012screening}, and give algorithms for multiple change-points with possibly heavy-tailed data in the \emph{offline} case with all data available up-front. The works of \citep{wang2022catoni, wang2023huber, shekhar2023sequential} give OCPD algorithms for heavy-tailed, but uni-variate data.


In many modern applications such as cloud-computing and monitoring, data is known to often be heavy-tailed \citep{nair2022fundamentals,loiseau2010investigating,nizam2016attack} and too complex to model with any simple parametric family \citep{barnett2016change,hallac2015network,dartmann2019big}. Given the velocity, variety and volume of modern data streams, performance of change-point detection is measured through false-positive rates in order to combat alert fatigue \citep{ruff2021unifying}, and algorithms must work for streams that have multiple change points. Motivated by these requirements, we seek an OCPD algorithm that simultaneously meet the following desiderata : it  {\em(i)} detects multiple change-points, {\em(ii)} makes no parametric assumptions on the distribution of data, {\em(iii)} works with potentially heavy-tailed data, {\em (iv)} works for high-dimensional data streams, and {\em(v)}  guarantees finite sample FPR.  

}

{\color{black}

\subsection{Main Contributions}

Our paper is the first to give an online algorithm satisfying all the $5$ desiderata listed above. Specifically, our algorithm gives finite sample guarantees for FPR and detection-delay without assuming that data comes from a specific parametric family or assuming strong tail conditions - such as that the data have sub-gaussian distributions. No previous algorithm for OCPD simultaneously achieves all desiderata. 
Our main technical contribution is to  provide a {\ttfamily clipped-SGD} algorithm with finite sample confidence bounds for heavy-tailed mean estimation, \emph{that hold for all confidence values simultaneously}, a result  of independent interest. We use these bounds to build a OCPD algorithm with finite sample FPR.

We further show good empirical performance across a variety of data streams with heavy-tailed, light-tailed, high dimensional or discrete distributions.
However while our algorithm is designed to work across different distributions, we observe theoretically and empirically that when data has additional structure such as being one-dimensional with sub-gaussian tails or is binary, then specialized OCPD algorithms for those cases yield better results than our method. Closing these gaps is an ongoing direction of research.




}

{\color{red}

}
\section{Problem Setup}
\label{sec:problem_formulation}

At each time $t$, a random vector $X_t \in \mathbb{R}^d$ is revealed to an OCPD algorithm. $X_t$ has a probability measure and expectation denoted by $\mathbb{P}_t$ and $\mathbb{E}_t$ respectively, and mean $\mathbb{E}_{t}[X_t] \in \mathbb{R}^d$. Subsequently, using all the samples observed so far - $X_1, \cdots, X_t$ - the algorithm outputs a binary decision denoting whether a change in mean has occurred  since time $t=1$ or the last time a change was output by the algorithm, whichever is larger. The goal of the OCPD algorithm is identify the change points as quickly as possible after they occur, with bounded false-positive rate (FPR). 
The observed datum $(X_t)_{t \geq 1}$ are independent, although not identically distributed with piece-wise constant mean. 

\begin{definition}[Piece-wise constant mean process]
Let $T$ be the time horizon (stream-length) and let $Q_T < T$ be the total number of change-points. A set of strictly increasing time-points $1 < \tau_1 < \tau_2 \cdots < \tau_{Q_T +1} := T+1$ are called change-points, if for all $c \in \{1,\cdots, Q_T\}$ 
\begin{itemize}
\item $\forall t \in [1,T]$, $X_t \sim \mathbb{P}_t$ independently.
    \item $\forall t \in [\tau_c, \tau_{c+1})$, the mean $\mathbb{E}_t[X_t] := \theta_c$ of the observation is constant and does not depend on $t$.
    \item $\forall c \in [1,Q_T]$, $\theta_{c} \neq \theta_{c+1}$.
\end{itemize}
\end{definition}
Thus, a piece-wise constant mean process is identified by the quadruple $\mathfrak{M} := (T, Q_T, (\tau_c)_{c=1}^{Q_T}, (\mathbb{P}_t)_{t=1}^T)$. Throughout, we use probability and expectation operators $\mathbb{P}$ and $\mathbb{E}$, to denote the joint product probability distribution $(\mathbb{P}_t)_{t=1}^T$. 

\subsection{Assumptions}
Let $\mathcal{P}$ be a family of probability measures  on $\mathbb{R}^d$ such that the probability distributions $\mathbb{P}_t$, for all $t$, are from this family, i.e., $\mathbb{P}_t \in \mathcal{P}$, $\forall t \in [1,T]$. Throughout this paper, we make the following non-parametric assumptions on the family $\mathcal{P}$. 

\begin{assumption}
There exists a convex compact set $\Theta \subset \mathbb{R}^d$ known to the algorithm, such that for all $ \mathbb{P} \in \mathcal{P}$, $\mathbb{E}_{X \in \mathbb{P}}[X] \in \Theta$. In words, the mean of all the distributions in the family belong to a known bounded set $\Theta$ such that $\max_{\theta_1, \theta_2 \in \Theta}\| \theta_1 - \theta_2\| := G$.
\label{assumption:bounded}
\end{assumption}

\begin{assumption}
     There exists $\sigma > 0$ {known} to the algorithm, such that for all $\mathbb{P} \in \mathcal{P}$ and $\theta \in \Theta$, $\mathbb{E}_{X \sim \mathbb{P}}[ \| X - \mathbb{E}_{X \sim \mathbb{P}}[X] \|_2^2] \leq \sigma^2$. In words, the second moment is uniformly bounded for all distributions in $\mathcal{P}$.
     \label{assumption_stochastic}
\end{assumption}

These assumptions are very general and encompass a wide range of families such as any bounded distribution, the set of sub-Gaussian distributions and heavy-tailed distributions that do not have finite higher moments. We seek algorithms that work without knowing the length of the data stream, the number of change-points and that do not make any assumptions on the underlying distributions generating the samples, beyond Assumptions  \ref{assumption:bounded} and \ref{assumption_stochastic}.

\subsection{Performance Measures}

Any OCPD algorithm is measured by two performance metrics -- {\em (i)} False-positive rate and {\em(ii)} Detection delay. We set notation to define these measures.

\begin{notation}
For every $1 \leq r \leq s < T$, we denote by $X_{r:s} := (X_r, X_{r+1}, \cdots, X_s)$ to be the set of observed vectors from time $r$ to time $s$, with both end-points $r$ and $s$ inclusive. 
\end{notation}

\begin{definition}[OCPD algorithm]
A sequence of measurable functions $\mathcal{A} := (\mathcal{A}_t)_{t\geq 1}$ is called an OCPD algorithm if for every time $t \geq 1$, $\mathcal{A}_t \in \{0,1\}$ and is measurable with respect to the sigma algebra generated by $X_{1:t}$. The interpretation is that if $\mathcal{A}_t = 1$ for some $t$, then the algorithm has detected a change at time $t$ and if $\mathcal{A}_t = 0$, no change is detected at time $t$. 
\end{definition}

\begin{notation}
For an OCPD algorithm $\mathcal{A}$ and for all $t \in [T]$, denote by $R^{(\mathcal{A})}(t) \in \mathbb{N}$ to be the random variable denoting the number of detections made till time $t$, i.e., $R^{(\mathcal{A})}(t) = \sum_{s=1}^t \mathcal{A}_s$.
\end{notation}

\begin{notation}
For an OCPD algorithm $\mathcal{A}$, and every $r \in \mathbb{N}$ and, denote by $t_{r}^{(\mathcal{A})}$ as the stopping time
\begin{align*}
    t_{r}^{(\mathcal{A})} := \min( \inf \{t \in [0,T] \text{ s.t. } R^{(\mathcal{A})}(t) \geq r \}, T+1),
\end{align*}
where the $\inf$ of an empty set is defined to be $\infty$. In words, $t_{r}^{(\mathcal{A})}$ is the stopping time when the OCPD algorithm detects a change for the $r$th time, or $T+1$, whichever is larger. 
\end{notation}

\begin{definition}[False Positive Detection]
The $r$th detection of an OCPD algorithm $\mathcal{A}$ is said to be a False Positive, if there exists no change-point between the $r-1$th and the $r$th detection. Formally, denote by the indicator (random) variable $\chi_r^{(\mathcal{A})} = \mathbf{1}(\not\exists c \in [1,Q_T] \text { s.t. } \tau_c \in (t_{r-1}^{(\mathcal{A})},t_{r}^{(\mathcal{A})}])$ to denote if the $r$th detection of $\mathcal{A}$ is a false-positive. Note that by definition, on the event that $R^{(\mathcal{A})}(T) < r$, $\chi_r^{(\mathcal{A})} = 0$. 
\end{definition}

\begin{definition}[False Positive Rate (FPR)]
An OCPD algorithm $\mathcal{A}$ is said to have false-positive rate bounded by $\delta \in (0,1)$ if 
\begin{align}
   \sup_{\mathfrak{M}} \mathbb{E}\left[ \frac{\sum_{r=1}^{T} \chi_r^{(\mathcal{A})}}{R^{(\mathcal{A})}(T)} \mathbf{1}(R^{(\mathcal{A})}(T)>0)\right] \leq \delta.
   \label{eqn:fpr_definition}
\end{align}

In words, an OCPD algorithm $\mathcal{A}$ has bounded false positive rate, if for every piece-wise constant mean process $\mathfrak{M}$, the expected fraction of false-positives made by the algorithm $\mathcal{A}$ is bounded by $\delta$. In Equation (\ref{eqn:fpr_definition}), we take the sum till $T$ because that is the maximum number of possible change points detected. If an algorithm only detects $s < T$ change points, then by definition $\chi_r^{(\mathcal{A})} = 0$ for all $r > s$.
\end{definition}

\begin{definition}[Worst-case Detection Delay]
For $n \in \mathbb{N}$ and $\Delta > 0$, let $X_1, X_2, \cdots, X_n, X_{n+1}, \cdots$ be an infinite stream with the following distribution. For every $t < n$, $X_t \stackrel{\text{ind}}{\sim} \mathbb{P}_t$ with $\mathbb{E}_{X \sim \mathbb{P}_t}[X] = \theta_1 \in \Theta$ and for every $t \geq n$, $X_t \stackrel{\text{ind}}{\sim} \mathbb{P}_t$ with $\mathbb{E}_{X \sim \mathbb{P}_t}[X] = \theta_2 \in \Theta$ with $\| \theta_1 - \theta_2 \| = \Delta$. Let $\mathfrak{M}^{(n,\Delta)}$ denote all such infinite piece-wise constant mean process. An algorithm $\mathcal{A}$ is said to have worst-case detection delay $\mathcal{D}(\Delta, n , \delta')$, if 
\begin{multline}
    \sup_{\mathfrak{M}^{(n,\Delta)}}\mathbb{P}\bigg[ \inf \{ t > n \text{: } \mathcal{A}_t = 1\} - n \geq  \mathcal{D}(\Delta, n, \delta')\bigg]  \leq \delta'
    \label{eqn:detection_delay_defn}
\end{multline}
 holds for all $n \in \mathbb{N}$, $\Delta > 0$ and $\delta' \in (0,1)$.
\end{definition}

In words, the detection delay function $\mathcal{D}(\Delta, n ,\delta')$ is such that  for every admissible process $\mathfrak{M}^{(n,\Delta)}$ that has a single change-point at time $n$ with jump magnitude $\Delta$, algorithm  $\mathcal{A}$ detects the change-point before time $n+ \mathcal{D}(\Delta, n ,\delta' )$, with probability at-least $1-\delta'$. Note that the delay metric is measured on data streams with exactly one change-point. Defining detection delay for streams with multiple change-points is ambiguous as there could be missed detections, with only a subset of the change-points being detected \citep{alami2020restarted}, \citep{maillard2019sequential}. The main question this paper studies is

\begin{center}
\textit{For each $\delta \in (0,1)$, does there exists an OCPD algorithm with FPR bounded by $\delta$ and having small worst-case detection-delay that only makes Assumptions \ref{assumption:bounded} and \ref{assumption_stochastic}  ? }.
\end{center}

Observe that it is trivial to achieve a FPR of $0$ for example the constant function where $\mathcal{A}(\cdot) = 0$, i.e., an algorithm that never detects change-point at all. However, this algorithm has a worst-case detection-delay of $\infty$, i.e., $\mathcal{D}(\Delta, n, \delta') = +\infty$ for all $\Delta > 0$, $n \in \mathbb{N}$ and $\delta' \in (0,1)$. Thus, the challenge is to design an algorithm that satisfies the FPR constraint of $\delta$ while having small, finite worst-case detection delay, without making parametric assumptions on the underlying data generating distributions.

\section{Online Robust Mean Estimation}
\label{sec:mean_estimation}

The central workhorse of our change-point detection algorithm is heavy-tailed online mean estimation. Suppose $X_1, X_2, \cdots$ are a sequence of independent random vectors, with the means $\mathbb{E}_t[X_t] = \theta^* \in \Theta$ being a constant independent of time $t$. Let $(\widehat{\theta}_t)_{t \geq 1}$ be a sequence of random variables such that $\widehat{\theta}_t$ is an estimate of $\theta$ based on the samples $X_1, \cdots, X_{t}$ defined through clipped-SGD algorithm described as follows. For a given non-negative sequence $(\eta_t)_{t \geq 1}$ and $\lambda > 0$, the estimate $\widehat{\theta}_0 \in \Theta$ is arbitrary,  $\widehat{\theta}_t$ for each $t \geq 1$, is given by
\begin{align}
    \widehat{\theta}_t \coloneqq \prod_{\Theta}(\widehat{\theta}_{t-1} - \eta_t \text{clip}(X_t - \widehat{\theta}_{t-1}, \lambda)),
    \label{eqn:sgd_update}
\end{align}
where, $\prod_{\Theta}$ is the projection operator onto the convex compact set $\Theta$ and for every $x \in \mathbb{R}^d$ and $\lambda > 0$,
\begin{align}
    \text{clip}(x,\lambda) = x \min \left( 1, \frac{\lambda}{\|x\|} \right).
\end{align}

Our main result on the convergence of the estimator $\widehat{\theta}_t$ to the true $\theta^*$ with increasing number of samples $t$ is the following. 

\begin{theorem}
For all times $t \geq 1$, when clipped SGD in Equation (\ref{eqn:sgd_update}) is run with $\lambda = 2G$ and $\eta_t = \frac{2}{(t+\gamma)}$ for $\gamma = \max \left(120 \lambda \sigma(\sigma+1),{320\sigma^2}+1 \right)$, then for every $t \geq 1$ and every $\delta \in (0,1)$, 
 \begin{align*}
   \mathbb{P} \left[ \| \widehat{\theta}_t - \theta^* \|_2^2 \geq {\mathcal{B}(t, \delta)} \right] \leq \frac{\delta}{t(t+1)},
 \end{align*}
where 
\begin{multline}
    \mathcal{B}(t, \delta) := C_t\bigg[ \frac{\gamma^2 G^2}{(t+1)^2} +\left(\frac{16  \sigma^2}{\lambda} +  4  \sigma^2 \right) \frac{1}{2(t+1)} \\+ \frac{96 \lambda^2 \ln \left( \frac{2t^2(t+1)}{\delta}\right)\sigma(\sigma+1) }{ (t+\gamma)\sqrt{t+1}}  \bigg],
    \label{eqn:defn_B}
\end{multline}
and $C_t  = \max(\frac{1024\sigma^4}{G^2\lambda^2}, \frac{8 \lambda \sqrt{\ln \left( \frac{2t^2(t+1)}{\delta} \right)}}{\gamma^2 G} )$.
\label{thm:main_mean_est}
\end{theorem}
{\color{black}
\begin{corollary}
There exists an universal constant $A > 0$ such that for all $t \geq 1$, when clipped SGD in Equation (\ref{eqn:sgd_update}) is run with parameters in Theorem \ref{thm:main_mean_est}
\begin{align*}
    \mathbb{P} \left[ \| \widehat{\theta}_t - \theta^* \| \geq A \max \left(\frac{\sigma^3}{\sqrt{t}} , \frac{\sigma \sqrt{\ln \left( \frac{t^3}{\delta}\right)}}{\sqrt{t}}\right) \right] \leq \frac{\delta}{t(t+1)},
\end{align*}
holds for every  $\delta \in (0,1)$.
\end{corollary}
}

Proof is in Appendix in Section \ref{sec:mean_estimation_proofs} and uses tools from \citep{bubeck2015convex}, \citep{gorbunov2020stochastic, tsai2022heavy} and \citep{victor1999general}. 

\begin{remark}
Compared to \citep{tsai2022heavy}, we do not need the failure probability $\delta$ in the input and we can give simultaneous confidence intervals for all failure probabilities $\delta$. In contrast, the algorithm of \citep{tsai2022heavy} requires $\delta \in (0,1)$ as an input and only guarantees that the estimate mean is close to the true mean, upto error probability of $\delta$. However, the bound in Theorem \ref{thm:main_mean_est} is off by logarithmic factors compared to \citep{tsai2022heavy}. Concretely, $C_t = O(1)$ for the algorithm of \citep{tsai2022heavy}, while it is $O( \log(t/\delta))$ for us. This is the price to have confidence intervals hold for all failure probabilities simultaneously as opposed to just having one single failure probability.
\end{remark}

\begin{remark}
Compared to the setting of \citep{tsai2022heavy}, our setting is \emph{weaker} as we assume that the domain $\Theta$ is compact with finite diameter $G$. This is what enables us to use an appropriately tuned learning rate and clipping parameter to make the algorithm any-time and obtain confidence intervals at all failure probabilities simultaneously. It is an open question whether the assumption that $\Theta$ is compact can be relaxed and if we can still make guarantees confidence interval holding for all failure probabilities $\delta$ for all $t$ for heavy tailed distributions.
\end{remark}

\begin{remark}
The constants in Theorem \ref{thm:main_mean_est} are not optimal. In Section \ref{sec:experiments}, we suggest an alternative set of constants that work well empirically across variety of settings.
\label{remark:sub_opt_constants}
\end{remark}

\begin{remark}
There have been significant recent advances in robust mean estimation \citep{diakonikolas2020outlier,lugosi2021robust,depersin2022robust,cherapanamjeri2020algorithms,diakonikolas2022streaming}, that are known to provide near optimal error bounds. However, unlike our method, none of these algorithms can give confidence bounds for all confidence values simultaneously.
\end{remark}

\begin{remark}
Theorem $4.3$ in \citep{devroye2016sub} proves that it is impossible to get a finite sample confidence bound to hold for all $\delta \in (0,1)$. Our result does not contradict this since the restriction on the allowable $\delta$ is \emph{implicit} in Theorem \ref{thm:main_mean_est}. Equation (\ref{eqn:defn_B}) gives that, for every $t \in \mathbb{N}$, as $\delta \searrow 0$, $\mathcal{B}(t, \delta) \nearrow \infty$. However, from Assumption \ref{assumption:bounded},  if $\mathcal{B}(t,\delta) \geq G$,  then the statement of Theorem \ref{thm:main_mean_est} is vacuous. Thus, Theorem \ref{thm:main_mean_est} gives a non-vacuous bounds only for $\delta \in (\delta_{min}^{(t)},1)$ where $\delta_{min}^{(t)} := \inf_{\delta > 0}\{ \mathcal{B}(t,\delta) < G \}$.
\end{remark}

{\color{black}
\subsection{Uniform over time bound}
{\color{black}
As a corollary of Theorem \ref{thm:main_mean_est}, we get the following bound that holds uniformly over all time. 
\begin{corollary}
There exists an universal constant $A > 0$ such that, when clipped SGD in Equation (\ref{eqn:sgd_update}) is run with parameters in Theorem \ref{thm:main_mean_est},
\begin{align*}
\mathbb{P}\left[ \exists t \in \mathbb{N} :  \| \widehat{\theta}_t - \theta^* \| \geq A \max \left(\frac{\sigma^3}{\sqrt{t}} , \frac{\sigma \sqrt{\ln \left( \frac{t^3}{\delta}\right)}}{\sqrt{t}}\right) \right] \leq \delta,
\end{align*}
holds for every  $\delta \in (0,1)$.
\label{cor:dim_free}
\end{corollary}
}

The proof follows by taking an union bound over all $t \geq 1$, i.e., summing over $t \geq 1$ on both the LHS and RHS of Corollary \ref{cor:dim_free} and noticing that $\sum_{t \geq 1}\frac{1}{t(t+1)}=1$. The bounds in Theorem \ref{thm:main_mean_est} and Corollary \ref{cor:dim_free} are \emph{dimension free}, i.e., the term $d$ does not appear in the bounds. The moment bound $\sigma$ plays the role of dimension. In particular, suppose that all distributions in the family $\mathcal{P}$ have covariance matrices bounded in the positive semi-definite sense by $\Sigma \in \mathbb{R}^{d \times d}$. In this case, by definition, $\sigma^2 \leq \text{Trace}(\Sigma)$ and  plays the role of dimension.

In the special case when the samples $(X_t)_{t\geq 1}$ are i.i.d. with sub-gaussian distributions with mean $\theta^*$ and covariance matrix $\Sigma$, \citep{NIPS2011_e1d5be1c,maillard2019sequential, chowdhury2022bregman} show that for all $\delta \in (0,1)$,
\begin{multline}
    \mathbb{P}\bigg[ \exists t \in \mathbb{N} : \bigg\|\frac{1}{t}\sum_{s=1}^tX_s - \theta^* \bigg\| \geq \\ \sqrt{2 \lambda_{max}(\Sigma) \left( 1+ \frac{1}{t}\right) \ln \left( \frac{(t+1)^{{\color{black}d}}}{\delta} \right)} \bigg] \leq \delta,
    \label{eqn:maillard_them}
\end{multline}
holds, where $\lambda_{max}(\Sigma)$ is the highest eigen-value of the co-variance matrix $\Sigma$. Thus, for the special case of sub-gaussian distributions, Equation (\ref{eqn:maillard_them}) has a better dependence on time $t$ compared to our Corollary \ref{cor:dim_free}. The improved dependence on time arises as Equation (\ref{eqn:maillard_them}) is based on the construction of a self normalized martingale and using the martingale stopping theorem to obtain uniform over time bounds while Corollary \ref{cor:dim_free} is based on a simple union bound.

However, Equation (\ref{eqn:maillard_them}) is not dimension free and depends on the scale of the problem through the term $d \lambda_{max}(\Sigma)$ which by definition is larger than $\text{Trace}(\Sigma)$. In many high dimensional settings,  $d \lambda_{max}(\Sigma)$ is much {larger} than $\text{Trace}(\Sigma)$ and thus algorithms and bounds depending explicitly on $d$ is un-desirable \citep{wainwright2019high, lugosi2019mean}. For the {uni-variate} heavy-tailed distributions, a sequence of works \citep{wang2022catoni, wang2023huber} establish  confidence bounds with sharp dependence on time by extending the martingale recipe developed in \citep{howard2021time}. In our draft, we are able to get dimension free bounds for heavy-tailed distributions, but at the cost of a compactness Assumption \ref{assumption:bounded} that are not needed in \citep{NIPS2011_e1d5be1c}. It is an open question if we can get dimension-free bounds with the improved time-dependence of the kind in Equation (\ref{eqn:maillard_them}) without the compactness assumption.

}

\section{Change-Point Detection Algorithm}

{\color{black}Our algorithm is described in Algorithm \ref{algo:learn_model} and is based on the following idea. A change point is detected in the time-interval $[r,t]$ if there exists $r < s < t$ such that confidence interval around the estimated mean of the observations $X_{r:s}$  is separated from the confidence interval around the estimated mean of the observations $X_{s+1:t}$. Further, in order to accommodate multiple change-points, the algorithm \emph{restarts} after every change detection, similar to \citep{alami2020restarted}. 
It is known that standard empirical mean is a poor estimator when the underlying distributions can potentially be heavy-tailed, as its confidence interval under only assumptions in \ref{assumption_stochastic} is wide \citep{lugosi2019mean}. 
To attain better confidence intervals, we use  the {\ttfamily clipped-SGD} Algorithm \ref{algo:learn_model} that gives a confidence interval for the estimated mean for every failure probability $\delta\in (0,1)$ simultaneously. Having multiple confidence intervals is crucial as we show that adaptively testing different intervals of times at different carefully chosen confidence intervals (Line $8$ of Algorithm \ref{algo:learn_model}) leads to the bounded FPR guarantee.

}




	\begin{algorithm*}[htb]
		\caption{{ Online {\ttfamily Clipped-SGD} Change Point Detection}}
		\label{algo:learn_model}

		\begin{algorithmic}[1]
				\STATE \textbf{Input}: {  $(\eta_t)_{t \geq 1}$,  $\lambda > 0$, $\theta_0 \in \Theta$, $\delta \in (0,1)$ the FPR guarantee } 
				\STATE $r \gets 1$
				\STATE $\widehat{\theta}_{t,t-1} \gets \theta_0$, for all $t \geq 1$.
				\STATE Set {\ttfamily Num-change-points }$ \gets 0$
			    \FOR {each time $t = 1, 2, \cdots , $}
			   \STATE Receive sample $X_t$ \\
			   \STATE   $\widehat{\theta}_{s,t} \gets \prod_{\theta}(\widehat{\theta}_{s,t-1} - \eta_{t-s}\text{clip}(X_{t} - \widehat{\theta}_{s,t-1}, \lambda))$, for every $r \leq s \leq t$.
			    \IF {$\exists s \in (r,t)$ such that $\|\widehat{\theta}_{r:s} - \widehat{\theta}_{s+1:t}\|_2^2 > \mathcal{B}\left(s-r,\frac{\delta}{2(t-r)(t-r+1)}\right) + \mathcal{B}\left(t-s-1, \frac{\delta}{2(t-r)(t-r+1)}\right)$ \COMMENT {$B(\cdot, \cdot)$ is defined in Equation (\ref{eqn:defn_B}}}
			    \STATE Set $\mathcal{A}_t$ $\gets 1$ \COMMENT {Change point detected}
			    \STATE $r \gets t+1$
			    \STATE Set {\ttfamily Num-change-points }$ \gets ${\ttfamily Num-change-points } $+1$ \COMMENT {Increment number of change-points detected}
			    \ELSE 
			    \STATE Set $\mathcal{A}_t$ $\gets 0$
			    \ENDIF
			 
			    \ENDFOR 

\end{algorithmic}
	\end{algorithm*}



\subsection{Connections to GLR}

Restating our algorithm, a change point is detected in a time-interval $[t_0,t]$ if 
\begin{align*}
    \exists s \in (t_0,t) \text{ s.t. } \| \widehat{\theta}_{t_0:s} - \widehat{\theta}_{s+1:t} \|^2 \geq \mathcal{C}(t_0, s, t, \delta),
\end{align*}
where the function $\mathcal{C}(\cdot)$ is given in Line $8$ of Algorithm \ref{algo:learn_model}. In the above re-statement, the estimates $\widehat{\theta}_{t_0:s}$ and $\widehat{\theta}_{s+1:t}$ are robust estimates of the mean based on the set of observations $\{X_{t0}, \cdots,X_s\}$ and $\{X_{s+1}, \cdots, X_t\}$ respectively. The {\ttfamily Improved-GLR} of \citep{maillard2019sequential} uses a detector that is structurally similar to the above equation except that they {\em (i)} use the empirical mean as they are dealing with sub-gaussian random variables, and {\em (ii)} use a function $\mathcal{C}(\cdot)$ derived from the Laplace method that gives confidence bounds with better dependence on time, but is not dimension free. In contrast, we use the robust mean estimator given by clipped-SGD and the function $\mathcal{C}(\cdot)$ is derived from the confidence guarantees that only require the existence of the second moment and make no other tail assumptions and yields dimension free bounds. The cost however is that the confidence bound derived from clipped SGD has a weaker dependence on time compared to that obtained by the Laplace's method \citep{maillard2019sequential}. 
\subsection{False-positive guarantee}
\label{sec:fpr_guarantee}

We will prove the following result on Algorithm \ref{algo:learn_model}. For a given process $\mathfrak{M}$, and every $r \in \mathbb{N}$, denote by the deterministic time $\tau_c^{(r)} := \inf\{ \tau_c : \tau_c > r \}$ be the first change-point after time $r$. 

\begin{theorem}[False Positives]
When Algorithm \ref{algo:learn_model} is run with parameters  $\lambda = 2G$, $\eta_t = \frac{2}{(t+\gamma)}$ for $\gamma = \max \left(120 \lambda \sigma(\sigma+1),{320\sigma^2}+1 \right)$ and $\delta \in (0,1)$, 
\begin{align*}
    \sup_{\mathfrak{M}, r}\mathbb{P}[ \exists t \in [r, \tau_c^{(r)}), \text{ s.t. } \mathcal{A}_{t}  = 1 \vert \mathcal{A}_r=1] \leq \delta,
\end{align*}
holds almost-surely.
\label{thm:fpr_main}
\end{theorem}

Proof is in Appendix in Section \ref{sec:proof_thm_fpr_main}. This result states that with probability at-most $\delta$, a true change-point \emph{does not} lie between any two consecutive detections made by the algorithm. This theorem implies the following lemma. 

\begin{lemma}
Under the conditions of Theorem \ref{thm:fpr_main}, the FPR condition in Equation (\ref{eqn:fpr_definition}) holds. 
\label{lem:fpr_connection}
\end{lemma}
The proof is in the Appendix in Section \ref{sec:proof_of_fpr_connection}. We emphasize that the guarantee in \ref{lem:fpr_connection} is a \emph{worst-case guarantee}. In other words, no matter the underlying distribution, as long as Assumptions \ref{assumption:bounded} and \ref{assumption_stochastic} are met, Algorithm \ref{algo:learn_model} will not have more than a $\delta$ fraction of false-positives.

\subsection{Worst-case Detection Delay Guarantee}

\begin{lemma}
If Algorithm \ref{algo:learn_model} is run with the parameters from Theorem \ref{thm:fpr_main}, then for every $n \in \mathbb{N}$, $\Delta > 0$ and $\delta' \in (0,1)$
\begin{multline}
   \mathcal{D}(n, \Delta, \delta') \leq \inf \bigg\{ d \in \mathbb{N} : \Delta^2 \geq \mathcal{B}\left( n-1, \frac{\delta'}{2} \right) + \\ \mathcal{B}\left( d, \frac{\delta'}{2} \right)  + \mathcal{B}\left( n-1, \frac{\delta}{2(n+d+1)(n+d)} \right) + \\ \mathcal{B}\left( d,  \frac{\delta}{2(n+d+1)(n+d)} \right) \bigg\},
   \label{eqn:delay_formula}
\end{multline}
where $\mathcal{D}(\cdot)$ and $\mathcal{B}(\cdot)$ are in Eqns (\ref{eqn:detection_delay_defn}) and (\ref{eqn:defn_B}) respectively. 
\label{lem:detection_delay}
\end{lemma}

Proof is in the Appendix in Section \ref{sec:proof_delay}. Lemma \ref{lem:detection_delay} is an \emph{upper bound on the worst case delay}. In other words, for any pre- and post-change distribution with norm of the means differing by $\Delta$, Algorithm \ref{algo:learn_model} will detect this change within delay of $\mathcal{D}(n,\Delta,\delta')$ with probability at-least $1-\delta'$. 


\begin{figure}
\centering
\begin{subfigure}{0.49\linewidth}
\centering
\includegraphics[width=0.99\linewidth]{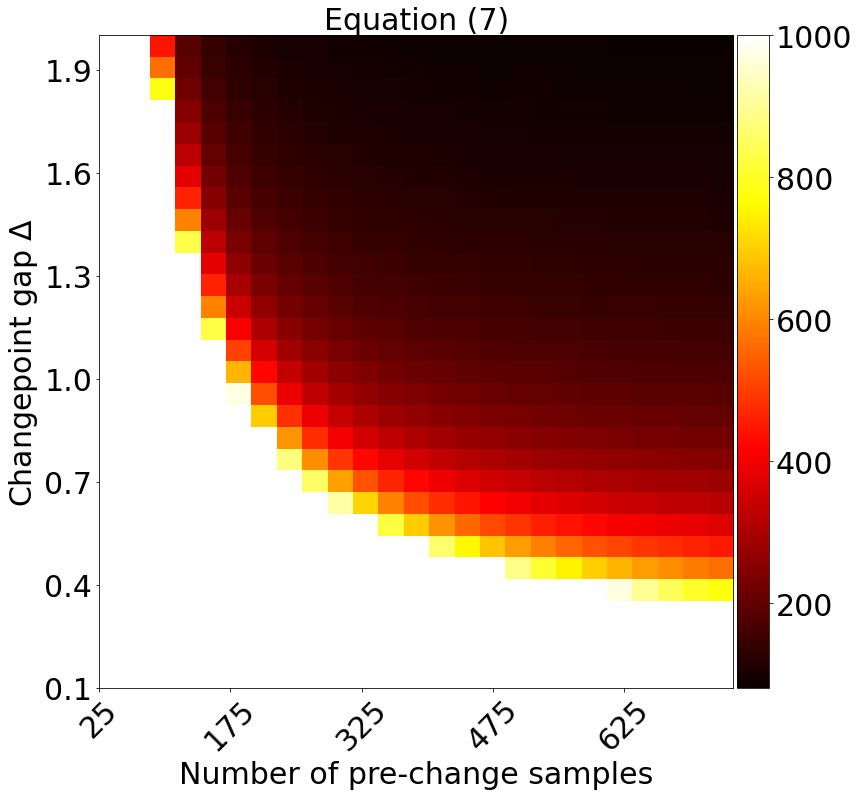}
\caption{}
\label{fig:heatmap}
\end{subfigure}
\begin{subfigure}{0.49\linewidth}
\centering
\includegraphics[width=0.93\linewidth]{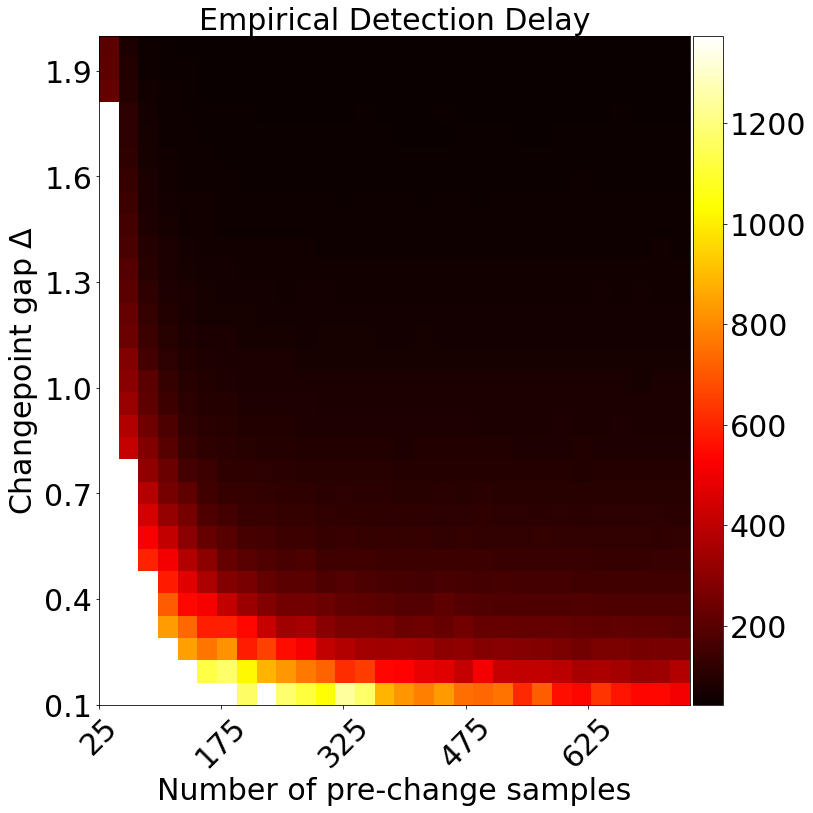}
\caption{}
\label{fig:empirical_heat_map}
\end{subfigure}
\caption{Figure $(a)$ plots the heat-map of $\mathcal{D}(n, \Delta, \delta')$ from Lemma \ref{lem:detection_delay} for fixed $\delta'=0.1$. The white cells represent infinity. Figure $(b)$ plots the $90$th quantile ($\delta'=0.1$) of the observed delay for Pareto distribution $d=32$ over $30$ runs. As can be seen, the observed detection delay in $(b)$ is much smaller than the worst case delay in $(a)$.}
\label{fig:heatmap_joint}
\end{figure}

For many specific choices of pre- and post-change distribution families however, we expect the observed detection delay to be much smaller than predicted by Lemma \ref{lem:detection_delay}.  This bound is conservative as it is worst-case over all distributions. In Figure \ref{fig:heatmap} we plot the bound in Lemma \ref{lem:detection_delay} for a fixed $\delta'=\delta=0.1$ as $n$ and $\Delta$ varies. We use the constants given in Section \ref{sec:synthetic_simulations} to plot Figure \ref{fig:heatmap}. In Figure \ref{fig:empirical_heat_map}, we plot the empirically observed detection delay for a sequence of $32$ dimensional Pareto distributed random vectors with shape parameter $2.01$. As can be seen in Figure \ref{fig:heatmap_joint}, the observed detection delay is much smaller than that indicated by Lemma $4.3$, which is a worst case over all distributions.


\begin{remark}
In the special case when the observations are Bernoulli random variables, the {\ttfamily R-BOCPD} algorithm of \citep{alami2020restarted} gives a smaller detection delay compared to ours -- our detection delay bound in \ref{lem:detection_delay} has additional poly-logarithmic factors of $\log(n/\delta)$ and sub-optimal constants compared to {\ttfamily R-BOCPD}. However, our bound holds for \emph{any} family of distributions, including high-dimensional and heavy tailed ones, while {\ttfamily R-BOCPD} can only be applied for Bernoulli distributions.
\end{remark}

\begin{corollary}[Un-detectable Change]
If $\Delta \leq \mathcal{O} \left( \frac{\log \left( \frac{n}{\delta} \right)}{\sqrt{n}} \right)$, then $   \mathcal{D}(n, \Delta, \delta') \leq \infty$ for all $\delta' \in (0,1)$, the delay bound in Lemma \ref{lem:detection_delay} is vacuous. 
\label{cor:undetectable}
\end{corollary}

\begin{remark}
The undetecable region consists of the grey/white areas of Figure \ref{fig:heatmap}. However, since Lemma \ref{lem:detection_delay} is only an upper-bound, the fact that $   \mathcal{D}(n, \Delta, \delta') = \infty$ \emph{does not imply} that our algorithm cannot detect the change (cf. Figure \ref{fig:empirical_heat_map}).
\end{remark}

\begin{remark}
In the case of sub-gaussian, exponential families, \citep{maillard2019sequential} give a lower bound for changes that not detectable by \emph{any} algorithm. When Algorithm \ref{algo:learn_model} is applied to sub-gaussian random variables from an exponential family, the detection-delay bound in Lemma \ref{lem:detection_delay} is sub-optimal by poly-logarithmic factors in $\log(n/\delta)$ compared to the lower bound. However, Algorithm \ref{algo:learn_model} and the delay bound in Lemma \ref{lem:detection_delay} holds for any class of distributions subject to Assumptions \ref{assumption_stochastic} and \ref{assumption:bounded}, while the bounds in \citep{maillard2019sequential} only applies to sub-gaussian observations from a known exponential family. 
\end{remark}

\begin{remark}
In parallel work, the FCS detector of \citep{shekhar2023sequential}, when combined with the heavy-tailed Catoni-style confidence sequences of \citep{wang2023huber} is shown to detect univariate mean changes as long as $\Delta \preceq \sqrt{\log ( \log(n)/\alpha)/n}$. Whether this rate is achievable in multivariate settings is left for future work
\end{remark}





\subsection{Change-point localization}

In practice, it is also crucial to identify the location where the change point occurred. In this section we describe how to modify Algorithm \ref{algo:learn_model} to also output the estimate of the location of change in addition to just detecting the existence of a change. Recall that for every $r \in \mathbb{N}$, $\tau_r^{(\mathcal{A})} \in \mathbb{N} \cup \{ \infty\}$ is the stopping time denoting the $r$th time, Algorithm $\mathcal{A}$ detects a change point. We modify Algorithm \ref{algo:learn_model} by additionally outputting for every $r \in \mathbb{N}$, a time interval $[s_{r;1}^{(\mathcal{A})}, s_{2;r}^{(\mathcal{A})}] \subseteq [\tau_{r-1}^{(\mathcal{A})},\tau_{r}^{(\mathcal{A})}]$ such that this is an interval that contains a change-point $\tau_c$.

\begin{figure*}
\centering
\begin{subfigure}{0.22\textwidth}
\includegraphics[width=0.99\linewidth]{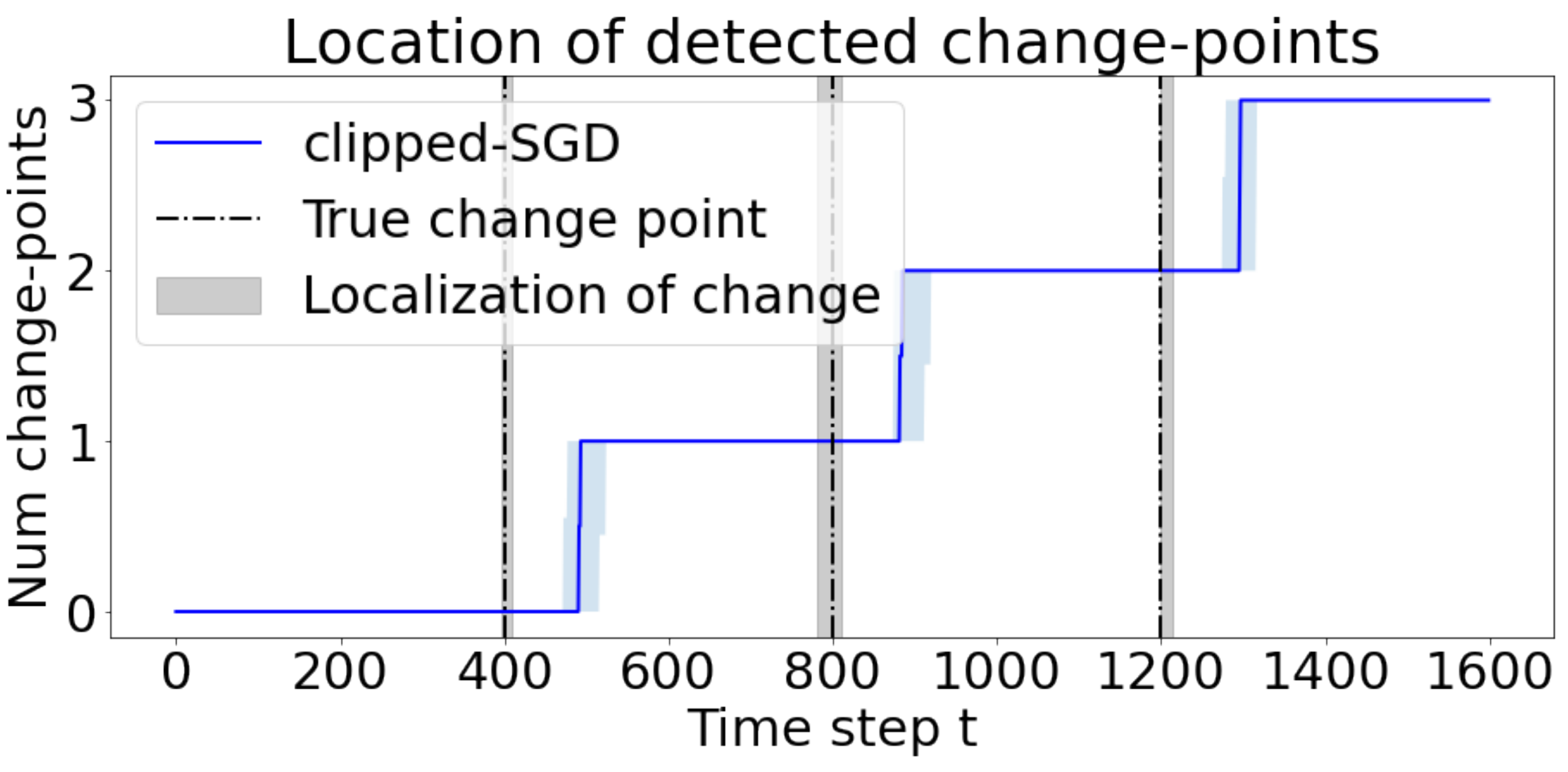}
\label{fig:local1}
\caption{Pareto $\Delta=1$}
\end{subfigure}
\begin{subfigure}{0.22\textwidth}
\includegraphics[width=0.99\linewidth]{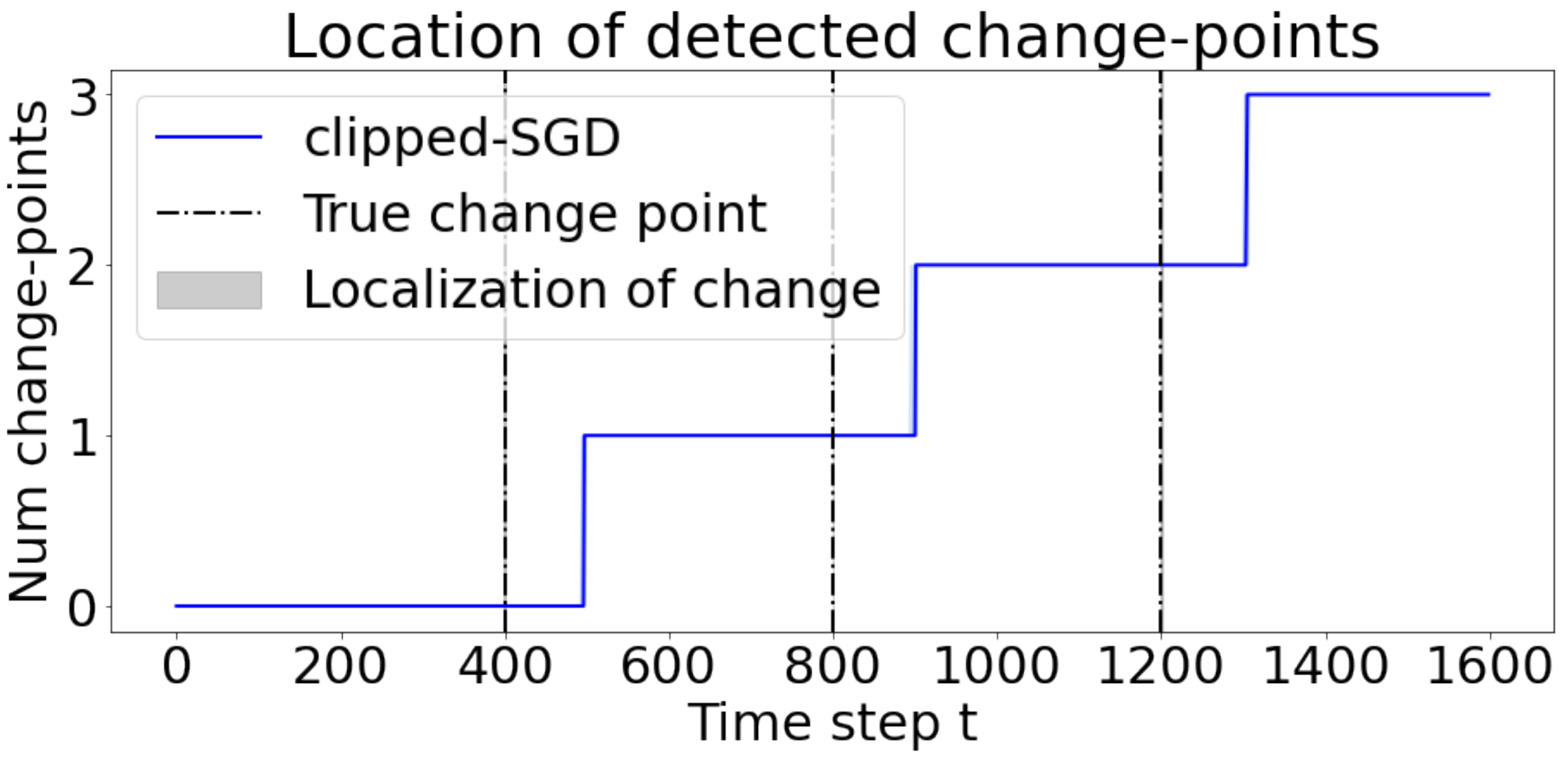}
\label{fig:local2}
\caption{Pareto $d=32, \Delta=1$}
\end{subfigure}
\begin{subfigure}{0.22\textwidth}
\includegraphics[width=0.99\linewidth]{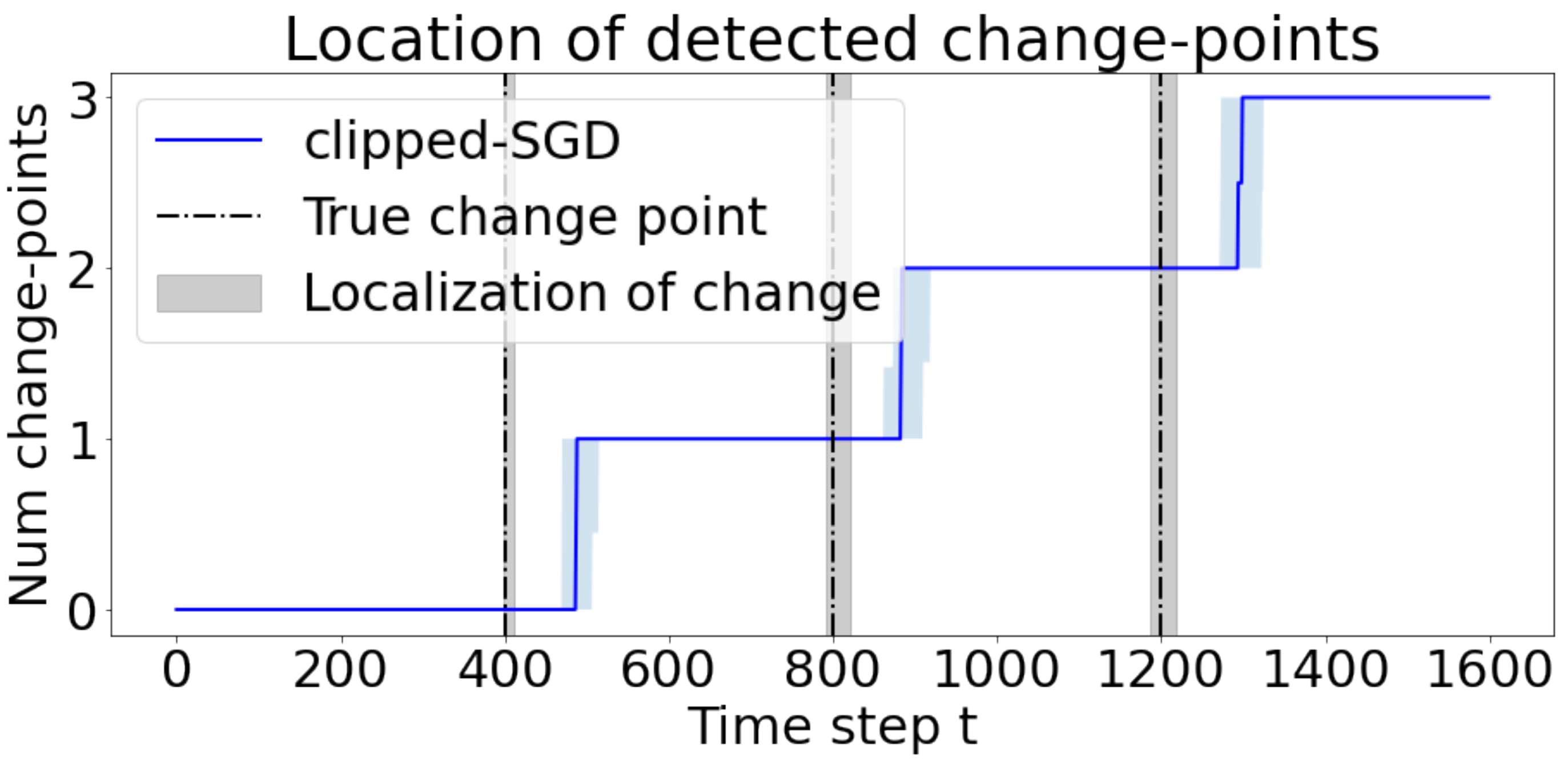}
\label{fig:local3}
\caption{Normal $\Delta=1$}
\end{subfigure}
\begin{subfigure}{0.22\textwidth}
\includegraphics[width=0.99\linewidth]{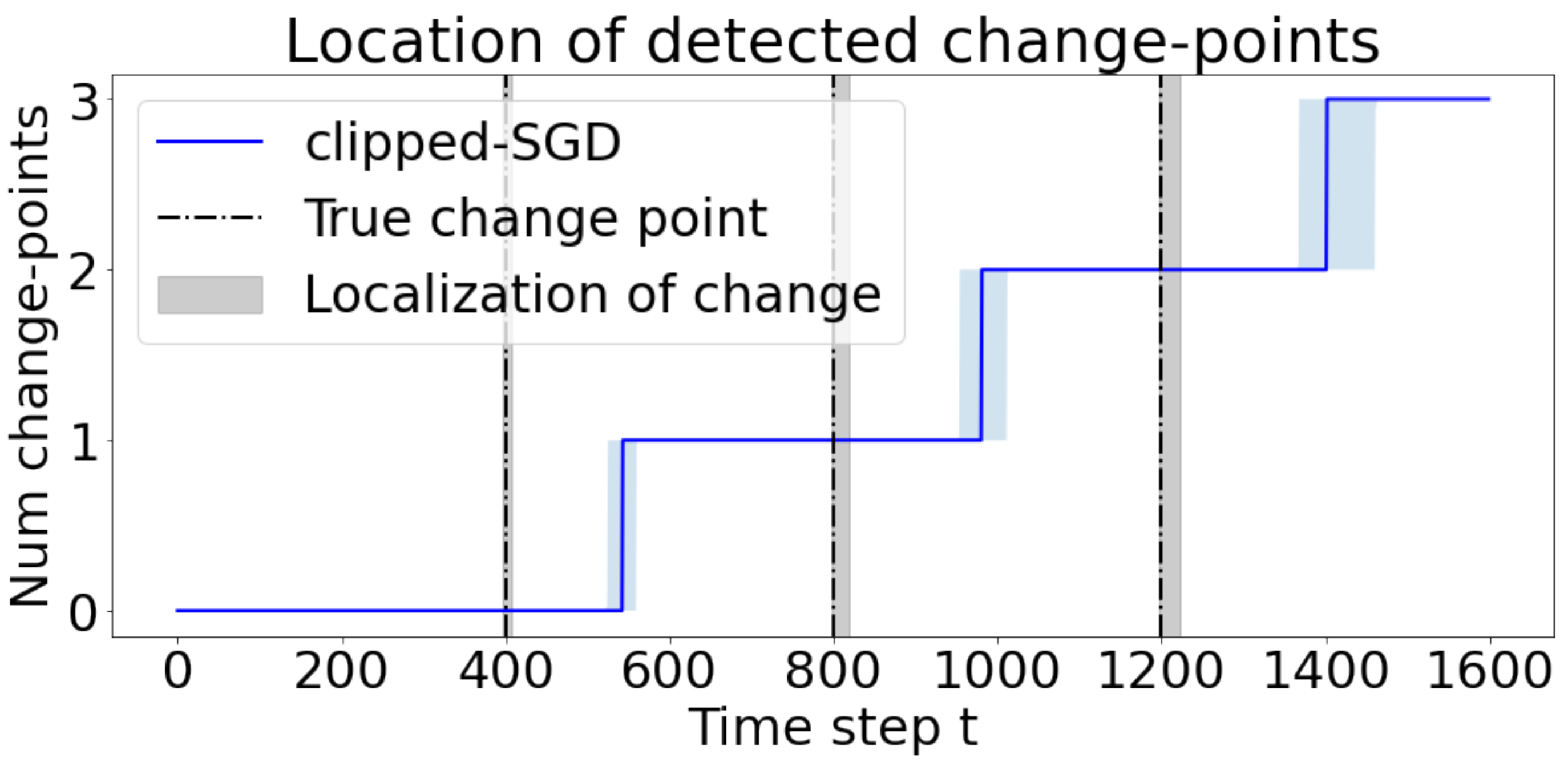}
\label{fig:local4}
\caption{Bernoulli $\Delta=0.7$}
\end{subfigure}
\caption{Plots showing that by Algorithm \ref{algo:learn_model_local} can detect and localize change-points across a variety of settings.}
\label{fig:localization}
\end{figure*}

 In order do so, we need an additional definition. For every $r < s < t$ and $\delta \in (0,1)$, denote by $\mathfrak{B}(r,s,t,\delta) \in \{0,1\}$ as the indicator variable that 
\begin{equation}
    \mathfrak{B}(r,s,t,\delta)=  \mathbf{1} \bigg(\|\widehat{\theta}_{r:s} - \widehat{\theta}_{s+1:t}\|_2^2 > \mathfrak{B_1} + \mathfrak{B_2}\bigg), \label{eqn:defn_mathfrak_B}
\end{equation}   
where $\mathfrak{B1}= \mathcal{B}\left(s-r,\frac{\delta}{2(t-r)(t-r+1)}\right)$ and $\mathfrak{B2}= \mathcal{B}\left(t-s-1, \frac{\delta}{2(t-r)(t-r+1)}\right) $.
The estimates of the location of change in a time-interval $[r,t]$ is all those time instants $s \in [r,t]$ such that $\mathfrak{B}(r,s,t,\delta)=1$. Line $12$ in Algorithm \ref{algo:learn_model_local} in Section \ref{sec:localization} in the Appendix, precisely defines the estimator. The empirical performance of this method is shown in Figure \ref{fig:localization}. We observe that this produces an accurate and sharp estimate of the change-point location in simulations.

\section{Experiments}
\label{sec:experiments}

\begin{figure*}[ht!]
\centering
\begin{subfigure}{0.22\textwidth}
\includegraphics[width=0.99\linewidth]{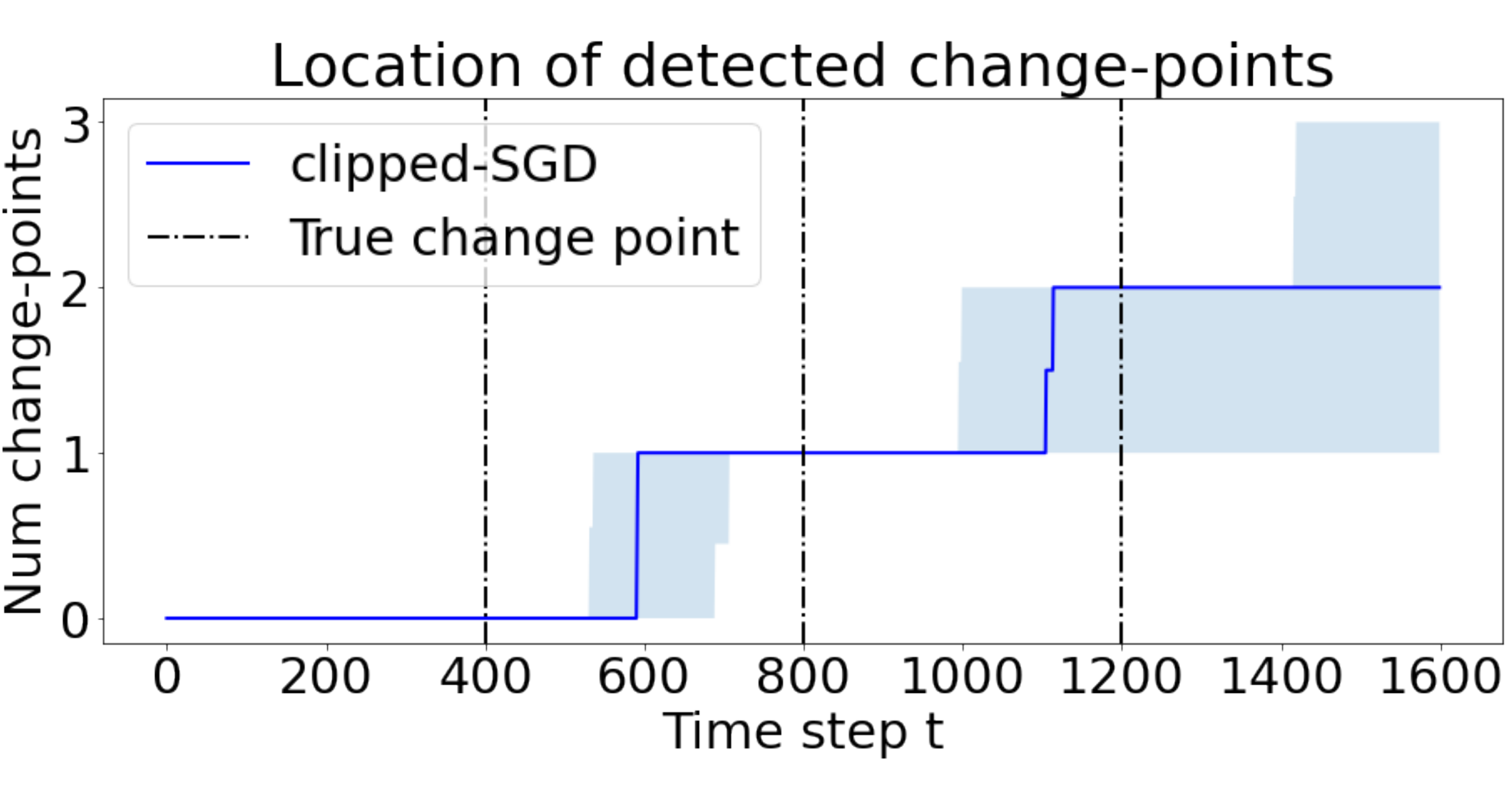}
    \caption{Pareto $\Delta=0.5$}
    \label{fig:ff1}
\end{subfigure}
\begin{subfigure}{0.22\textwidth}
\includegraphics[width=0.99\linewidth]{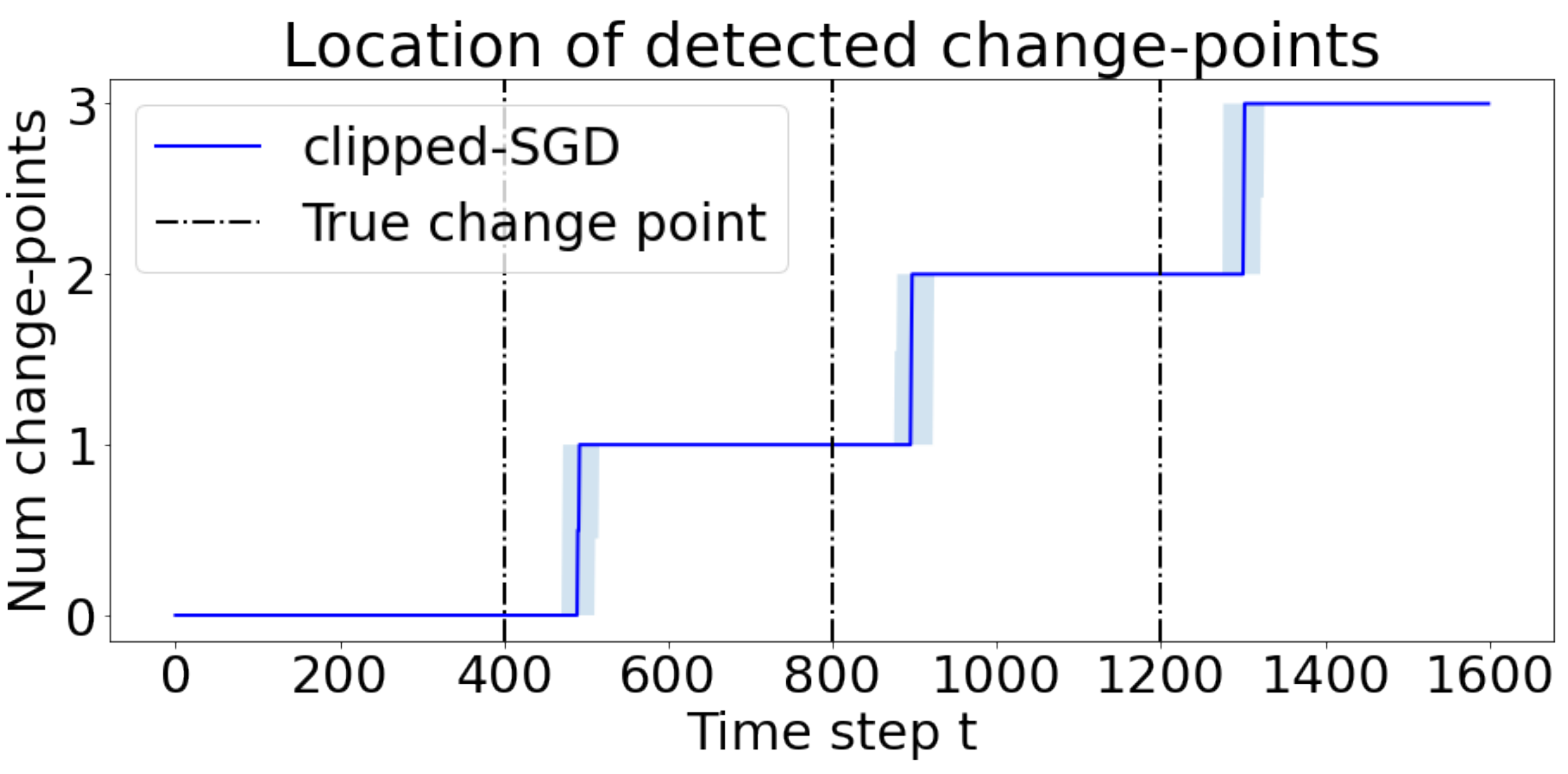}
    \caption{Pareto $\Delta=1$}
    \label{fig:ff2}
\end{subfigure}
\begin{subfigure}{0.22\textwidth}
\includegraphics[width=0.99\linewidth]{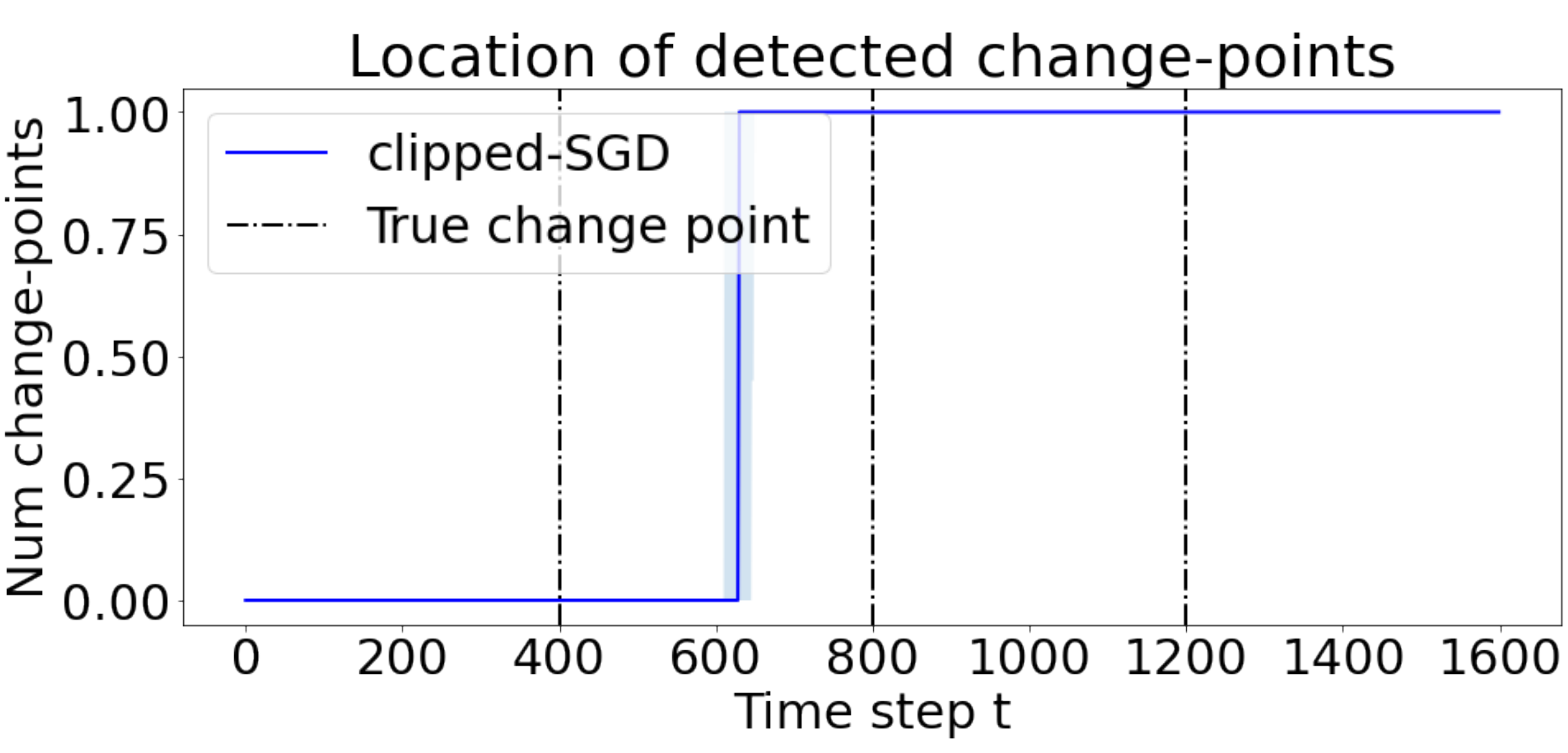}
    \caption{Pareto $d=32,\Delta=0.5$}
    \label{fig:ff3}
\end{subfigure}
\begin{subfigure}{0.22\textwidth}
\includegraphics[width=0.99\linewidth]{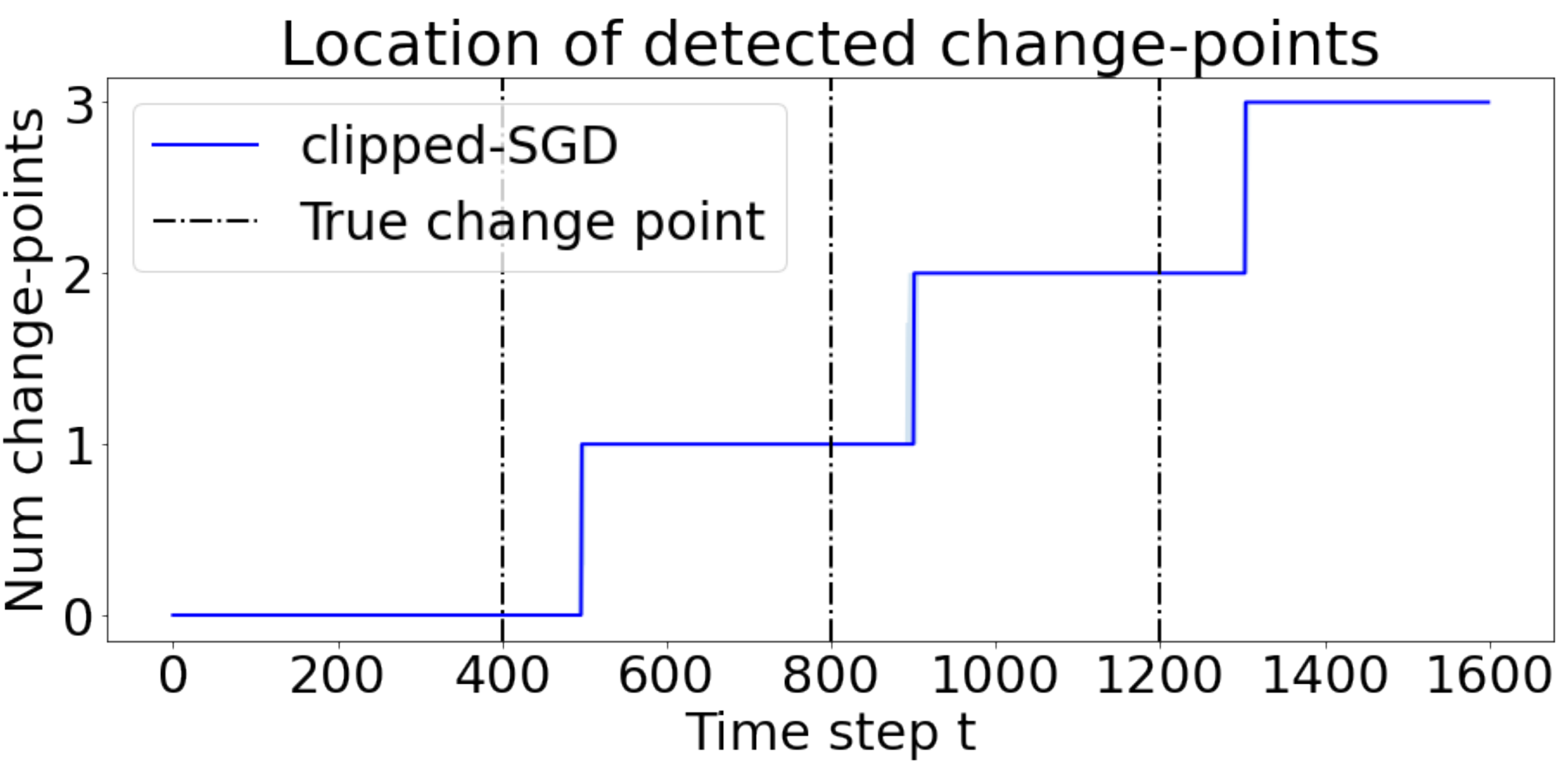}
    \caption{Pareto $d=32, \Delta=1$}
    \label{fig:ff4}
\end{subfigure}
\begin{subfigure}{0.22\textwidth}
\includegraphics[width=0.99\linewidth]{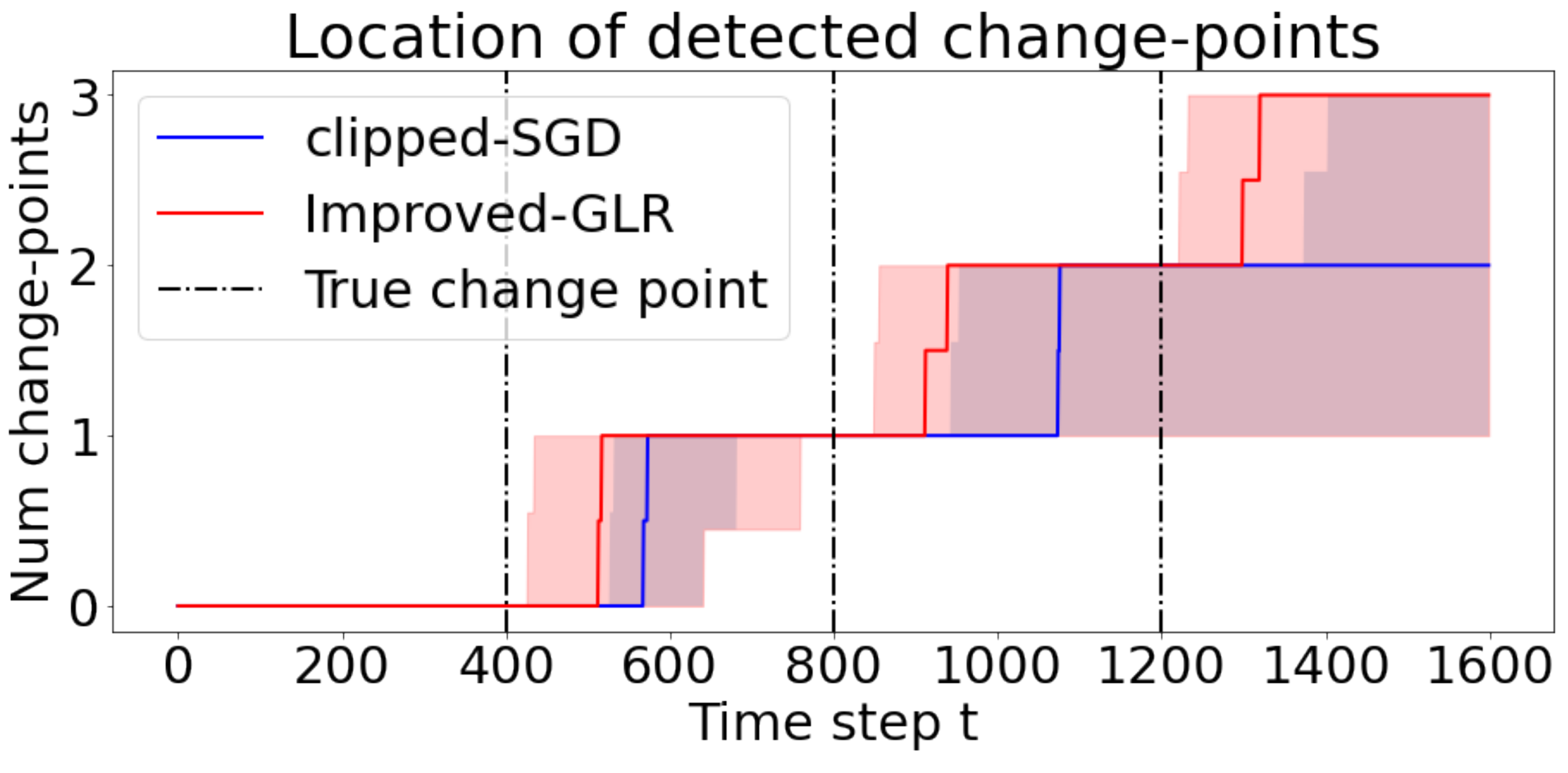}
    \caption{Normal $\Delta=0.5$}
    \label{fig:ff5}
\end{subfigure}
\begin{subfigure}{0.22\textwidth}
\includegraphics[width=0.99\linewidth]{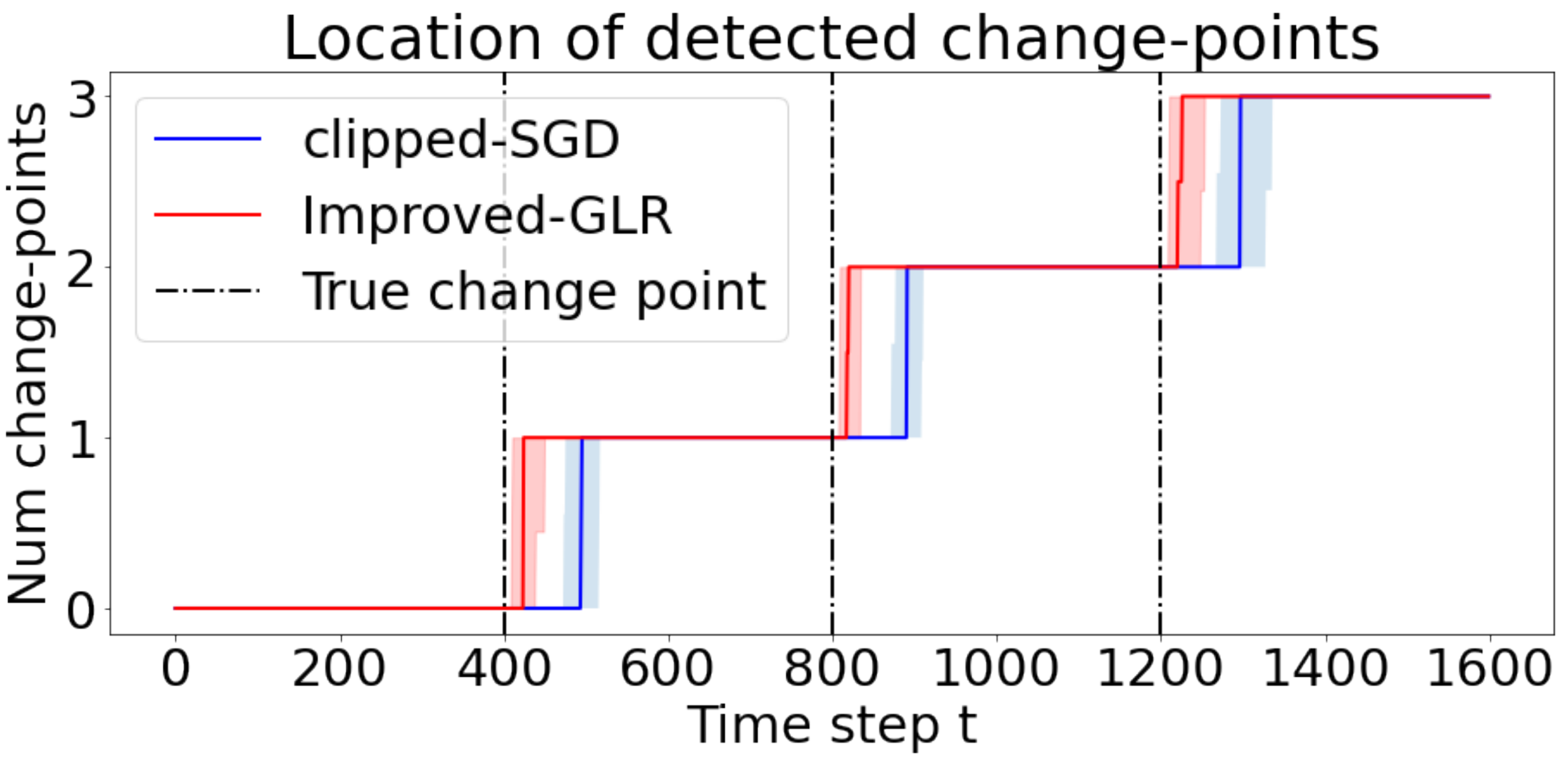}
    \caption{Normal $ \Delta=1$}
    \label{fig:ff6}
\end{subfigure}
\begin{subfigure}{0.22\textwidth}
\includegraphics[width=0.99\linewidth]{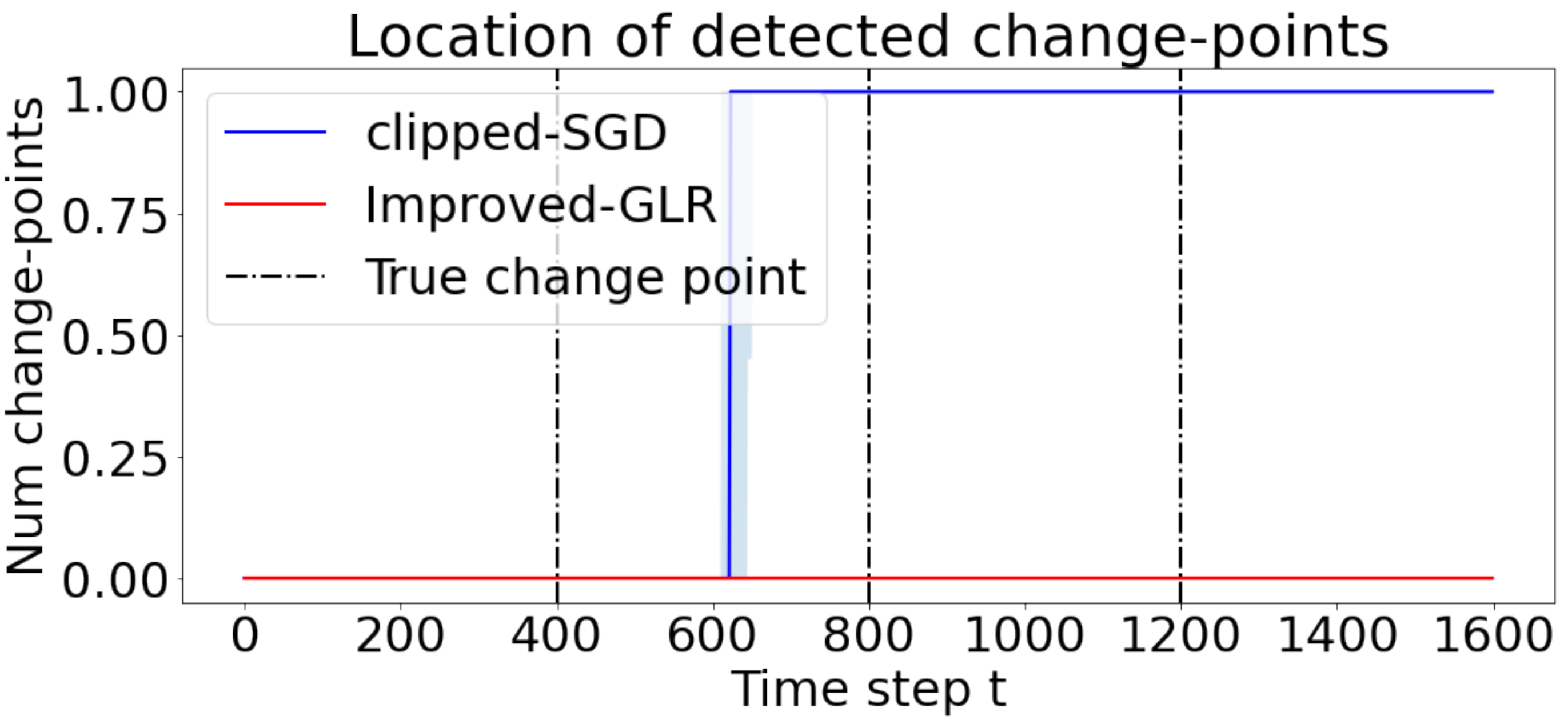}
    \caption{Normal $d=32$, $\Delta=0.5$}
    \label{fig:ff7}
\end{subfigure}
\begin{subfigure}{0.22\textwidth}
\includegraphics[width=0.99\linewidth]{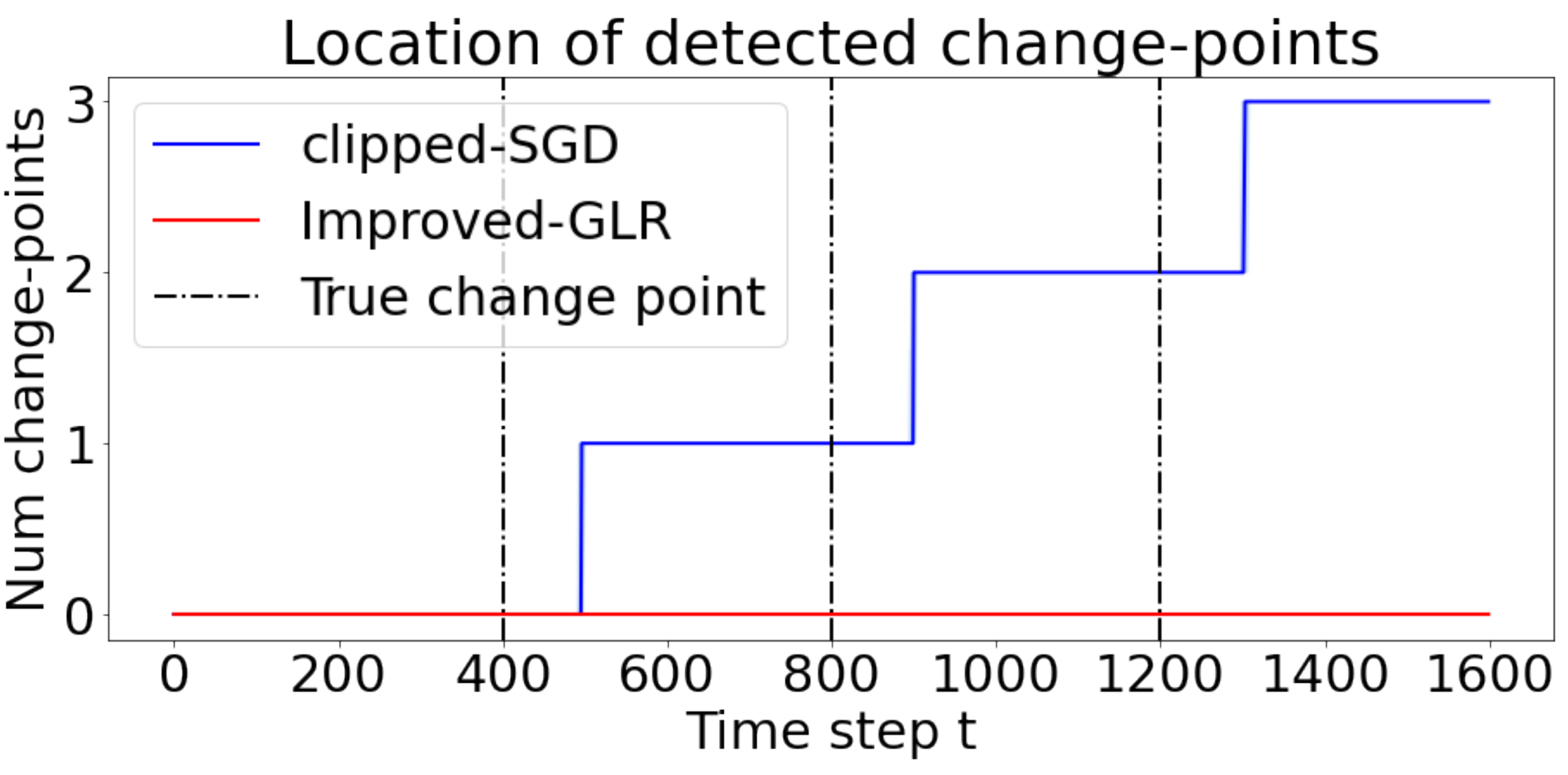}
    \caption{Normal $d=32, \Delta=1$}
    \label{fig:ff8}
\end{subfigure}
\begin{subfigure}{0.22\textwidth}
\includegraphics[width=0.99\linewidth]{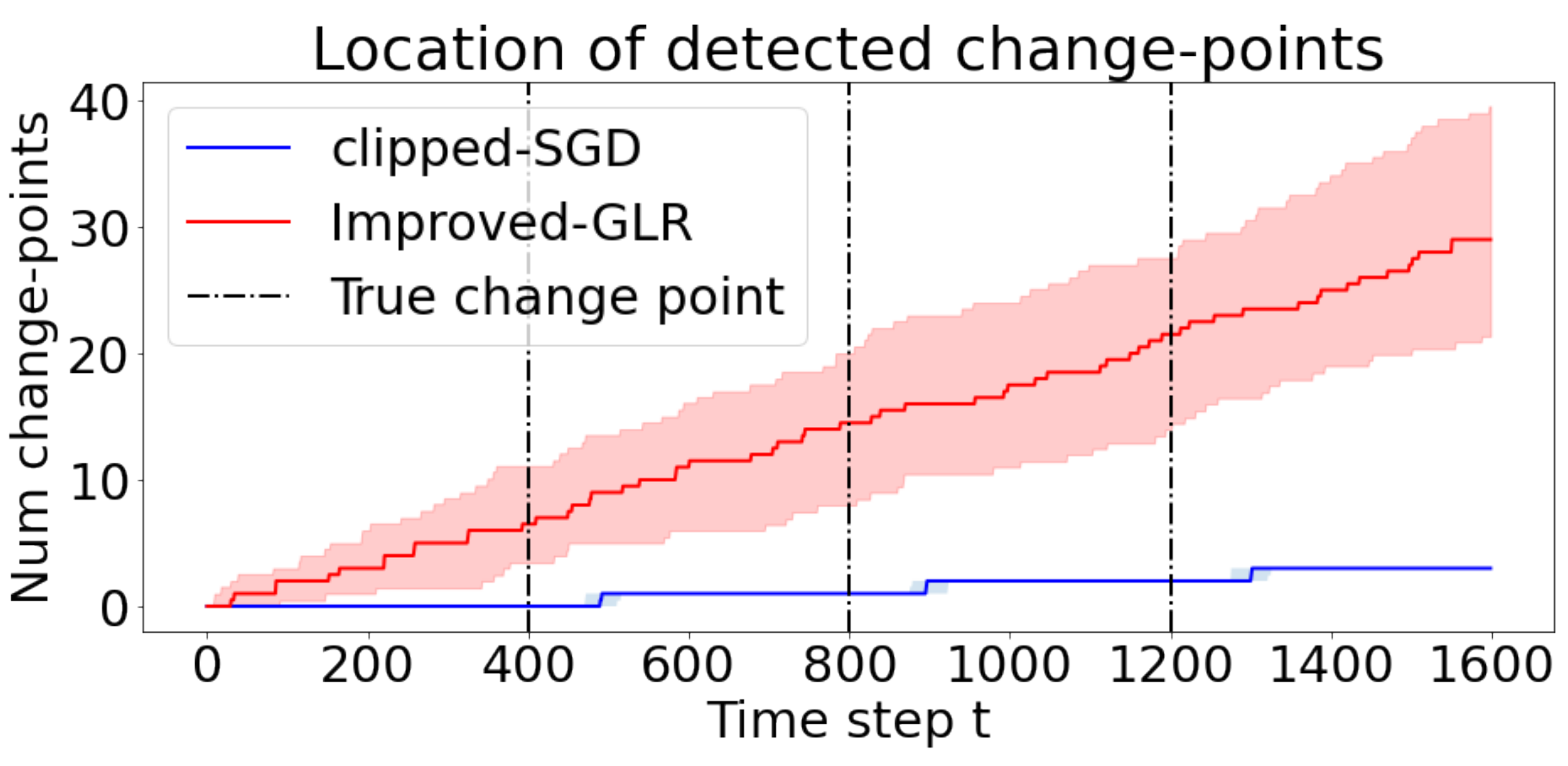}
    \caption{Pareto $ \Delta=1$}
    \label{fig:ff9}
\end{subfigure}
\begin{subfigure}{0.22\textwidth}
\includegraphics[width=0.99\linewidth]{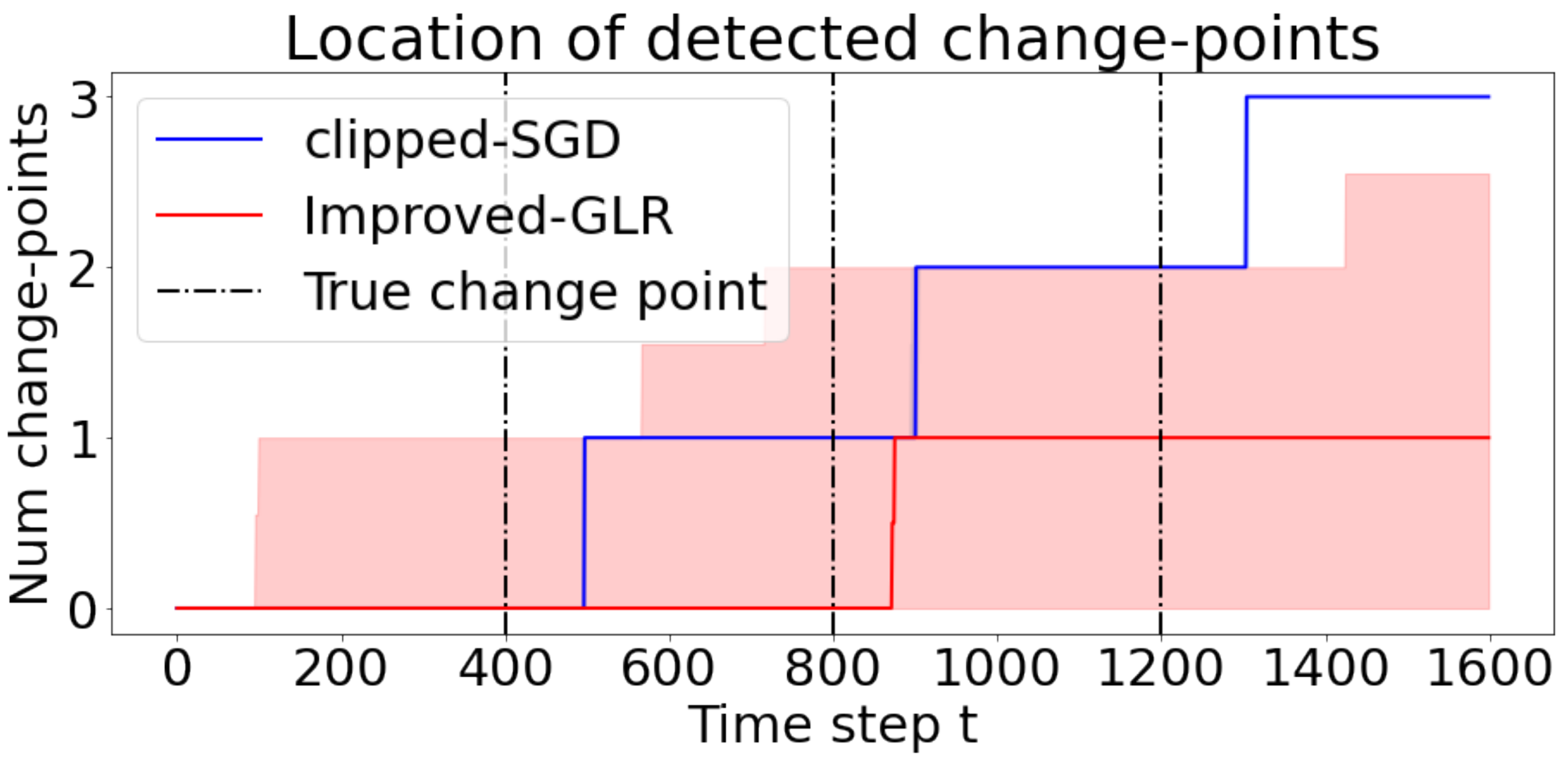}
    \caption{Pareto $d=32, \Delta=1$}
    \label{fig:ff10}
\end{subfigure}
\begin{subfigure}{0.22\textwidth}
\includegraphics[width=0.99\linewidth]{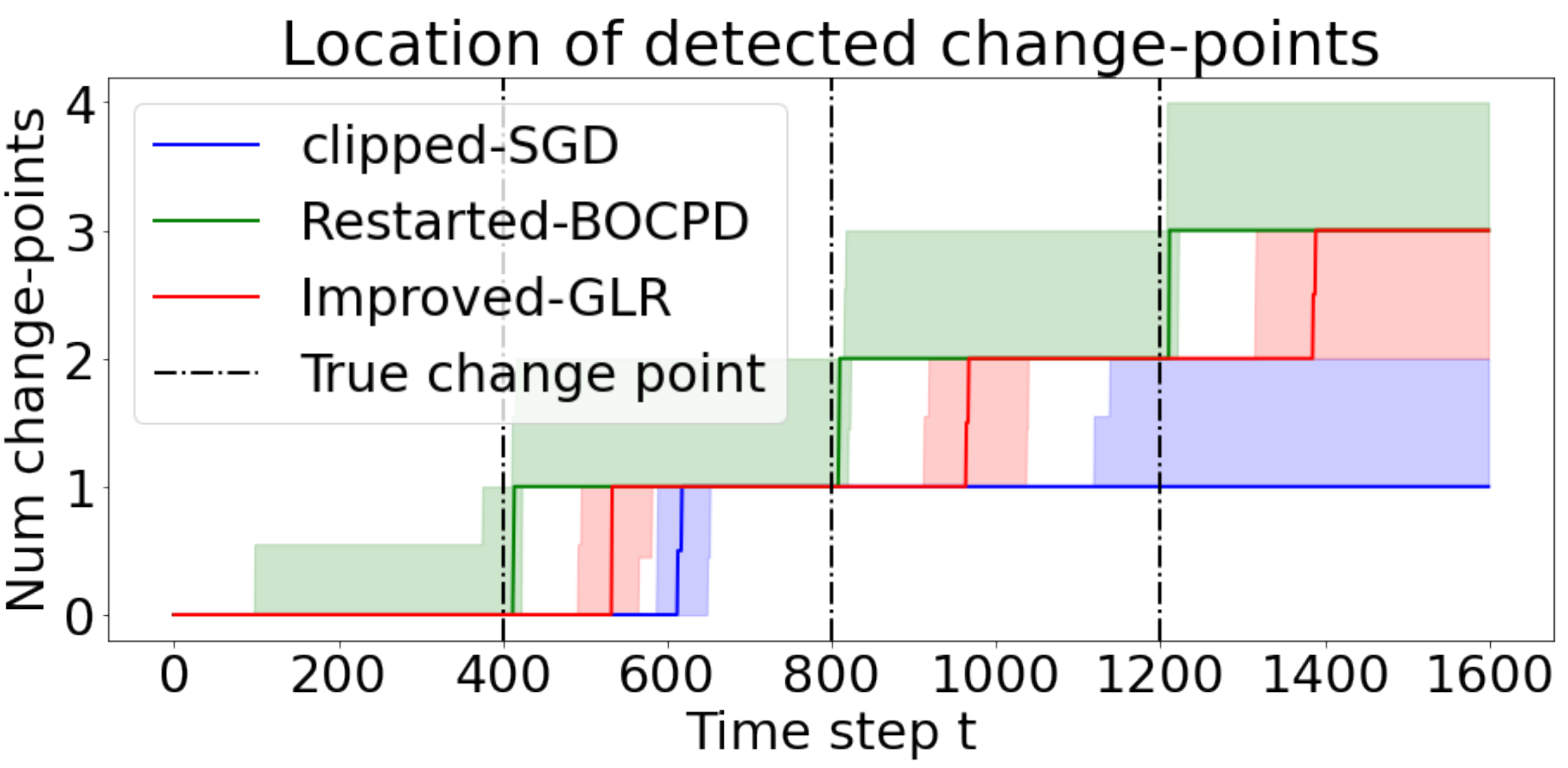}
    \caption{Bernoulli with $\Delta=0.5$}
    \label{fig:ff11}
\end{subfigure}
\begin{subfigure}{0.22\textwidth}
\includegraphics[width=0.99\linewidth]{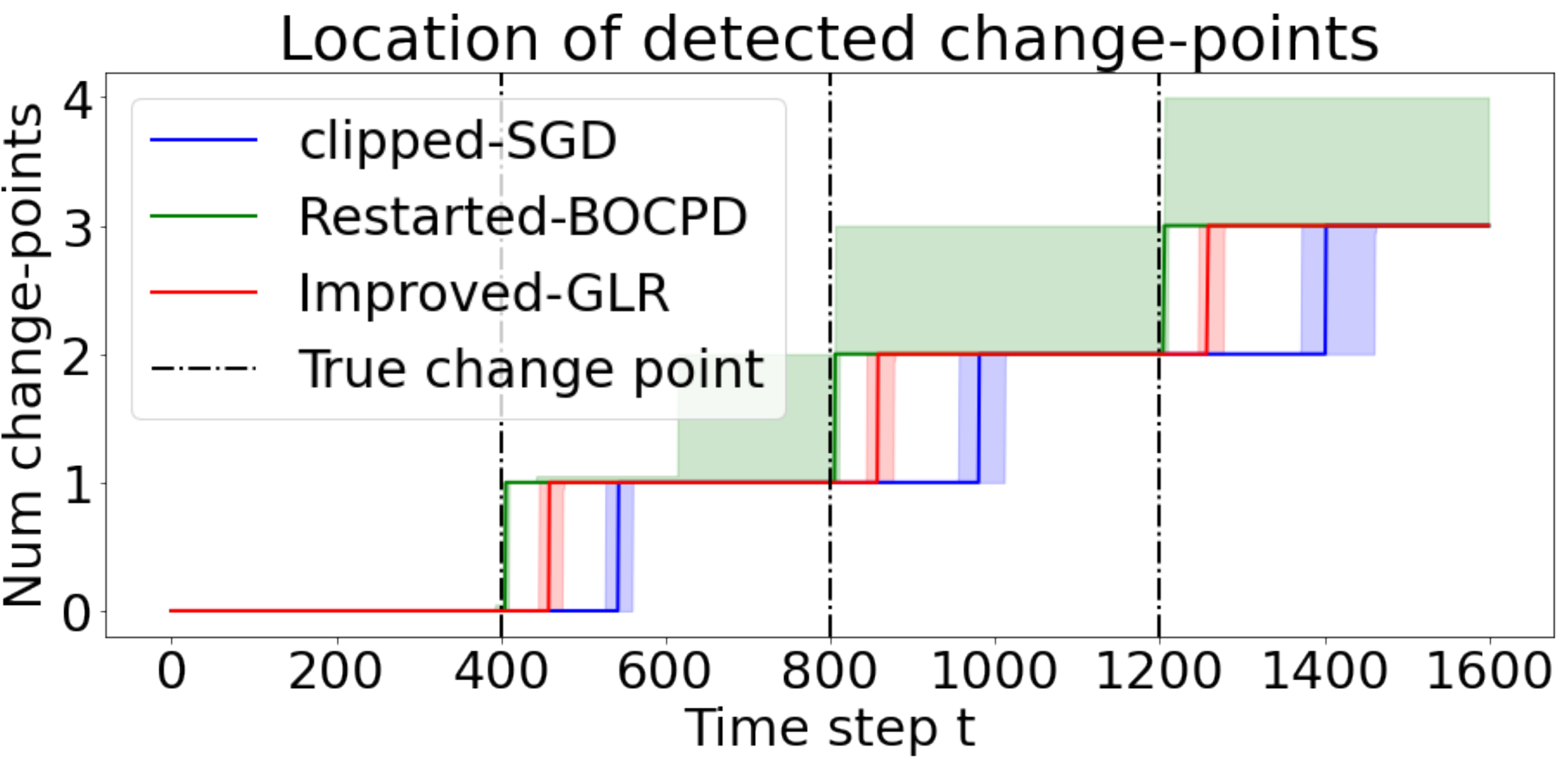}
    \caption{Bernoulli with $\Delta=0.7$}
    \label{fig:ff12}
\end{subfigure}
\caption{Empirical performance of Algorithm \ref{algo:learn_model}  in a variety of scenarios. Exact details of each plot in Section \ref{sec:synthetic_simulations}.}
\label{fig:all_figs} 
\end{figure*}

In this section we give numerical evidence to show that Algorithm \ref{algo:learn_model} can be applied across variety of settings. Line $8$ of Algorithm \ref{algo:learn_model} relies on confidence bounds for high-dimensional estimation where the global constants are not optimized. This is an artifact of the proof analysis in robust estimation \citep{lugosi2019mean, vershynin2018high}. Thus, we modify the absolute constants used in Theorem \ref{thm:fpr_main} as follows. We use $\gamma = \max \left({\color{red}4} \lambda \sigma(\sigma+1),{{\color{red}8}\sigma^2}+1 \right)$ with the color red highlighting the changes from the definition in Theorem \ref{thm:fpr_main}. The constant $C_t$ is modified as follows $     C_t  = \max(\frac{ {\color{red}0.5}\sigma^4}{G^2\lambda^2}, \frac{ {\color{red}1} \lambda \sqrt{\ln \left( \frac{2t^2(t+1)}{\delta} \right)}}{\gamma^2 G} ).
$
In addition, we use the following definition of $\mathcal{B}(\cdot, \cdot)$ 
\begin{multline}
    \mathcal{B}(t, \delta) := {\color{black}C_t}\bigg[ \frac{{\color{black}\gamma^2} G^2}{t+1} +\left(\frac{{\color{red}2}  \sigma^2}{\lambda} +  {\color{red}1}  \sigma^2 \right) \frac{1}{2(t+1)} \\+ \frac{{\color{red}2} \lambda^2 \ln \left( \frac{2t^2(t+1)}{\delta}\right)\sigma(\sigma+1) }{ (t+\gamma)\sqrt{t+1}}  \bigg],
    \label{eqn:defn_B_simulation}
\end{multline}
where $C_t$ and $\gamma$ are the modified values stated above. Further, in all simulations we assume $\Theta = \mathbb{R}^d$ to be the whole plane.

\subsection{Synthetic Simulations}
\label{sec:synthetic_simulations}

Here, demonstrate that Algorithm \ref{algo:learn_model} with choice of hyper-parameters in Equation (\ref{eqn:defn_B_simulation}) is practical and can be applied across a variety of data generating distributions -- either heavy-tailed, or high-dimensional or both and still obtains bounded false-positive rates and a much lower detection delay compared to what the conservative bound in Lemma \ref{lem:detection_delay} would indicate. 

\subsubsection{Setup}
\label{subsec:simulation_setup}
In Figure \ref{fig:all_figs}, we construct synthetic situations and introduce change-points with each change lasting $400$ time-units. In all experiments, we choose the family of distributions $\mathfrak{M}$ such that $\sigma=1$, $G = 1$2. At each time $t$, a sample is drawn from the appropriate distribution that we detail below and presented to the change-point algorithm. The true-change points and the median detection times along with the 95 percentile upper and lower confidence bands are show in Figure \ref{fig:all_figs}. These are estimated by averaging $30$ independent runs for each setting in Figure \ref{fig:all_figs}.

\textbf{Heavy-tailed distribution:} In Figures \ref{fig:ff1}, \ref{fig:ff2} and \ref{fig:ff9}, the sample at every time-point is drawn from a Pareto distribution with shape-parameter $2.01$. This implies that the third central moment of the distribution is infinity. The mean of the samples in the time-durations $t \in [0,400) \cup [800, 1200)$ is $0$ in all figures and the mean at times $t \in [400, 800) \cup [1200,1600)$ is $\Delta = 0.5, 1, 1$  respectively in Figures \ref{fig:ff1}, \ref{fig:ff2} and \ref{fig:ff9}. In Figure \ref{fig:ff3}, \ref{fig:ff4} and \ref{fig:ff10}, we consider the observation at time $t$ to be $32$ dimensional isotropic random vector with norm having Pareto distributions with shape parameter $2.01$. The mean vector at times $[0,400) \cup [800,1200)$ is $0 \in \mathbb{R}^{32}$ and at times $t \in [400,800) \cup [1200,1600)$ is $\frac{\Delta}{\sqrt{32}}[{1, \cdots, 1}] \in \mathbb{R}^{32}$, where $\Delta = 0.5, 1, 1$  respectively in Figures \ref{fig:ff3}, \ref{fig:ff4} and \ref{fig:ff10} respectively.

\textbf{Gaussian distribution:} In Figures \ref{fig:ff5} and \ref{fig:ff6}  the sample at every time-point is drawn from a unit variance Gaussian distribution. The mean of the samples in the time-durations $t \in [0,400) \cup [800, 1200)$ in all three figures is $0$ and the mean at times $t \in [400, 800) \cup [1200,1600)$ in the two figures \ref{fig:ff5} and \ref{fig:ff6} are $\Delta = 0.5$ and $\Delta=1$ respectively. In Figures \ref{fig:ff7} and \ref{fig:ff8}  we consider the observation at time $t$ to be $32$ dimensional isotropic gaussian random vector with co-variance on each axis being $1/\sqrt{32}$. The mean vector at times $[0,400) \cup [800,1200)$ is $0 \in \mathbb{R}^{32}$ and at times $t \in [400,1600) \cup [1200,1600)$ is  $\frac{\Delta}{\sqrt{32}}[{1, \cdots, 1}] \in \mathbb{R}^{32}$.

\textbf{Bernoulli distribution:} In Figures \ref{fig:ff11} and \ref{fig:ff12}, the data was $\{0,1\}$ valued Bernoulli random variable with means at times $[0,400) \cup [800,1200)$ was $0.7$ and $0.85$ respectively in the two figures, and the means at times $[400,800) \cup [1200,1600)$ are $0.3, 0.15$ respectively in the two figures.

\vspace{-2mm}
\subsubsection{Baselines}

We consider the {\ttfamily Improved-GLR} of \citep{maillard2019sequential} and {\ttfamily R-BOCPD} of \citep{alami2020restarted} as baselines since they have been empirically demonstrated to be state-of-art, and are the only other algorithms to possess finite sample, non-asymptotic FPR guarantees. The {\ttfamily Improved-GLR} can be applied to any distribution, while  its theoretical guarantees only hold for sub-gaussian distributions. The {\ttfamily R-BOCPD} algorithm is only applicable to binary data, and thus we only use it on the Bernoulli distributed setting. 

\begin{table*}[h!]
\small
    \centering
    \begin{tabular}{c|c|c||c|c|c}
         \textbf{Distribution} & $\mathbf{d}$ & $\mathbf{\Delta}$ & \textbf{Algorithm }\ref{algo:learn_model} & \textbf{Improved GLR} \citep{maillard2019sequential}  & \textbf{R-BOCPD} \citep{alami2020restarted}  \\ \hline 
         \multirow{4}{*}{Normal} & $1$ &$1$  & $274 \pm 38$ & {$\mathbf{64\pm45}$} & \multirow{4}{*}{N/A} \\
         & $32$ &$1$ & $\mathbf{300 \pm 6}$ & $2400 \pm 0$ &  \\
         & $1$ & $0.5$ & $694 \pm 191$ & $\mathbf{356 \pm 150}$  &\\ 
         & $32$ & $0.5$  & $\mathbf{1427 \pm 14}$ & $2400 \pm 1$ & \\ \hline
         \multirow{4}{*}{Pareto} &$1$&$1$& $\mathbf{296 \pm 35}$ & $19913\pm 8143$ &\multirow{4}{*}{N/A} \\
         & $32$ & $1$  & $\mathbf{302 \pm 7}$ & $1616 \pm 921$ & \\
         & $1$ & $0.5$ & $\mathbf{868 \pm 365}$ & $1891 \pm 663$ & \\
         & $32$ & $0.5$ & $\mathbf{1431 \pm 14}$ & $1667 \pm 653$ & \\ \hline 
         \multirow{2}{*}{Bernoulli} & - & $0.7$ & $515 \pm 49$ & $181 \pm 23$ & $\mathbf{23 \pm 479}$ \\
         & - & $0.5$ & $1509 \pm 53$ & $1466 \pm 762$ & $\mathbf{63 \pm 380}$ \\ \hline
         
    \end{tabular}
    \caption{Quantitative summary of Figure \ref{fig:all_figs} by comparing regret, where lower is better. Our method achieves lower regret across variety of settings of distribution, dimension and change magnitude.}
    \label{tab:fig_table}
\end{table*}

\subsubsection{Results}

 \textbf{Figure \ref{fig:all_figs} shows that our algorithm is the only one to attain bounded FPR across  heavy-tailed, Gaussian, high dimensional and Bernoulli distribution.}
 
For Pareto distribution, Figures \ref{fig:ff8} and \ref{fig:ff10} show that the {\ttfamily Improved-GLR} algorithm  has a large number of False Positives. Intuitively this occurs because the {\ttfamily Improved-GLR} algorithm assumes sub-gaussian tails and thus large deviations that are typical for the heavy-tailed Pareto distributions are mistaken for a change. (See also Figure \ref{fig:sample_path}). In contrast, from Figures \ref{fig:ff1}, \ref{fig:ff2}, \ref{fig:ff3}, \ref{fig:ff4} and \ref{fig:ff10}, we see that our algorithm consistently attains bounded false-positive rates and finite detection delay guarantees across choices of $\Delta$ and dimension $d$. 

On gaussian distributed data, both our algorithm \ref{algo:learn_model} and the {\ttfamily Improved-GLR}  obtains similar performance in-terms of false-positive rates. However, the  the median detection time of our algorithm is larger than the 95th percentile detection time of {\ttfamily Improved-GLR}. In Bernoulli distributed data, all methods attain similar False-positive guarantees; however, the specialized algorithm of {\ttfamily R-BOCPD} is superior in terms of detection delay compared to ours and the {\ttfamily Improved-GLR}.



%






{\color{black}

In Table \ref{tab:fig_table}, we summarize Figure \ref{fig:all_figs} by measuring  \emph{regret}. For any OCPD algorithm $\mathcal{A}$, we can define a function $R^{(\mathcal{A})} : [T] \to \mathbb{N}$ where $R^{(\mathcal{A})}(t) = \sum_{s \leq t} \mathcal{A}_s$ is the total number of change-points detected upto time $t$. Similarly, for any $t \in [T]$, the ground-truth function $R^*(t) = \max \{ c : \tau_c \leq t \}$ is the number of true changes till time $t$. The regret of  algorithm $\mathcal{A}$ is defined as $\sum_{t=1}^T | R^{\mathcal{A}}(t) - R^*(t)|$. This measure is non-negative and is $0$ if and only if the output of the algorithm matches the ground truth. In Table \ref{tab:fig_table}, we give the median value of regret along with $95$\% confidence interval.
We observe in Table \ref{tab:fig_table} that our method achieves lower regret across a variety of situations - whether the data is heavy-tailed, light tailed high dimensional or discrete. 


}

\subsubsection{Change-point localization}

In Figure \ref{fig:localization}, we demonstrate sharpness of change-point localization (detailed in Algorithm \ref{algo:learn_model_local}). The setting in Figure \ref{fig:localization} is identical to that of Figure \ref{fig:all_figs} with the boundary of the shaded region representing the $5$th quantile for the starting point and the $95$th quantile for the ending point of the change location interval output in Line $12$ of Algorithm \ref{algo:learn_model_local}. The localization region is biased towards the right, which is expected since our algorithm is designed to minimize false positives even in the worst-case. 

\begin{figure}
    \centering
    \includegraphics[width=0.8\linewidth]{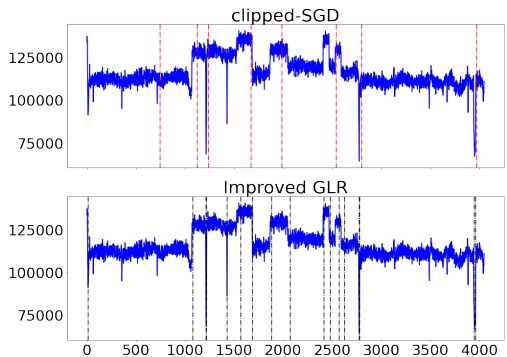}
    \caption{Performance of change-point detection of Algorithm \ref{algo:learn_model} and the {\ttfamily Improved-GLR} on real data.}
    \label{fig:real_data}
\end{figure}

\subsection{Real-Data}

In Figure \ref{fig:real_data} we show the performance of Algorithm \ref{algo:learn_model} and the {\ttfamily Improved-GLR} on the well-log dataset \citep{oruanaidh1996numerical}. This dataset consists of $4050$ measurements in the range $[6 \times 10^4, 10^5]$ of nuclear-magnetic-response taken during drilling of a well. The data
are used to interpret the geophysical structure of the
rock surrounding the well. The variations in mean
reflect the stratification of the earth’s crust. 
We process the data by dividing it by $10^{4.5}$ and run Algorithm \ref{algo:learn_model} with $G=10$, $\sigma=1$ and {\ttfamily Improved GLR} with $\sigma=1$. The detected change-points are shown in Figure \ref{fig:real_data}. Figure \ref{fig:real_data} shows that  Algorithm \ref{algo:learn_model} is comparable to {\ttfamily Improved-GLR} in terms of false-positives.

\section{Conclusions }

We introduced a new method based on clipped-SGD, to  detect change-points with guaranteed finite-sample FPR, without parametric or tail assumptions. The key technical contribution is to give an anytime online mean estimation algorithm, that provides a confidence bound for the mean at all confidence levels simultaneously. We also give a finite-sample, high probability bound on the detection delay as a function of the gap between the means and number of pre-change observations. We further corroborate empirically that ours is the only algorithm to detect change-points with bounded FPR, across multi-dimensional heavy tailed, gaussian or binary-valued data streams. 

Our work opens several interesting directions for future work. Obtaining sharp confidence intervals for estimating the mean of a random vector without the existence of variance was shown in \citep{cherapanamjeri2022optimal, wang2022catoni}. Extending the tools from therein to further relax the second moment assumption we considered is a natural direction of future work.  Another open question is to see if the martingale methods can be extended to the high-dimensional to get dimension free confidence bounds. Further, we observe in simulations that our method attains `sharp' localization empirically. Understanding the three-way trade-off between sharpness of localization, FPR and detection delay is an important area of future work. 

{
\textbf{Acknowledgements} AS thanks Aaditya Ramdas for several useful comments that improved the presentation.}
\balance
\bibliography{change-point-detect}

\onecolumn 

\appendix 
\section{Change-point Localization}
\label{sec:localization}



	
	\begin{algorithm*}[htb]
		\caption{{ Online {\ttfamily Clipped-SGD} Change Point Detection and Localization}}
		\label{algo:learn_model_local}

		\begin{algorithmic}[1]
				\STATE \textbf{Input}: {  $(\eta_t)_{t \geq 1}$,  $\lambda > 0$, $\theta_0 \in \Theta$, $\delta \in (0,1)$ FPR guarantee } 
				\STATE $r \gets 1$
				\STATE $\widehat{\theta}_{t,t-1} \gets \theta_0$, for all $t \geq 1$.
				\STATE Set $ \tau_c^{(0)} \gets 0$
				\STATE Set {\ttfamily Num-change-points }$ \gets 0$
			    \FOR {each time $t = 1, 2, \cdots , $}
			   \STATE Receive sample $X_t$ \\
			   \STATE   $\widehat{\theta}_{s,t} \gets \prod_{\theta}(\widehat{\theta}_{s,t-1} - \eta_{t-s}\text{clip}(X_{t} - \widehat{\theta}_{s,t-1}, \lambda))$, for every $r \leq s \leq t$.
			    \IF {$\exists s \in (r,t)$ such that $\|\widehat{\theta}_{r:s} - \widehat{\theta}_{s+1:t}\|_2^2 > \mathcal{B}\left(s-r,\frac{\delta}{2(t-r)(t-r+1)}\right) + \mathcal{B}\left(t-s-1, \frac{\delta}{2(t-r)(t-r+1)}\right)$ \COMMENT {$B(\cdot, \cdot)$ is defined in Equation (\ref{eqn:defn_B}}}
			    \STATE Set \textbf{Restart}$_t$ $\gets 1$ \COMMENT {Change point detected}
			    \STATE Set {\ttfamily Num-change-points }$ \gets ${\ttfamily Num-change-points } $+1$ \COMMENT {Increment number of change-points detected}
			    {\color{red}\STATE Output time interval $[\inf\{ s \in (r,t) \text{ s.t. } \mathfrak{B}(r,s,t,\delta)=1 \}, \sup \{ s \in (r,t) \text{ s.t. } \mathfrak{B}(r,s,t,\delta)=1\}]$ as the location of the change-point \COMMENT {$\mathfrak{B}()$ defined in Equation (\ref{eqn:defn_mathfrak_B})}}
			    \STATE $r \gets t+1$
			    \ELSE 
			    \STATE Set \textbf{Restart}$_t$ $\gets 0$
			    \ENDIF
			 
			    \ENDFOR 

\end{algorithmic}
	\end{algorithm*}


\section{Proof for Robust Estimation in Theorem \ref{thm:main_mean_est}}
\label{sec:mean_estimation_proofs}


We follow the same proof architecture as that of Proof of \citep{tsai2022heavy}. Throughout the proof, we let $m=1$ to be the strong convexity parameter of the quadratic loss function $x \to \frac{1}{2} \| x - x_0 \|^2$, for some $x_0 \in \mathbb{R}^d$.

Fix a time $t \in \mathbb{N}$. We define a sequence of random variable $(\psi_t)_{t \geq 1}$  as follows. 
\begin{align*}
    {\psi}_t := \text{clip}( (X_t- \widehat{\theta}_{t-1}), \lambda) - (\theta^* - \widehat{\theta}_{t-1}),
\end{align*}


Consider any time $t$. We have 
\begin{align}
    \| \theta_{t} - \theta^* \|_2^2 &=  \| \prod_{\Theta}(\widehat{\theta}_{t-1} - \eta_t \text{clip}(X_t - \widehat{\theta}_{t-1}, \lambda)) - \theta^*\|_2^2, \\
    &\stackrel{(a)}{\leq}  \| \widehat{\theta}_{t-1} - \eta_t \text{clip}(X_t - \widehat{\theta}_{t-1}, \lambda) - \theta^* \|_2^2, \\
    &= \|\widehat{\theta}_{t-1} - \eta_t({\psi}_t + (\theta^* - \widehat{\theta}_{t-1})) - \theta^*\|_2^2, \nonumber \\
    &= \|\widehat{\theta}_{t-1} - \theta^*\|_2^2 + \eta_t^2 \| {\psi}_t + (\theta^* - \widehat{\theta}_{t-1})\|_2^2 - 2 \eta_t \langle \widehat{\theta}_{t-1} - \theta^*, {\psi}_t + (\theta^* - \widehat{\theta}_{t-1}) \rangle, \nonumber \\
    &\stackrel{(b)}{\leq} \|\widehat{\theta}_{t-1} - \theta^*\|_2^2 + 2\eta^2_t \| {\psi}_t \|_2^2 + 2 \eta^2_t \|(\theta^* - \widehat{\theta}_{t-1})\|_2^2 - 2 \eta_t \langle \widehat{\theta}_{t-1} - \theta^*, {\psi}_t+ (\theta^* - \widehat{\theta}_{t-1}) \rangle, \label{eqn:proof_decomposition_first_step_mean_estimate}
\end{align}
Step $(a)$ follows since $\Theta$ is a convex set, $\|\mathcal{P}_{\Theta}(\widehat{\theta}_t) - \theta^*\| \leq \| \widehat{\theta}_t - \theta^*\|$, since $\theta^* \in \Theta$. In step $(b)$, we use the fact that $\|a+b\|_2^2 \leq 2\|a\|_2^2 + 2 \|b\|_2^2$, for all $a,b \in \mathbb{R}^d$. 
Substituting Equation (\ref{eqn:convexity_bound_inner_prod}) into (\ref{eqn:proof_decomposition_first_step_mean_estimate}), we get that 
\begin{multline*}
    \|\theta^* - \theta_{t} \|_2^2 \leq  \|\widehat{\theta}_{t-1} - \theta^*\|_2^2 + 2\eta^2_t \| {\psi}_t \|_2^2  - 2 \eta_t \langle \widehat{\theta}_{t-1} - \theta^*_{t}, {\psi}_t  \rangle
     \\ + 2 \eta^2_t \left( (M+m) \langle (\theta^* - \widehat{\theta}_{t-1}), \widehat{\theta}_{t-1} - \theta^*_t\rangle - mM \| \widehat{\theta}_{t-1} - \theta^*\|_2^2 \right)- 2\eta_t \langle (\theta^* - \widehat{\theta}_{t-1}), \widehat{\theta}_{t-1} - \theta^*_t \rangle.
\end{multline*}
Re-arranging the equation above yields 
\begin{multline*}
    \|\theta^* - \theta_{t} \|_2^2 \leq (1-2\eta^2_t mM)\|\widehat{\theta}_{t-1} - \theta^*\|_2^2 + 2\eta^2_t \| {\psi}_t \|_2^2    - 2 \eta_t \langle \widehat{\theta}_{t-1} - \theta^*, {\psi}_t \rangle  \\ - 2 \eta_t(1 - \eta_t \left( (M+m) \right)\langle (\theta^* - \widehat{\theta}_{t-1}), \widehat{\theta}_{t-1} - \theta^*\rangle.
\end{multline*}
Further substituting Equation (\ref{eqn:convexity_gradient_bound}) into the display above yields that 
\begin{align*}
    \|\theta^* - \widehat{\theta}_t\|_2^2 &\leq  (1-2\eta_t m + 2 \eta^2_t m^2 ) \| \widehat{\theta}_{t-1} - \theta^* \|_2^2  + 2 \eta^2_t \| {\psi}_t \|_2^2 - 2\eta_t \langle \widehat{\theta}_{t-1} - \theta^*, {\psi}_t \rangle, \\
    &\leq (1-\eta_t m  ) \| \widehat{\theta}_{t-1} - \theta^* \|_2^2  + 2 \eta^2_t \| {\psi}_t \|_2^2 - 2\eta_t \langle \widehat{\theta}_{t-1} - \theta^*, {\psi}_t \rangle,
\end{align*}
where the inequality comes from the fact that if $\eta_t m < 1 \implies 2 \eta_t m - 2\eta^2_t m^2 > \eta m$. 
\begin{align}
        \|\theta^* - \widehat{\theta}_t\|_2^2 \leq  (1-\eta_t m ) \| \widehat{\theta}_{t-1} - \theta^* \|_2^2 +    2 \eta^2_t \| {\psi_t} \|_2^2  - 2\eta_t \langle \widehat{\theta}_{t-1} - \theta^*, {\psi}_t  \rangle.
        \label{eqn:proof_decomp_disp_1_mean_estimate}
\end{align}

Unrolling the recursion yields, 
\begin{align*}
                \|\theta^* - \widehat{\theta}_t\|_2^2 \leq   \prod_{u=1}^{t}(1-\eta_u m ) \| \theta_1 - \theta^* \|_2^2   +  2 \eta^2_t \sum_{s=1}^{t-1}\prod_{u=1}^{s}(1-\eta_{t-u+1} m ) \| {\psi}_{t-s+1} \|_2^2  \\ - 2\eta_t \sum_{s=1}^{t-1}\prod_{u=1}^{s}(1-\eta_{t-u+1} m ) \langle \theta_{t-s} - \theta^*, {\psi}_{t-s+1}  \rangle .
\end{align*}

Using the fact that $\prod_{u=1}^{s}(1-\eta_{t-u+1} m ) = \frac{(t-s+\gamma-3)(t-s+\gamma-2)}{(t+\gamma)(t+\gamma-1)}$, we get that
\begin{align}
                \|\theta^* - \widehat{\theta}_t\|_2^2 \leq   \frac{ (\gamma-2)(\gamma-1)\| \theta_1 - \theta^* \|_2^2}{(t+\gamma)(t+\gamma-1)}   \\ - 2\eta_t \sum_{s=1}^{t-1}\frac{(t-s+\gamma-3)(t-s+\gamma-2)\langle \theta_{t-s} - \theta^*, {\psi}_{t-s+1}  \rangle}{(t+\gamma)(t+\gamma-1)}  . \label{eqn:proof_one_step_unrolled_mean_estimate}
\end{align}

Denote by $\psi_t:= \psi_t^{(b)} + \psi_t^{(v)}$, where $\psi_t^{(b)} := \mathbb{E}_{Z_t}[ \psi_t \vert \mathcal{F}_{t-1}]$ and $\psi_t^{(v)} := \psi_t - \psi_t^{(b)}$. Using this in the display above and using that fact that $\|a+b\|_2^2 \leq 2 \|a\|_2^2 + 2 \|b\|_2^2$, we get 
\begin{align}
                \|\theta^*_t - \theta\|_2^2 &\leq   \frac{ (\gamma-1)(\gamma-2)\| \theta_1 - \theta^* \|_2^2}{(t+\gamma)(t+\gamma-1)}  + 4  \eta^2_t \sum_{s=1}^{t-1}\frac{(t-s+\gamma-3)(t-s+\gamma-2)\| {\psi}_{t-s+1} \|_2^2}{(t+\gamma)(t+\gamma-1)}  +    \nonumber \\ & - 2\eta_t \sum_{s=1}^{t-1}\frac{(t-s+\gamma-3)(t-s+\gamma-2)\langle \theta_{t-s} - \theta^*, {\psi}_{t-s+1}^{(b)}  \rangle}{(t+\gamma)(t+\gamma-1)} \nonumber \\ &- 2\eta_t \sum_{s=1}^{t-1}\frac{(t-s+\gamma-3)(t-s+\gamma-2)\langle \theta_{t-s} - \theta^*, {\psi}_{t-s+1}^{(v)}  \rangle}{(t+\gamma)(t+\gamma-1)} \nonumber . \label{eqn:proof_one_step_unrolled_with_bias_variance_mean_estimate}
\end{align}
Further simplifying by adding and subtracting $\mathbb{E}_{Z_t}[\| \psi_t^{(v)}\|_2^2 \vert \mathcal{F}_{t-1}]$ to be above display, we get
\begin{align}
                \|\theta^* - \widehat{\theta}_t\|_2^2 &\leq   \frac{(\gamma-1)(\gamma -2) \| \theta_1 - \theta^* \|_2^2}{(t+\gamma)(t+\gamma-1)}   + 4  \eta^2_t \sum_{s=1}^{t-1}\frac{(t-s+\gamma-3)(t-s+\gamma-2)\| {\psi}_{t-s+1}^{(b)} \|_2^2}{(t+\gamma)(t+\gamma-1)}  \\ &+  4  \eta^2_t \sum_{s=1}^{t-1}\frac{(t-s+\gamma-3)(t-s+\gamma-2)\mathbb{E}_{Z_{t-s+1}}[\| {\psi}_{t-s+1}^{(v)} \|_2^2\vert \mathcal{F}_{t-s}]}{(t+\gamma)(t+\gamma-1)}   \nonumber \\ & +  4 \eta^2_t \sum_{s=1}^{t-1}\frac{(t-s+\gamma-3)(t-s+\gamma-2) (\| {\psi}_{t-s+1}^{(v)} \|_2^2 - \mathbb{E}_{Z_{t-s+1}}[\| \psi_{t-s+1}^{(v)}\|_2^2 \vert \mathcal{F}_{t-s}])}{(t+\gamma)(t+\gamma-1)}    \\ & - 2\eta_t \sum_{s=1}^{t-1}\frac{ (t-s+\gamma-3)(t-s+\gamma-2)\langle \theta_{t-s} - \theta^*, {\psi}_{t-s+1}^{(b)} \rangle}{(t+\gamma)(t+\gamma-1)} \nonumber \\ & - 2\eta \sum_{s=1}^{t-1}\frac{ (t-s+\gamma-3)(t-s+\gamma-2)\langle \theta_{t-s} - \theta^*, {\psi}_{t-s+1}^{(v)}  \rangle}{(t+\gamma)(t+\gamma-1)}  . \label{eqn:proof_one_step_unrolled_with_bias_variance_martingale_mean_estimate}
\end{align}


\begin{lemma}[Lemma F.5 \citep{gorbunov2020stochastic}]
If $\lambda \geq 2G$, the following inequalities hold almost-surely for all times $t$.
\begin{align}
    \| \psi_t^{(v)}\| &\leq 2 \lambda \mathbf{1}_{\sigma > 0} \label{eqn:bound_on_variance_norm} \\
    \| \psi_t^{(b)} \|_2 &\leq \frac{4\sigma^2}{\lambda} \label{eqn:bound_on_bias_norm} \\
    \mathbb{E}_{Z_t}[ \| \psi_t^{(v)}\|_2^2 \vert \mathcal{F}_{t-1}] &\leq 10\sigma^2 \label{eqn:bound_expected_variance_norm}
\end{align}
\label{lem:bounds_on_norms}
\end{lemma}

Simplifying Equation (\ref{eqn:proof_one_step_unrolled_with_bias_variance_martingale_mean_estimate}) using bounds in Lemma \ref{lem:bounds_on_norms}, along with the fact that for all $1 \leq s \leq t$ and $\gamma \geq 1$, $\frac{(t-s+\gamma-3)(t-s+\gamma-2)}{(t+\gamma)(t+\gamma-1)} \leq \frac{t-s+\gamma}{t+\gamma}$ we get
\begin{align}
                \|\theta^* - \widehat{\theta}_t\|_2^2 &\leq    \frac{ (\gamma-1)(\gamma-2)\| \theta_1 - \theta^* \|_2^2}{(t+\gamma)(t+\gamma-1)}   + \frac{16 \eta^2_t \sigma^2}{\lambda} \sum_{s=1}^{t-1}\frac{t-s+\gamma}{t+\gamma} +  4 \eta^2_t \sigma^2 \sum_{s=1}^{t-1}\frac{t-s+\gamma}{t+\gamma}  \nonumber \\ &+  4 \eta^2_t \sum_{s=1}^{t-1}\frac{(t-s+\gamma)(\| {\psi}_{t-s+1}^{(v)} \|_2^2 - \mathbb{E}_{Z_{t-s+1}}[\| \psi_{t-s+1}^{(v)}\|_2^2 \vert \mathcal{F}_{t-s+1}])}{t+\gamma} \nonumber \\ &  + 2\eta_t \sum_{s=1}^{t-1}\frac{(t-s+\gamma) \| \theta_{t-s} - \theta^*\| \| {\psi}_{t-s+1}^{(b)} \|}{t+\gamma} + - 2\eta_t \sum_{s=1}^{t-1}\frac{(t-s+\gamma) \langle \theta_{t-s} - \theta^*, {\psi}_{t-s+1}^{(v)}  \rangle}{t+\gamma}  . \label{eqn:proof_one_step_unrolled_with_bias_variance_martingale_simplified_1_mean_estimate}
\end{align}
Further applying the bound that $\| \psi_t^{(b)}\| \leq \frac{4 \sigma^2}{{\lambda}}$
\begin{align}
                \|\theta^* - \widehat{\theta}_t\|_2^2 &\leq    \frac{ (\gamma-1) (\gamma-2)\| \theta_1 - \theta^* \|_2^2}{(t+\gamma)(t+\gamma-1)}   + \underbrace{\left(\frac{16 \eta^2_t \sigma^2}{\lambda} +  4 \eta^2_t \sigma^2 \right)\sum_{s=1}^{t-1}\frac{t-s+1}{t+\gamma}}_{\text{Term }1}  \nonumber \\ &+  \underbrace{4 \eta^2_t \sum_{s=1}^{t-1}\frac{(t-s+\gamma)(\| {\psi}_{t-s+1}^{(v)} \|_2^2 - \mathbb{E}_{Z_{t-s+1}}[\| \psi_{t-s+1}^{(v)}\|_2^2 \vert \mathcal{F}_{t-s+1}])}{t+\gamma}}_{\text{Term }2}   \nonumber \\ & + \underbrace{\frac{8\sigma^2 \eta_t}{\lambda}  \sum_{s=1}^{t-1}\frac{(t-s+\gamma) \| \theta_{t-s} - \theta^*\| }{t+\gamma}}_{\text{Term }3}  \underbrace{- 2\eta_t \sum_{s=1}^{t-1}\frac{(t-s+\gamma) \langle \theta_{t-s} - \theta^*, {\psi}_{t-s+1}^{(v)}  \rangle}{t+\gamma}}_{\text{Term }4} . \label{eqn:proof_one_step_unrolled_with_bias_variance_martingale_simplified_2_mid_mean_estimate}
\end{align}

\subsection{Probabilistic analysis}

\textbf{Definitions}
\\

For every $t \geq 1$, denote by the constant 
\begin{align}
    C_t  = \max\left(\frac{1024\sigma^4}{G^2m^2\lambda^2}, \frac{8 \lambda \sqrt{\ln \left( \frac{2t^3}{\delta} \right)}}{\gamma^2 G} \right).
    \label{eqn:defn_C}
\end{align}

Denote by the deterministic constant $\xi_u^{(t)}$ for $u= 1,\cdots, t$ as
\begin{align}
    \left(\xi_u^{(t)}\right)^2 := C_t\bigg[ \left(\frac{16  \sigma^2}{\lambda} +  4  \sigma^2 \right) \frac{1}{2m^2(u+1)} + \frac{96 \lambda^2 \ln \left( \frac{2t^3}{\delta}\right)\sigma(\sigma+1) }{m (u+\gamma)\sqrt{u+1}}  \bigg].
    \label{eqn:xi_t_defn}
\end{align}

From the definition, the following in-equalities hold. 
\begin{proposition}
For all times $u \in \{1,\cdots, t\}$,
\begin{align}
    \sum_{s=1}^{u-1}(u-s+\gamma)\xi_s^{(t)} &\leq 2(u+\gamma)\sqrt{u+1} \xi_u^{(t)}, \\
    \sum_{s=1}^{u-1}(\xi_s^{(t)})^2 &\leq 2(u+1)\ln(u+1) (\xi_u^{(t)})^2
    \label{eqn:xi_t_sum}
\end{align}
\label{prop:sum_of_xi_squared}
\end{proposition}
\begin{proof}
This follows from the following fact. 
\begin{proposition}
For all $ u \in \mathbb{N}$ and $\gamma \geq 0$, we have 
\begin{align*}
    \sum_{s=1}^{u-1}\frac{u-s+\gamma}{\sqrt{u+1}} \leq 2(u+\gamma)\sqrt{u+1}.
\end{align*}
\end{proposition}
\end{proof}

For each time $u \in \{1,\cdots, t\}$, denote by the random variable $\nu_u^{(t)}$ by 
\begin{align*}
    \nu_u^{(t)} := \twopartdef { \theta_{u} - \theta^* } {\| \theta_u - \theta^*\|^2 \leq (\xi_u^{(t)})^2 + \frac{C_t\gamma^2 G^2}{(u+1)}} {0} {\text{otherwise}}
\end{align*}

For every $u \in \{1,\cdots,t\}$, denote by the event $\mathcal{E}^{(t)}_{u;1}$ to be the one in which the following inequality holds for all $u \in \{1,\cdots, t\}$. 
\begin{multline}
 \mathcal{E}^{(t)}_{u;1} \coloneqq \bigg \{   4 \eta^2_t \sum_{s=1}^{u-1}\frac{(u-s+\gamma)(\| {\psi}_{u-s+1}^{(v)} \|_2^2 - \mathbb{E}_{Z_{u-s+1}}[\| \psi_{u-s+1}^{(v)}\|_2^2 \vert \mathcal{F}_{u-s+1}])}{t+\gamma} \\\leq \frac{96 \lambda^2 \ln \left( \frac{2t^2(t+1)}{\delta}\right)\sigma(\sigma+1) }{m (u+\gamma)\sqrt{u+1}} \bigg\}. \label{eqn:mean_martingale_term_1}
\end{multline}
and $\mathcal{E}^{(t)}_{u;2}$ as 
\begin{align}
   \mathcal{E}^{(t)}_{u;2} \coloneqq \bigg\{ - 2\eta_u \sum_{s=1}^{u-1}\frac{(u-s+\gamma) \langle \upsilon_{u-s}, {\psi}_{u-s+1}^{(v)}  \rangle}{t+\gamma} &\leq\frac{\xi_u^{(t)}\ln \left( \frac{2t^2(t+1)}{\delta}\right)}{10 \sqrt{u+1}} +  \frac{C_u \gamma^2 G^2}{4(u+1)}. \bigg\} \label{eqn:mean_martingale_term_2}
\end{align}

Denote by the event $\mathcal{E}^{(t)}$ as 
\begin{align}
    \mathcal{E}^{(t)} \coloneqq \bigcap_{u=1}^t \left(\mathcal{E}^{(t)}_{u;1} \cap \mathcal{E}^{(t)}_{u;2} \right).
    \label{eqn:event_e_t}
\end{align}

\begin{lemma}
For all $t\geq 1$,
\begin{align*}
    \mathbb{P}[\mathcal{E}^{(t)}] \geq 1 - \frac{\delta}{t(t+1)}.
\end{align*}

\label{lem:martingale_mean_estimate}
\end{lemma}

We now prove by induction hypothesis that 

\begin{lemma}
For every $t$, under the event $\mathcal{E}^{(t)}$,  the following holds.
\begin{align}
    \| \widehat{\theta}_u - \theta^*\|_2^2 \leq \frac{C_t\gamma^2 G^2}{(u+1)^2} + (\xi_u^{(t)})^2  ,
    \label{eqn:induction_hypothesis}
\end{align}
for all $u \in \{1,\cdots, t\}$.
\end{lemma}
\label{lem:induction}
\begin{proof}

\begin{proof}[Proof of Lemma \ref{lem:induction}]
We will prove this lemma by induction on $u$ by analyzing Equation (\ref{eqn:proof_one_step_unrolled_with_bias_variance_martingale_simplified_2_mid_mean_estimate}). The base-case of $u=1$ holds trivially with probability $1$ since $C_t > 1$, $\forall t \geq 1$ and $\gamma > 2$.
\\

Now, assume that on the event $\mathcal{E}^{(t)}$, the induction hypothesis in Equation (\ref{eqn:induction_hypothesis}) holds for all times $1,\cdots, u-1$. We prove this by expanding Equation (\ref{eqn:proof_one_step_unrolled_with_bias_variance_martingale_simplified_2_mid_mean_estimate}) and bounding each of the terms. 
\\

\textbf{Term 1} 
\\

It is easy to verify that 
\begin{align*}
    \left(\frac{16 \eta^2_u \sigma^2}{\lambda} +  4 \eta^2_u \sigma^2 \right)\sum_{s=1}^{u-1}\frac{u-s+\gamma}{u+\gamma} &\leq \left(\frac{16  \sigma^2}{\lambda} +  4  \sigma^2 \right)\frac{u}{2m^2(u+\gamma)^2}, \\
    &\leq \frac{\left( \frac{16\sigma^2}{\lambda} + 4 \sigma^2\right)}{2m^2(u+1)}.
\end{align*}
The last inequality follows since $\gamma^2 > 1$.

\textbf{Term 2}
\\

First notice that 
\begin{multline*}
    4 \eta^2_u \sum_{s=1}^{u-1}\frac{(u-s+\gamma)(\| {\psi}_{u-s+1}^{(v)} \|_2^2 - \mathbb{E}_{Z_{u-s+1}}[\| \psi_{u-s+1}^{(v)}\|_2^2 \vert \mathcal{F}_{u-s+1}])}{t+\gamma} \leq \\ \frac{4 \eta_u}{u+\gamma} \sum_{s=1}^{u-1}{(\| {\psi}_{u-s+1}^{(v)} \|_2^2 - \mathbb{E}_{Z_{u-s+1}}[\| \psi_{u-s+1}^{(v)}\|_2^2 \vert \mathcal{F}_{u-s+1}])}
\end{multline*}

From the definition of event $\mathcal{E}^{(t)}$ in Equation (\ref{eqn:event_e_t}), we get that 
\begin{align*}
    \text{Term }2 \leq \frac{96 \lambda^2 \ln \left( \frac{2t^2(t+1)}{\delta}\right)\sigma(\sigma+1) }{m (u+\gamma)\sqrt{u+1}}.
\end{align*}

\textbf{Term 3}
\\

\begin{align*}
    \frac{8\sigma^2 \eta_u}{\lambda}  \sum_{s=1}^{u-1}\frac{(u-s+\gamma) \| \theta_{u-s} - \theta^*\| }{u+\gamma} &\leq \frac{8 \sigma^2 }{m\lambda(u+\gamma)^2} \sum_{s=1}^{u-1}\left((u-s+\gamma){\xi_{u-s}^{(t)}} + \sqrt{C_t}\gamma G\frac{(u-s+\gamma)}{(u-s+1)}\right), \\
    &\stackrel{(\ref{eqn:xi_t_sum})}{\leq} \frac{16 \sigma^2 \sqrt{(u+1)}\xi_u^{(t)}}{m(u+\gamma)} + \frac{8 \sqrt{C_t}\sigma^2\gamma^2 G u }{m\lambda(u+\gamma)^2}, \\
     &\stackrel{}{\leq} \frac{16 \sigma^2 \sqrt{(u+1)}\xi_u^{(t)}}{m(u+\gamma)} + \frac{8 \sqrt{C_t} \sigma^2\gamma^2 G  }{m\lambda(u+\gamma)}, \\
    &\stackrel{(a)}{\leq} \frac{\xi_u^{(t)}}{10\sqrt{u+1}} +  \frac{{\color{black}C_t \gamma^2 G^2} }{4(u+1)}.
\end{align*}
The last inequality follows since $\gamma \geq \frac{320\sigma^2}{m}+1 \implies \frac{8 \sigma^2 (u+1)^{1/2}\log(u+1)}{m(u+\gamma)} \leq \frac{1}{10\sqrt{u+1}}$, for all $u \leq t$ and the fact that  {\color{black}$C_t \geq \frac{1024\sigma^4}{G^2m^2\lambda^2}$}.
\\ 

\textbf{Term 4}
\\

The definition of event $\mathcal{E}^{(t)}$ in Equation (\ref{eqn:event_e_t}) gives that $\text{Term }4 \leq \frac{\xi_u^{(t)}\ln \left( \frac{2t^2(t+1)}{\delta}\right)}{10\sqrt{u+1}} + \frac{C_t \gamma^2 G^2}{4(u+1)}$
\\






{\color{black}

Now, adding in the bounds together into Equation (\ref{eqn:proof_one_step_unrolled_with_bias_variance_martingale_simplified_2_mid_mean_estimate}), 
\begin{multline*}
    \| \widehat{\theta}_u - \theta^{*} \|_2^2 \leq \frac{\gamma^2G^2}{u+1} + \frac{\left( \frac{16\sigma^2}{\lambda} + 4 \sigma^2\right)}{2m^2(u+1)} + \frac{\xi_u^{(t)}}{10\sqrt{u+1}} + \frac{1600 \lambda^2 \ln \left( \frac{2t^2(t+1)}{\delta}\right)\sigma(\sigma+1) }{m (u+\gamma)\sqrt{u+1}} \\ + \frac{\xi_u^{(t)}\ln \left( \frac{2t^2(t+1)}{\delta}\right)}{10\sqrt{u+1}} + \frac{C_t \gamma^2 G^2}{2(u+1)}.
\end{multline*}
Now using the fact that $\frac{\xi_u^{(t)}\ln \left( \frac{2 t^3}{\delta}\right)}{\sqrt{u+1}} \leq (\xi_u^{(t)})^2$, we get that 
\begin{align*}
    \| \widehat{\theta}_u - \theta^{*} \|_2^2 &\leq \left(1+\frac{C_t}{2} \right)\frac{\gamma^2G^2}{u+1} + \frac{\left( \frac{16\sigma^2}{\lambda} + 4 \sigma^2\right)}{2m^2(u+1)} + \frac{(\xi_u^{(t)})^2}{5} + \frac{96 \lambda^2 \ln \left( \frac{2t^2(t+1)}{\delta}\right)\sigma(\sigma+1) }{m (u+\gamma)\sqrt{u+1}}.
\end{align*}
Substituting the definition of $\xi_u^{(t)}$ from Equation (\ref{eqn:xi_t_defn}), we get that 
\begin{align*}
     \| \widehat{\theta}_u - \theta^{*} \|_2^2 &\leq \left(1 + \frac{C_t}{2} \right)\left[\frac{\gamma^2G^2}{u+1} + \frac{\left( \frac{16\sigma^2}{\lambda} + 4 \sigma^2\right)}{2m^2(u+1)} + \frac{96 \lambda^2 \ln \left( \frac{2t^2(t+1)}{\delta}\right)\sigma(\sigma+1) }{m (u+\gamma)\sqrt{u+1}}   \right] , \\
    &\leq (\xi_u^{(t)})^2 + \frac{C_t \gamma^2 G^2}{u+1}.
\end{align*}
The last inequality follows since $C_t  = \max\left(\frac{1024\sigma^4}{G^2m^2\lambda^2}, \frac{8 \lambda \sqrt{\ln \left( \frac{2t^3}{\delta} \right)}}{\gamma^2 G} \right) \implies C_t \geq 2$. 
}

\end{proof}

\end{proof}


\subsection{Proof of Lemma \ref{lem:martingale_mean_estimate}}

We first reproduce an useful result.

\begin{lemma}[Freedman’s inequality\citep{victor1999general}]
Suppose $Y_1, \cdots, Y_T$ is a bounded martingale with respect to a filtration $(\mathcal{F}_t)_{t=0}^T$ with $\mathbb{E}[Y_t \vert \mathcal{F}_{t-1}] = 0$ and $\mathbb{P}[|Y_t| \leq B] = 1$ for all $t \in \{1,\cdots, T\}$. Denote by $V_s := \sum_{n=1}^s \text{Var}(Y_n \vert \mathcal{F}_{n-1})$ be the sum of conditional variances. Then, for every $a, v > 0$, 
\begin{align}
    \mathbb{P} \left( \exists n \in [1,T] \text{ such that } \sum_{t=1}^n Y_t \geq a \text{ and } V_n \leq v \right) \leq \exp \left( \frac{-a^2}{2(v+ Ba)} \right).
    \label{eqn:martingale_diff_thm_bound}
\end{align}
\label{lem:martingale_diff}
\end{lemma}
Re-arranging the above inequality, we see that if 
\begin{align}
    a \geq B \ln \left( \frac{2 T}{\delta} \right) + \sqrt{ \left(B \ln \left( \frac{2 T}{\delta} \right)\right)^2 + 2v \ln \left( \frac{2 T}{\delta} \right)  },
    \label{eqn:martingale_useful_bound}
\end{align}
then the RHS of Equation (\ref{eqn:martingale_diff_thm_bound}) is bounded above by $\frac{\delta}{2}$. 

\begin{proof}[Proof of Lemma \ref{lem:martingale_mean_estimate}]
\textbf{Proof of Equation (\ref{eqn:mean_martingale_term_1})}
\\

Fix a $u \in \{1,\cdots, t\}$. For $s \in \{1,\cdots, u-1\}$, denote by the random variable $Y_s^{(u)} :=  \frac{(u-s+\gamma)}{u+\gamma}(\| {\psi}_{u-s+1}^{(v)} \|_2^2 - \mathbb{E}_{Z_{u-s+1}}[\| \psi_{u-s+1}^{(v)}\|_2^2 \vert \mathcal{F}_{u-s}])$. Thus, 
\begin{align*}
    4 \eta^2_u \sum_{s=1}^{u-1}\frac{(u-s+\gamma)(\| {\psi}_{u-s+1}^{(v)} \|_2^2 - \mathbb{E}_{Z_{u-s+1}}[\| \psi_{u-s+1}^{(v)}\|_2^2 \vert \mathcal{F}_{u-s+1}])}{u+\gamma} \leq 4\eta_u^2 {\sum_{s=1}^{u-1}Y_s^{(u)}}.
\end{align*}
Observe that the sequence $(Y_s^{(u)})_{s=1}^{u-1}$ is a martingale difference sequence with respect to the filtration $(\mathcal{G}_s)_{s=1}^{t-1}$, where $\mathcal{G}_s := \mathcal{F}_{u-s}$. Furthermore, observe that with probability $1$, $| Y_s^{(u)}| \leq 4\lambda^2\mathbf{1}_{\sigma > 0} + 4\lambda^2\mathbf{1}_{\sigma > 0} \leq 8\lambda^2\mathbf{1}_{\sigma > 0}$. We can bound the conditional variance as 
\begin{align*}
    \sum_{s=1}^{u-1}\text{Var}(Y_s^{(u)} \vert \mathcal{G}_s) &\leq  \sum_{s=1}^{u-1}\left(\frac{(u-s+\gamma)}{u+\gamma}\right)^2\mathbb{E}_{Z_{u-s}}[ (\| {\psi}_{u-s+1}^{(v)} \|_2^2 - \mathbb{E}_{Z_{u-s+1}}[\| \psi_{u-s+1}^{(v)}\|_2^2 \vert \mathcal{F}_{u-s}])^2 \vert \mathcal{F}_{u-s} ], \\
    &\stackrel{\ref{eqn:bound_on_variance_norm}}{\leq}  8\lambda^2 \sum_{s=1}^{u-1}\mathbb{E}_{Z_{u-s}}[ |\| {\psi}_{u-s+1}^{(v)} \|_2^2 - \mathbb{E}_{Z_{u-s+1}}[\| \psi_{u-s+1}^{(v)}\|_2^2 \vert \mathcal{F}_{u-s}]| \vert \mathcal{F}_{u-s} ], \\
    &\leq 8  \lambda^2 \sum_{s=1}^{u-1} 2 \mathbb{E}_{Z_{u-s}}[ |\| {\psi}_{u-s+1}^{(v)} \|_2^2 \vert \mathcal{F}_{u-s}], \\
    &\stackrel{\ref{eqn:bound_expected_variance_norm}}{\leq} 160 \lambda^2 \sigma^2 (u-1).
\end{align*}
Now, putting $B := 8  \lambda^2$ and $v = 160  \lambda^2 \sigma^2 u$, we get from Equation (\ref{eqn:martingale_useful_bound}) that with probability at-least $1-\delta/(2t^2(t+1))$, 
\begin{align*}
    \sum_{s=1}^{u-1}Y_s^{(u)} &\leq 8  \lambda^2  \ln \left( \frac{2t^2(t+1)}{\delta} \right)\mathbf{1}_{\sigma > 0} + \sqrt{\left(8  \lambda^2  \ln \left( \frac{2t^2(t+1)}{\delta} \right)\mathbf{1}_{\sigma > 0} \right)^2 + 160 \lambda^2 \sigma^2 u \ln \left( \frac{2t^2(t+1)}{\delta} \right)}, \\
    &\stackrel{(a)}{\leq} 32  \lambda^2 \ln \left( \frac{2t^2(t+1)}{\delta} \right)\sigma(\sigma+1)\sqrt{u+1}.
    \end{align*}
Step $(a)$ follows from the fact that $\lambda \geq 1$. Thus, we have with probability at-least $1-\frac{\delta}{2t^2(t+1)}$, 
\begin{align*}
  4 \eta^2_u \sum_{s=1}^{u-1}\frac{(u-s+\gamma)(\| {\psi}_{u-s+1}^{(v)} \|_2^2 - \mathbb{E}_{Z_{u-s+1}}[\| \psi_{u-s+1}^{(v)}\|_2^2 \vert \mathcal{F}_{u-s+1}])}{u+\gamma} &\leq 96\eta_u^2   \lambda^2 \ln \left( \frac{2t^2(t+1)}{\delta} \right)\sigma(\sigma+1)\sqrt{u+1}, \\
    &\leq  \frac{96\lambda^2 \ln \left( \frac{2t^2(t+1)}{\delta} \right)\sigma(\sigma+1)\sqrt{u+1}}{m^2(u+\gamma)^2}, \\
    &\leq \frac{96\lambda^2 \ln \left( \frac{2t^2(t+1)}{\delta} \right)\sigma(\sigma+1)}{m^2(u+\gamma)\sqrt{u+1}}.
\end{align*}
Now taking an union bound over all $u \in \{1,\cdots,t\}$ yields that with probability at-least $1- \frac{\delta}{2t(t+1)}$, for all time $u \in \{1, \cdots, t\}$, 
\begin{align*}
     4 \eta^2_u \sum_{s=1}^{u-1}\frac{(t-s+\gamma)(\| {\psi}_{u-s+1}^{(v)} \|_2^2 - \mathbb{E}_{Z_{u-s+1}}[\| \psi_{u-s+1}^{(v)}\|_2^2 \vert \mathcal{F}_{u-s+1}])}{t+\gamma} &\leq \frac{96 \lambda^2 \ln \left( \frac{2t^2(t+1)}{\delta}\right)\sigma(\sigma+1) }{m (u+\gamma)\sqrt{u+1}}
\end{align*}

\textbf{Proof of Equation (\ref{eqn:mean_martingale_term_2})}
\\

{\color{black}
\begin{align*}
    - 2\eta_u \sum_{s=1}^{u-1}\frac{(u-s+\gamma) \langle \upsilon_{u-s}, {\psi}_{u-s+1}^{(v)}  \rangle}{u+\gamma} \leq \frac{2}{m(u+\gamma)^2} \sum_{s=1}^{u-1}{ \langle \theta_{u-s} - \theta^*, {\psi}_{u-s+1}^{(v)}  \rangle}
\end{align*}

Fix a $u \in \{1, \cdots, t\}$ and denote by $Y_s^{(u)} := (u-s+\gamma) \langle \theta_{u-s} - \theta^{*}, \psi_{u-s+1}^{(v)} \rangle$. Since $\theta_{u-s}$ is measurable with respect to the sigma-algebra generated by $\mathcal{F}_{u-s}$, the conditional expectation $\mathbb{E}[Y_s^{(u)} \vert \mathcal{F}_{u-s}] = 0$. Thus, $(Y_s^{(u)})_{s=1}^{u-1}$ is a martingale difference sequence with respect to the filtration $(\mathcal{F}_{u-s})_{s=1}^{u-1}$. Furthermore, we have from Equation (\ref{eqn:bound_on_variance_norm}) that $|Y_s^{(u)}| \leq 2 (u-s+\gamma) \left(\xi_{u-s}^{(t)} + \frac{\gamma R_1}{(u+\gamma - 1)} \right)\lambda  \leq 2\lambda(u+\gamma)\xi_t^{(t)} + 2\lambda \gamma G$. We can now bound the sum of conditional variances as 
\begin{align*}
    \sum_{s=1}^{u-1} \text{Var}(Y_s^{(u)} \vert \mathcal{F}_{u-s}) &\leq \sum_{s=1}^{u-1} 4 (u-s+\gamma)^2 (\xi_{u-s}^{(t)})^2 \lambda^2\sigma^2 + 4\lambda^2 G^2, \\
    &\stackrel{(\ref{eqn:xi_t_sum})}{\leq} 12 \lambda^2 \sigma^2 (u+\gamma)^2(u+1) \log(u+1) (\xi_u^{(t)})^2 + 4\lambda^2\gamma^2G^2 u . 
\end{align*}
Step $(a)$ follows since $\eta m <1$. Now applying the bound in Equation (\ref{eqn:martingale_useful_bound}) with $B := 2\lambda(u+\gamma)\xi_u^{(t)} + 2\lambda G$ and $v = 12 \lambda^2 \sigma^2 (u+\gamma)^2(u+1) \log(u+1) (\xi_u^{(t)})^2 + 4 \lambda^2  \gamma^2 G^2u$, we get that with probability at-least $1-\delta/(2t^2(t+1))$, 
\begin{align*}
   \sum_{s=1}^{u-1}&{(u-s+\gamma) \langle \upsilon_{u-s}, {\psi}_{u-s+1}^{(v)}  \rangle}\leq 2 \lambda \left((u+\gamma)\xi_u^{(t)} + R_1\right) \ln \left( \frac{2 t^2(t+1)}{\delta} \right) +  \bigg[\left(2 \lambda \left((u+\gamma)\xi_u^{(t)} + G\right)  \ln \left( \frac{2 t^2(t+1)}{\delta} \right) \right)^2 \\& +  \left( \lambda^2 \sigma^2 (u+\gamma)^2(u+1) \log(u+1) (\xi_u^{(t)})^2 + 4\lambda^2\gamma^2G^2 (u+1) \right) \ln \left( \frac{2t^2(t+1)}{\delta}\right)\bigg]^{\frac{1}{2}}, \\
    &\stackrel{}{\leq} {6(u+\gamma)\sqrt{u+1} \log(u+1) (\xi_u^{(t)}) \lambda \sigma(\sigma+1) \ln \left( \frac{2 t^2(t+1)}{\delta}\right)} + 2\lambda \gamma G \sqrt{(u+1) \ln \left( \frac{2t^2(t+1)}{\delta} \right)}.
\end{align*}
Thus,

\begin{align*}
   - 2\eta_u  \sum_{s=1}^{u-1}\frac{(u-s+\gamma) \langle \upsilon_{u-s}, {\psi}_{u-s+1}^{(v)}  \rangle}{u+\gamma}  &\leq \frac{12\sqrt{u+1} \log(u+1) (\xi_u^{(t)}) \lambda \sigma(\sigma+1) \ln \left( \frac{2t^2(t+1)}{\delta}\right)}{(u+\gamma)} + \frac{C_t \gamma G}{10(u+1)}, \\
    &\leq \frac{\xi_u^{(t)}\ln \left( \frac{2t^2(t+1)}{\delta}\right)}{10 \sqrt{u+1}} + \frac{C_t G}{10(u+1)}.
\end{align*}
The first inequality follows since $C_t \geq \frac{8 \lambda \sqrt{\ln \left( \frac{2t^3}{\delta} \right)}}{\gamma^2 G}$. The last inequality follows since for all times $u \leq t$, we have 
\begin{align*}
    \frac{12{\sqrt{u+1}} \log(u+1) \lambda \sigma(\sigma+1) \ln \left( \frac{2t^2(t+1)}{\delta}\right)}{(u+\gamma)} \leq \frac{\ln \left( \frac{2t^2(t+1)}{\delta}\right)}{10}
\end{align*}
as a consequence of $\gamma \geq 120 \lambda \sigma(\sigma+1)$.
 }

\end{proof}

\section{Proofs from Section \ref{sec:fpr_guarantee}}

\subsection{Proof of Theorem \ref{thm:fpr_main}}
\label{sec:proof_thm_fpr_main}

We bound this probability using the result of \ref{thm:main_mean_est} and a simple union bound argument. For any process $\mathfrak{M}$, observe that 
\begin{align}
    \mathbb{P}[\exists t \in [r+1,\tau_c^{(r)}) \text{ s.t.} \mathcal{A}_{t} =1 \vert \mathcal{A}_r = 1] &= \mathbb{P}[\cup_{t=r+1}^{\tau_c-1} \mathcal{A}_{t} = 1 \vert \mathcal{A}_r = 1] \nonumber \\
    &\leq \sum_{t=r+1}^{\tau_c-1} \mathbb{P}[\mathcal{A}_{t} = 1 \vert \mathcal{A}_r = 1]. \label{eqn:fpr_proof_mid_point_1}
\end{align}

We now examine the above Equation to bound it. For any fixed $t \in (r, \tau_c^{(r)})$
\begin{align}
   \mathbb{P}[ & \mathcal{A}_{t} = 1 \vert \mathcal{A}_r = 1] 
    =  \mathbb{P}\left[ \bigcup_{s=r+1}^{t-1} \| \widehat{\theta}_{r:s} - \widehat{\theta}_{s+1:t} \| \geq \mathcal{B}\left(s-r, \frac{\delta}{2t(t+1)}\right) + \mathcal{B}\left(t-s-1, \frac{\delta}{2t(t+1)}\right)\right], \nonumber\\
    &\leq \sum_{s=r+1}^{t-1} \left(\mathbb{P}\left[ \| \widehat{\theta}_{r:s} - \theta_{c-1} \| \geq \mathcal{B}\left(s-r, \frac{\delta}{2t(t+1)}\right) \right] +  \mathbb{P}\left[ \| \widehat{\theta}_{s+1:t} - \theta_{c-1} \| \geq \mathcal{B}\left(t-s-1, \frac{\delta}{2t(t+1)}\right) \right]\right), \nonumber\\
    &\stackrel{(a)}{\leq} \sum_{s=r+1}^{t-1} \left( \frac{\delta}{2t(t+1)(s-r)(s-r+1)} + \frac{\delta}{2t(t+1)(t-s-1)(t-s)} \right), \nonumber\\
    &= \frac{\delta}{2t(t+1)} \left( \sum_{s=r+1}^{t-1} \frac{1}{(s-r)(s-r+1)} + \sum_{s=r+1}^{t-1} \frac{1}{(t-s-1)(t-s)}\right), \nonumber\\
    &\leq \frac{\delta}{2t(t+1)} \left( \sum_{s=1}^{t-1-r} \frac{1}{s(s+1)} + \sum_{s=1}^{t-1-r}\frac{1}{s(s+1)} \right), \nonumber\\
    &\stackrel{(b)}{\leq} \frac{\delta}{t(t+1)}. \label{eqn:fpr_inter_2}
\end{align}

Since for all $t < \tau_c^{(r)}$, the mean of the random variables $X_{r+1}, \cdots, X_t$ are identical and equal to $\theta_{c-1}$ (see notation in Section \ref{sec:problem_formulation}), Theorem \ref{thm:main_mean_est} gives rise to inequality $(a)$. Step $(b)$ follows from the fact that $\sum_{s \geq 1} \frac{1}{s(s+1)} = 1$. Now substituting the bound from Equation (\ref{eqn:fpr_inter_2}) into Equation (\ref{eqn:fpr_proof_mid_point_1}), we get that 
\begin{align*}
      \mathbb{P}[\exists t \in [r+1,\tau_c^{(r)}) \text{ s.t. } \mathcal{A}_{t} =1 \vert \mathcal{A}_r = 1] &\leq \sum_{t=r+1}^{\tau_c-1} \frac{\delta}{t(t+1)}, \\
      &\leq \sum_{t \geq 1} \frac{\delta}{t(t+1)}, \\
      &=\delta. 
\end{align*}

Since the above bound holds for all $r$ and process $\mathfrak{M}$, we have 
\begin{align*}
    \sup_{\mathfrak{M}, r}\mathbb{P}[\exists t \in [r+1,\tau_c^{(r)}) \text{ s.t.} \mathcal{A}_{t}  =1 \vert \mathcal{A}_r = 1] \leq \delta. 
\end{align*}

\subsection{Proof of Lemma \ref{lem:fpr_connection}}
\label{sec:proof_of_fpr_connection}

Recall from the definition that the $r$th detection is false if
\begin{align*}
    \chi_r^{(A)} = \mathbf{1}(\not\exists c \text{ s.t. } \tau_c \in (t_{r-1}^{(A)}, t_r^{(A)}]).
\end{align*}

We will show that $\mathbb{E}[\chi_r^{(A)} ] \leq \delta$. This will then conclude the proof of the lemma. 

\begin{align*}
    \mathbb{E}[\chi_r^{(A)} ] &= \mathbb{P}[\not\exists c \text{ s.t. } \tau_c \in (t_{r-1}^{(A)}, t_r^{(A)}]], \\
    &= \mathbb{E} \left[ \mathbb{P}[\not\exists c \text{ s.t. } \tau_c^{(s)} \in (s, t_r^{(A)}]]\bigg| t_{r-1}^{(A)} = s \right], \\
    &\leq \mathbb{E} \left[ \mathbb{P}[\cup_{t=s+1}^{\infty} \tau_c^{(s)} = t, t_{r}^{(\mathcal{A})} < t]\bigg \vert t_{r-1}^{(A)} = s  \right], \\
     &\leq \mathbb{E} \left[ \mathbb{P}[\exists t \in [s+1, \tau_c^{(s)}), \mathcal{A}_{t} = 1 ]\bigg\vert t_{r-1}^{(A)} = s  \right], \\
      &\stackrel{(a)}{\leq} \mathbb{E} \left[ \mathbb{P}[\exists t \in [s+1, \tau_c^{(s)}), \mathcal{A}_{t} = 1 \vert \mathcal{A}_{s} = 1  ]\bigg\vert t_{r-1}^{(A)} = s  \right], \\
     &\stackrel{(b)}{\leq} \delta.
\end{align*}
Inequality $(a)$ follows from the fact that on the event $t_{r-1}^{(\mathcal{A})} = s$, $\mathcal{A}_s = 1$. Inequality $(b)$ follows from Theorem \ref{thm:fpr_main}.
\section{Proof of Lemma \ref{lem:detection_delay}}
\label{sec:proof_delay}



The proof follows from a straightforward application of Theorem \ref{thm:main_mean_est} as follows. Let $n \in \mathbb{N}, \Delta > 0$ and $\delta' \in (0,1)$ be arbitrary. 

\begin{align}
    \mathbb{P}[ \mathcal{D}(n, \Delta, \delta') \geq d ] &=  \mathbb{P}[ \cap_{s=1}^{n+d} \mathcal{A}(X_{1:s}) = 0 ], \nonumber \\
    &=\mathbb{P}\left[ \bigcap_{s=1}^{n+d} \| \widehat{\theta}_{1:s} - \widehat{\theta}_{s+1:n+d} \|_2^2 \leq \mathcal{B}\left(s,\frac{\delta}{2(n+d)(n+d+1)}\right) + \mathcal{B}\left(n+d-s-1, \frac{\delta}{2(n+d)(n+d+1)}\right) \right], \nonumber\\
    &\leq \mathbb{P}\left[  \| \widehat{\theta}_{1:n-1} - \widehat{\theta}_{n:n+d} \|_2^2 \leq \mathcal{B}\left(n-1,\frac{\delta}{2(n+d)(n+d+1)}\right) + \mathcal{B}\left(d, \frac{\delta}{2(n+d)(n+d+1)}\right) \right].
    \label{eqn:delay_proof_1}
\end{align}

From triangle-inequality, we know that 
\begin{align}
    \| \widehat{\theta}_{1:n-1} - \widehat{\theta}_{n:n+d} \|_2^2 &\geq \| \theta_1 - \theta_2 \|_2^2 - \| \widehat{\theta}_{1:n-1} - \theta_1 \|_2^2 - \|  \widehat{\theta}_{n:n+d} - \theta_2 \|_2^2, \nonumber\\
    &= \Delta^2 - \| \widehat{\theta}_{1:n-1} - \theta_1 \|_2^2 - \|  \widehat{\theta}_{n:n+d} - \theta_2 \|_2^2. 
    \label{eqn:delay_proof_2}
\end{align}

Thus, substituting Equation (\ref{eqn:delay_proof_2} into Equation (\ref{eqn:delay_proof_1}), we get that 
\begin{multline*}
  \mathbb{P}[ \mathcal{D}(n, \Delta, \delta') \geq d ] \leq     \mathbb{P}\bigg[ \Delta^2 - \| \widehat{\theta}_{1:n-1} - \theta_1 \|_2^2 - \|  \widehat{\theta}_{n:n+d} - \theta_2 \|_2^2 \leq \\ \mathcal{B}\left(n-1,\frac{\delta}{2(n+d)(n+d+1)}\right) + \mathcal{B}\left(d, \frac{\delta}{2(n+d)(n+d+1)}\right) \bigg].
\end{multline*}

Denote by the events $\mathcal{E}_i$ for $i \in \{1,2\}$ as
\begin{align*}
    \mathcal{E}_1 &:= \left\{ \| \widehat{\theta}_{1:n-1} - \theta_1 \|_2^2 > \mathcal{B}\left(n-1, \frac{\delta'}{2}\right) \right\}, \\
    \mathcal{E}_2 &:= \left\{ \| \widehat{\theta}_{n:n+d} - \theta_2 \|_2^2 > \mathcal{B}\left(d, \frac{\delta'}{2}\right) \right\}, \\
\end{align*}
Denote by $\mathcal{E} := \mathcal{E}_1 \cup \mathcal{E}_2$. Theorem \ref{thm:main_mean_est} gives that $\mathbb{P}[\mathcal{E}_1] \leq \frac{\delta'}{2(n(n+1))} \leq \frac{\delta'}{2}$ and $\mathbb{P}[\mathcal{E}_2] \leq \frac{\delta'}{2d(d+1)} \leq \frac{\delta'}{2}$. Thus, an union bound gives that $\mathbb{P}[\mathcal{E}] \leq \delta'$. Let $d' \in \mathcal{G}$ be arbitrary, where
\begin{align}
   \mathcal{G} := \bigg\{ d \in \mathbb{N} : \Delta^2 \geq \mathcal{B}\left( n-1, \frac{\delta'}{2} \right) + \mathcal{B}\left( d, \frac{\delta'}{2} \right)  + \mathcal{B}\left( n, \frac{\delta}{2(n+d+1)(n+d)} \right) +  \mathcal{B}\left( d,  \frac{\delta}{2(n+d+1)(n+d)} \right) \bigg\}
   \label{eqn:delay_proof_3}
\end{align}

\textbf{Claim} : If the event $\mathcal{E}^{c}$ holds, then $\mathcal{D}(n,\Delta, \delta) \leq d$ for all $d \in \mathcal{G}$. 

 Suppose $d \in \mathcal{G}$ and event $\mathcal{E}^c$ holds. Then, we know by triangle inequality in Equation (\ref{eqn:delay_proof_2}) that 

\begin{align}
    \| \widehat{\theta}_{1:n-1} - \widehat{\theta}_{n:n+d} \|_2^2 &\geq \| \theta_1 - \theta_2 \|_2^2 - \| \widehat{\theta}_{1:n-1} - \theta_1 \|_2^2 - \|  \widehat{\theta}_{n:n+d} - \theta_2 \|_2^2, \nonumber\\
    &= \Delta^2 - \| \widehat{\theta}_{1:n-1} - \theta_1 \|_2^2 - \|  \widehat{\theta}_{n:n+d} - \theta_2 \|_2^2, \\
    &\stackrel{(a)}{\geq} \Delta^2 -  \mathcal{B}\left(n-1, \frac{\delta'}{2}\right) -  \mathcal{B}\left(d, \frac{\delta'}{2}\right), \\
    &\stackrel{(b)}{\geq} \mathcal{B}\left( n, \frac{\delta}{2(n+d+1)(n+d)} \right) +  \mathcal{B}\left( d,  \frac{\delta}{2(n+d+1)(n+d)} \right).
    \label{eqn:delay_proof_4}
\end{align}
Step $(a)$ follows from the definition of event $\mathcal{E}$ and on the assumption of the claim that event $\mathcal{E}^c$ holds. Step $(b)$ follows from the fact that $d \in \mathcal{G}$ is arbitrary (cf. Equation (\ref{eqn:delay_proof_3}). The last step says from Line $8$ of Algorithm \ref{algo:learn_model} that if no detection has been made till time $n+d$, then under the event $\mathcal{E}^c$, time step $d$ is a detection time. Since event $\mathcal{E}^c$ holds with probability at-least $1-\delta'$ , this concludes the proof.

\begin{figure}
    \centering
    \includegraphics[width=0.4\linewidth]{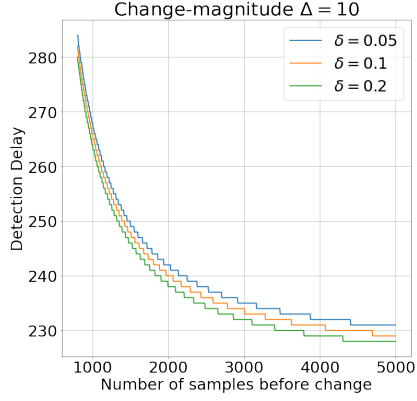}
    \caption{Plot of $ \mathcal{D}(n, \Delta, \delta')$ in Lemma \ref{lem:detection_delay} for fixed $\Delta=10, \delta=0.1$.}
    \label{fig:detection_delay}
\end{figure}


 \subsection{Useful convexity based inequalities}
 
Let $f : \Theta \to \mathbb{R}$ be a strongly convex function with strong convexity parameters $0 < m \leq M < \infty$. Denote by $\theta^* := \arg\min_{\theta \in \Theta}f(\theta)$. Since $f(\cdot)$ is convex and $\Theta$ is convex and compact, the existence and uniqueness of $\theta^*$ is guaranteed. Strong convexity gives that for any $\widehat{\theta}_{t-1} \in \Theta$,
\begin{align}
   f({\theta}^*) \geq f(\widehat{\theta}_{t-1}) + \langle \nabla f(\widehat{\theta}_{t-1}),  \theta^*-\widehat{\theta}_{t-1}  \rangle + \frac{m}{2} \|  \theta^* -\widehat{\theta}_{t-1} \|_2^2.
\end{align}
Further since $\theta^* = \arg\min_{\theta \in \Theta}f(\theta)$., we have that 
\begin{align*}
 f(\widehat{\theta}_{t-1}) - f(\theta^*_t) \geq \frac{m}{2}\| \widehat{\theta}_{t-1} - \theta^*\|_2^2.
\end{align*}
Putting these two together, we see that 
\begin{align}
    \langle \nabla f(\widehat{\theta}_{t-1}), \widehat{\theta}_{t-1} - \theta^* \rangle \geq m \|\widehat{\theta}_{t-1} - \theta^*\|_2^2.
    \label{eqn:convexity_gradient_bound}
\end{align}
Also,  We further use the following lemma. 
\begin{lemma}[Lemma $3.11$ from \citep{bubeck2015convex}]
Let $g : \mathbb{R}^d \to \mathbb{R}$ be a $M$ smooth and $m$ strongly convex function. Then for all $x,y \in \mathbb{R}^d$,
\begin{align*}
    \langle \nabla g(x) - \nabla g(y), x-y \rangle \geq \frac{mM}{M+m} \| x-y\|_2^2 + \frac{1}{M + m}\| \nabla g(x) - \nabla g(y) \|_2^2.
\end{align*}
\end{lemma}
By substituting $x = \widehat{\theta}_{t-1}$, $y = \theta^*_{t}$ and $g(\cdot) = f(\cdot)$ and by leveraging the fact that $\nabla f(\theta^*) = 0$, we get the inequality that 
\begin{align*}
    \langle \nabla f(\widehat{\theta}_{t-1}), \widehat{\theta}_{t-1} - \theta^* \rangle \geq \frac{mM}{m+M}\|\widehat{\theta}_{t-1} - \theta^* \|_2^2 + \frac{1}{M+m}\|\nabla f(\widehat{\theta}_{t-1}) \|_2^2. 
\end{align*}
Re-arranging, we see that 
\begin{align}
    \|\nabla f(\widehat{\theta}_{t-1}) \|_2^2 \leq (M+m) \langle \nabla f(\widehat{\theta}_{t-1}), \widehat{\theta}_{t-1} - \theta^*\rangle - mM \| \widehat{\theta}_{t-1} - \theta^*\|_2^2.
        \label{eqn:convexity_bound_inner_prod}
\end{align}

\section{Additional Simulations}

\begin{figure*}[ht!]
\centering
\begin{subfigure}{0.32\linewidth}
\includegraphics[width=0.99\linewidth]{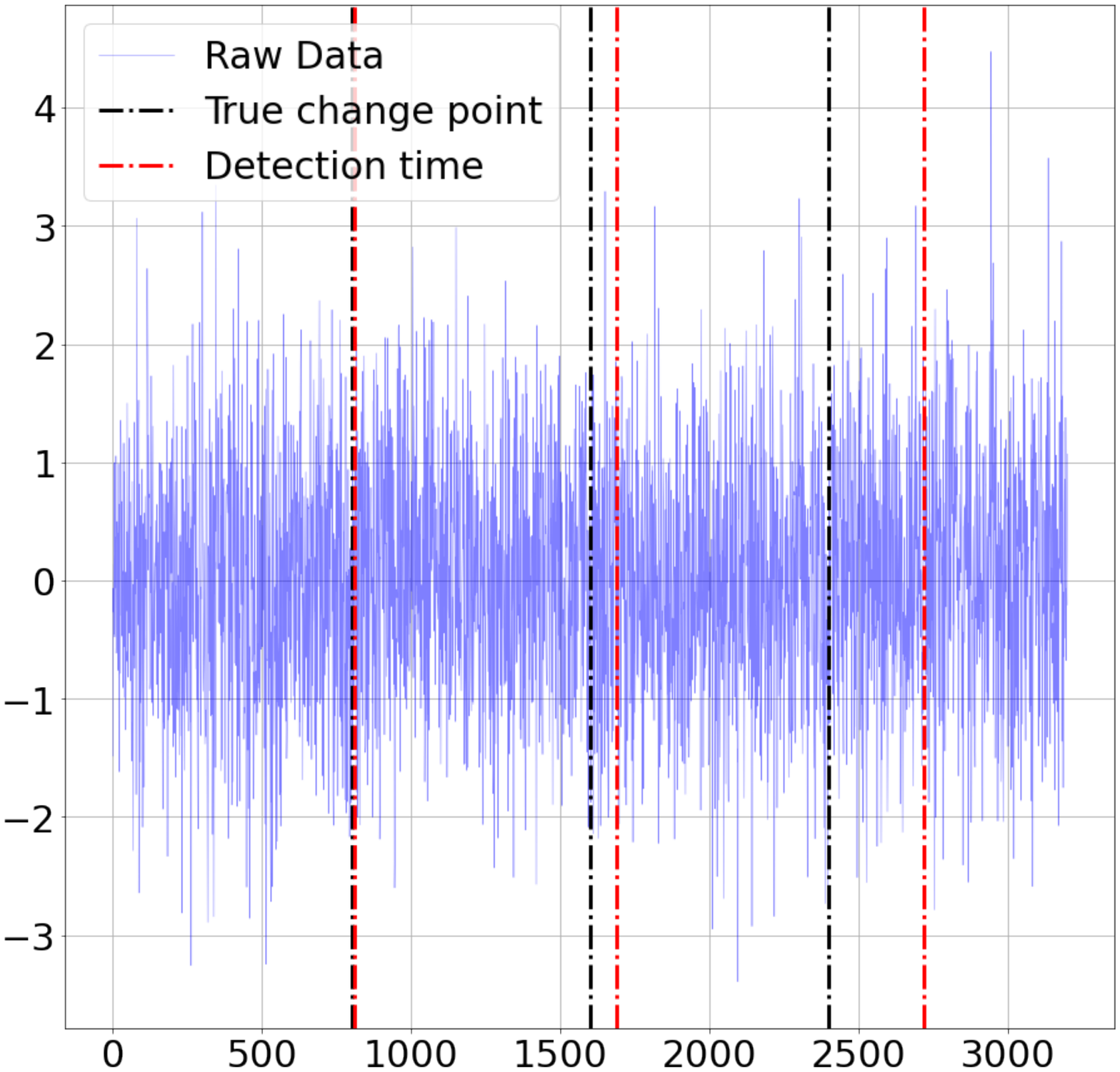}
\caption{Unit-variance Gaussian.}
\label{fig:fig1}
\end{subfigure}
\begin{subfigure}{0.32\linewidth}
\includegraphics[width=0.99\linewidth]{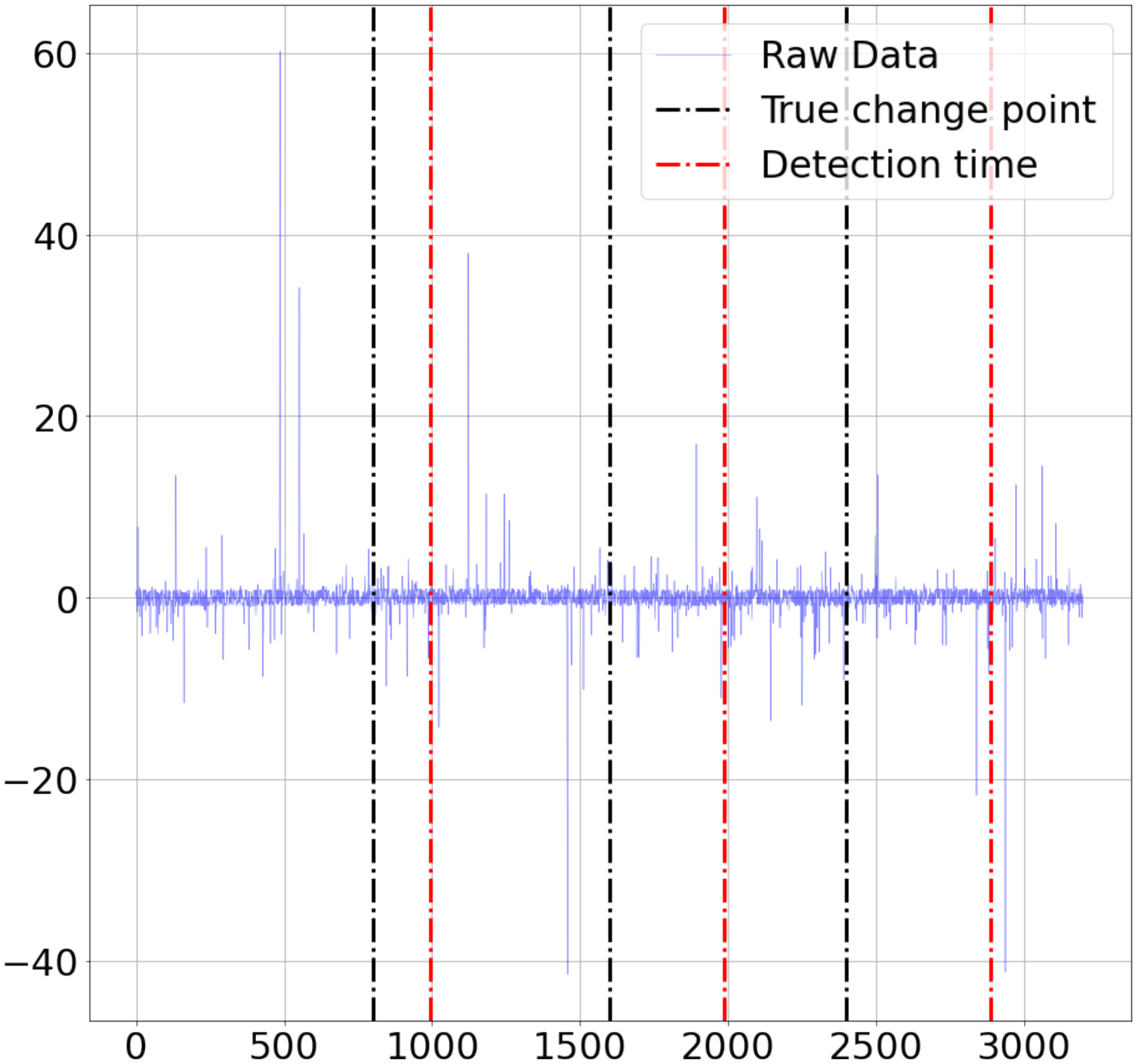}
\caption{Pareto with $s=2.1$.}
\label{fig:fig2}
\end{subfigure}
\begin{subfigure}{0.32\linewidth}
\includegraphics[width=0.99\linewidth]{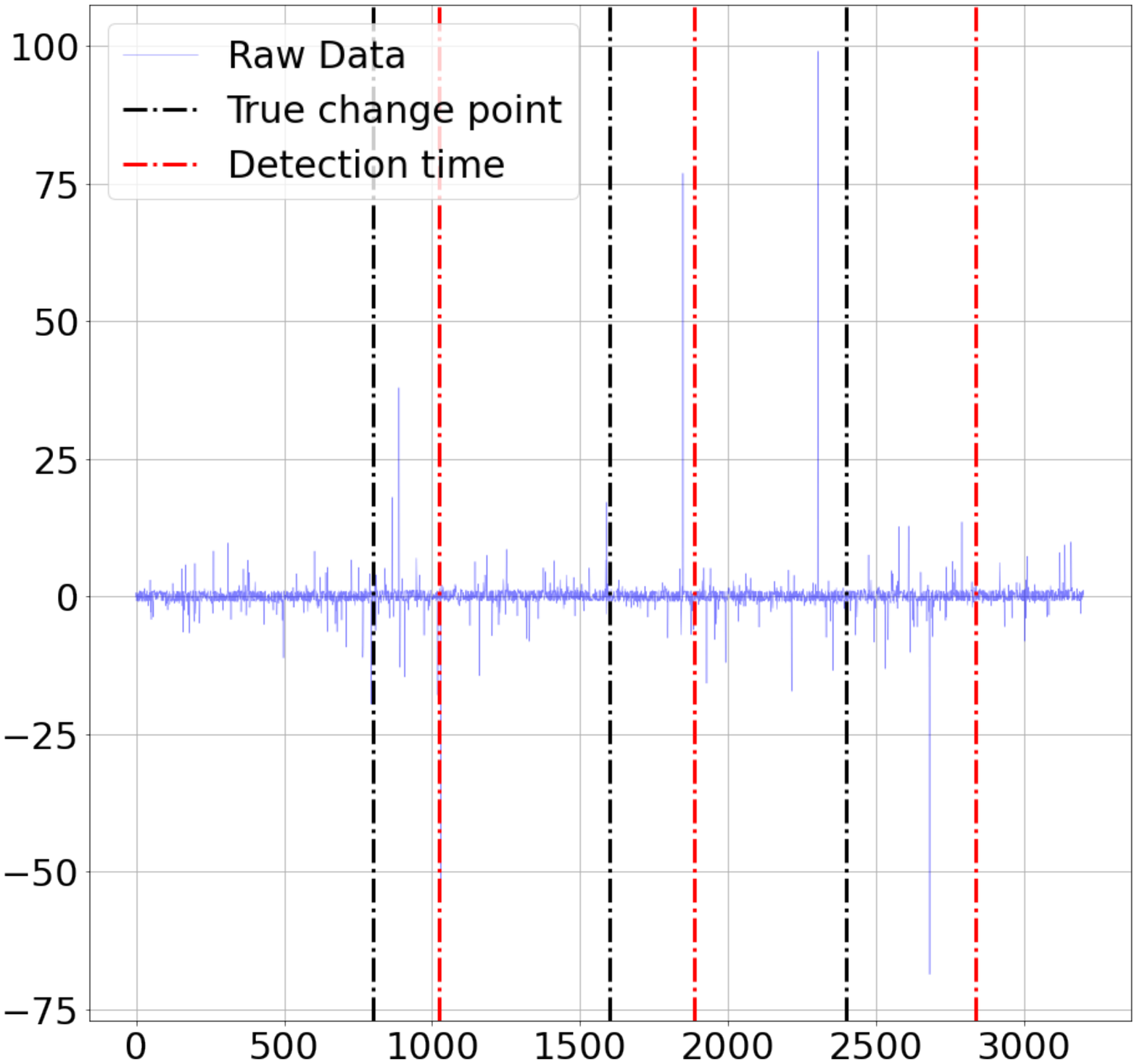}
\caption{Pareto with $s=2.01$.}
\label{fig:f3}
\end{subfigure}
\begin{subfigure}{0.32\linewidth}
\includegraphics[width=0.99\linewidth]{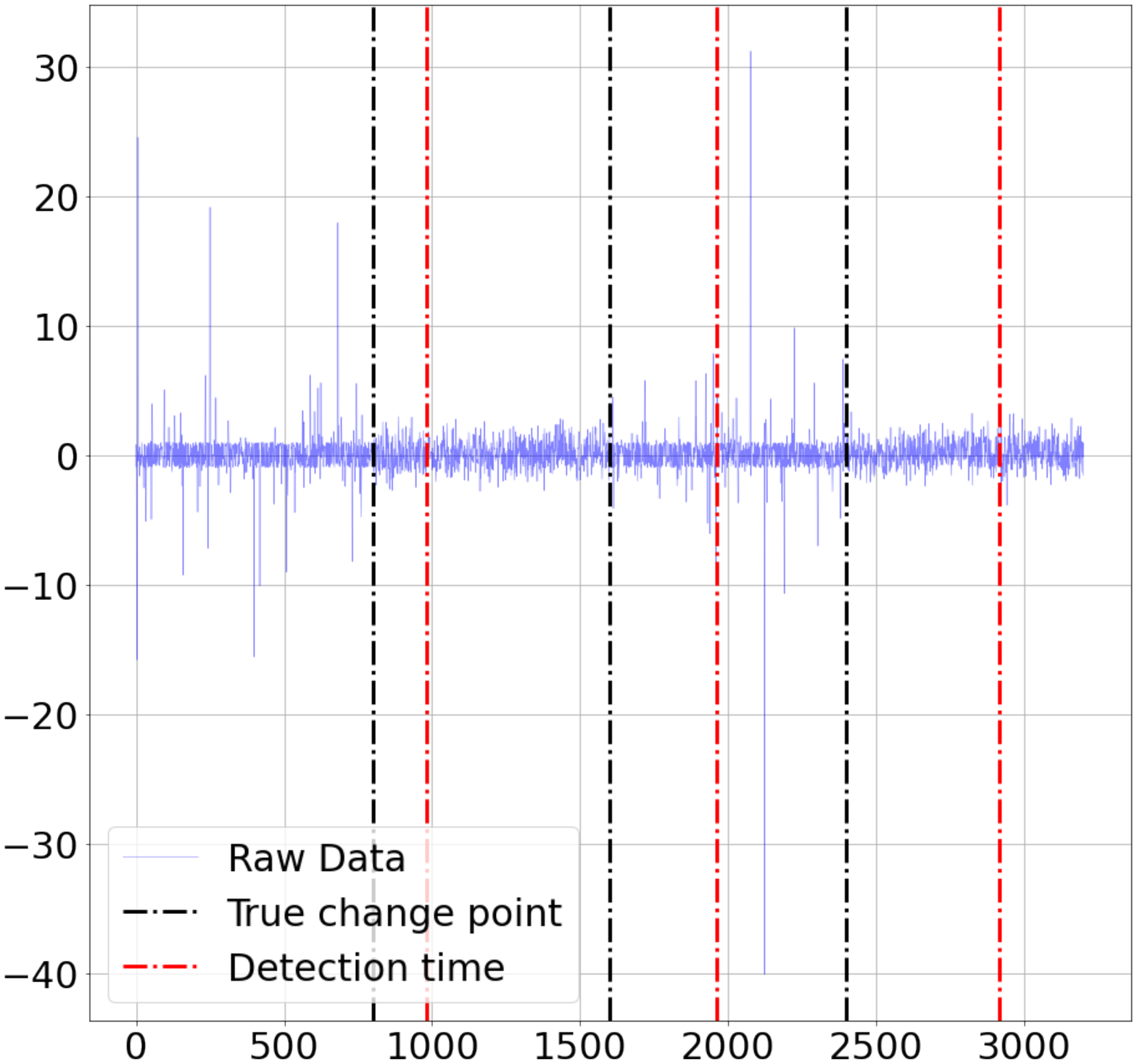}
\caption{Alternate Pareto $s=2.01$ and Gaussian.}
\label{fig:f4}
\end{subfigure}
\begin{subfigure}{0.32\linewidth}
\includegraphics[width=0.99\linewidth]{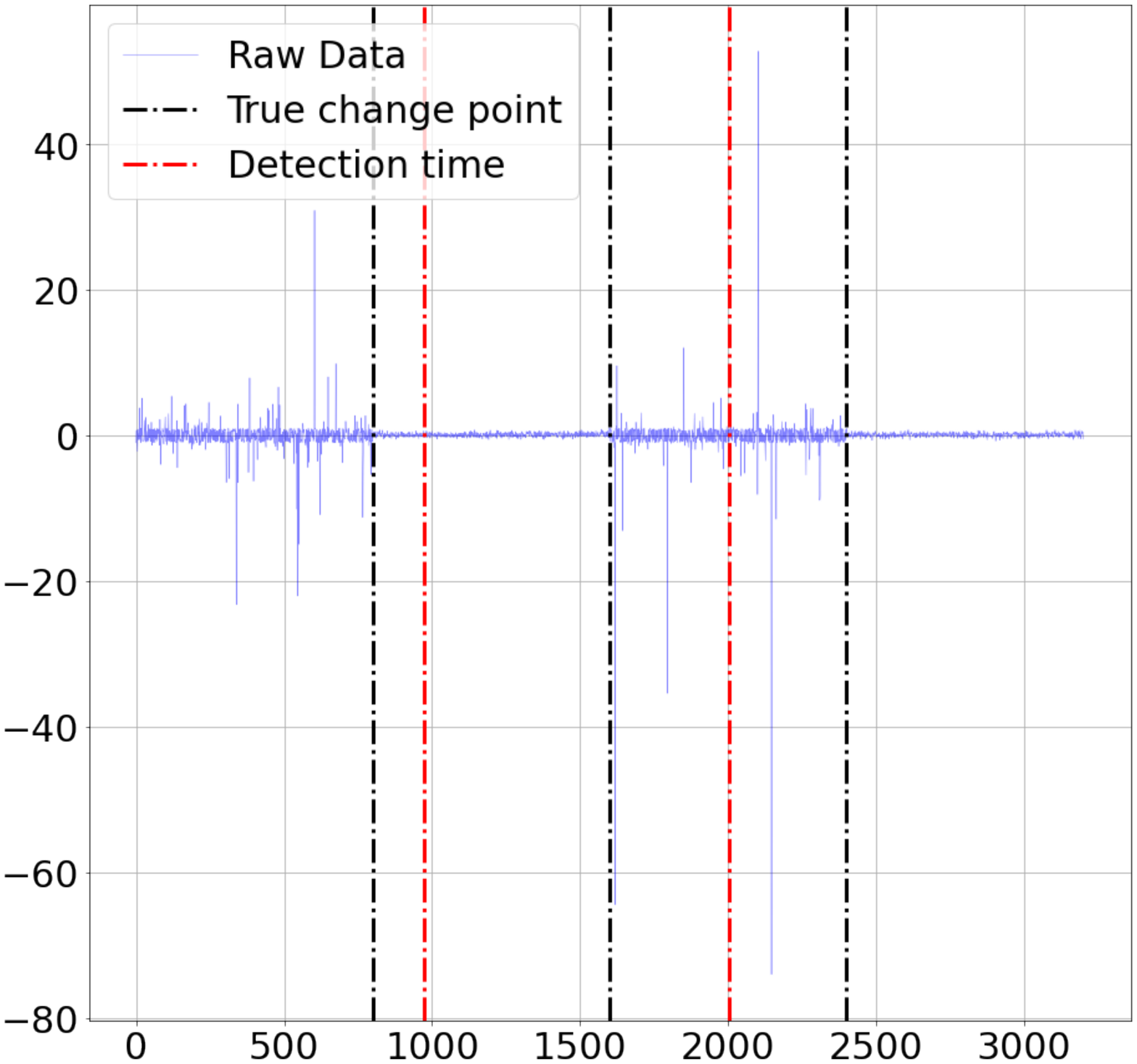}
\caption{Alternate Pareto $s=2.01$ and Gaussian}
\label{fig:f5}
\end{subfigure}
\begin{subfigure}{0.32\linewidth}
\includegraphics[width=0.99\linewidth]{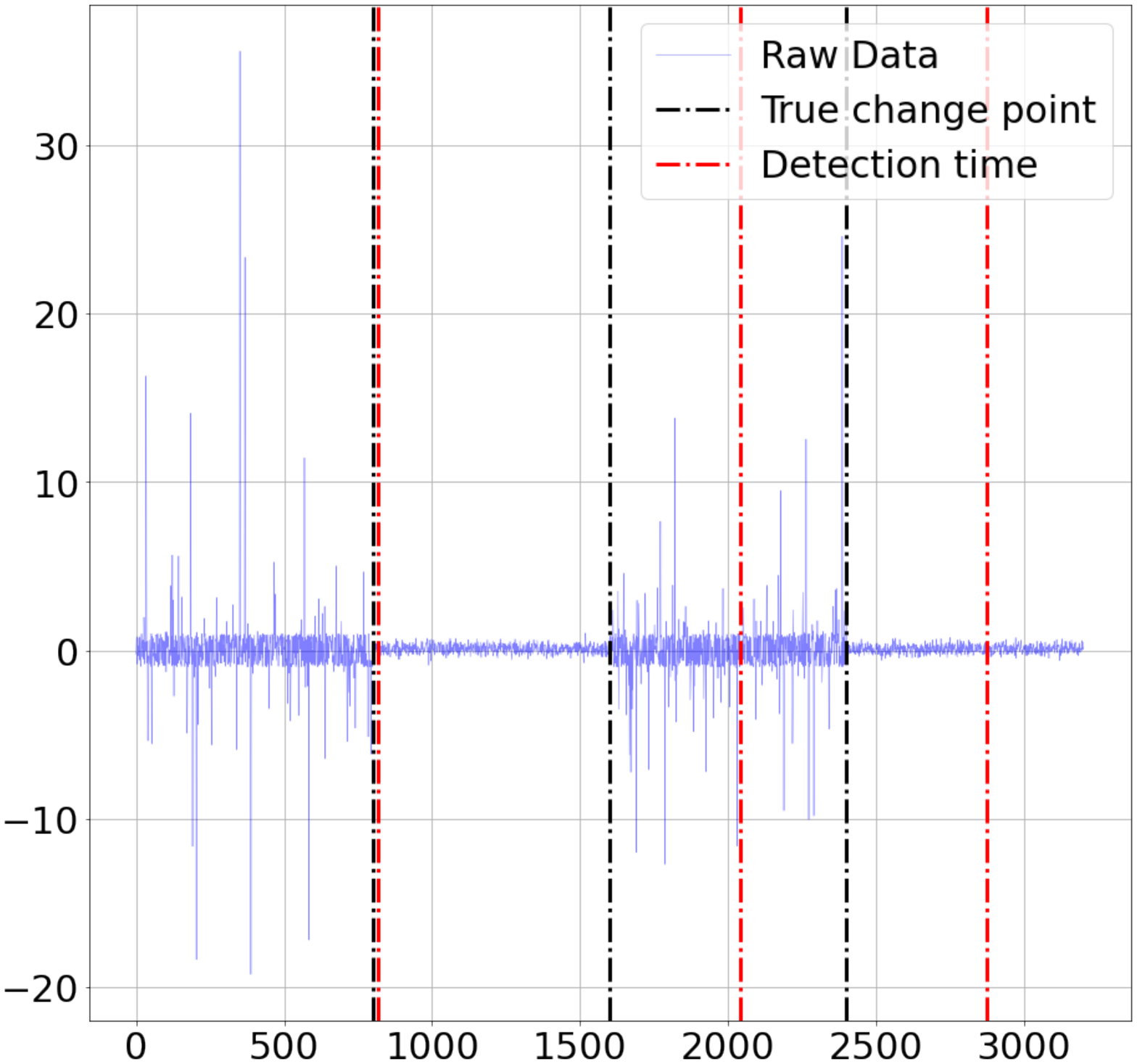}
\caption{Alternate Pareto $s=2.01$ and Gaussian}
\label{fig:f6}
\end{subfigure}
\begin{subfigure}{0.32\linewidth}
\includegraphics[width=0.99\linewidth]{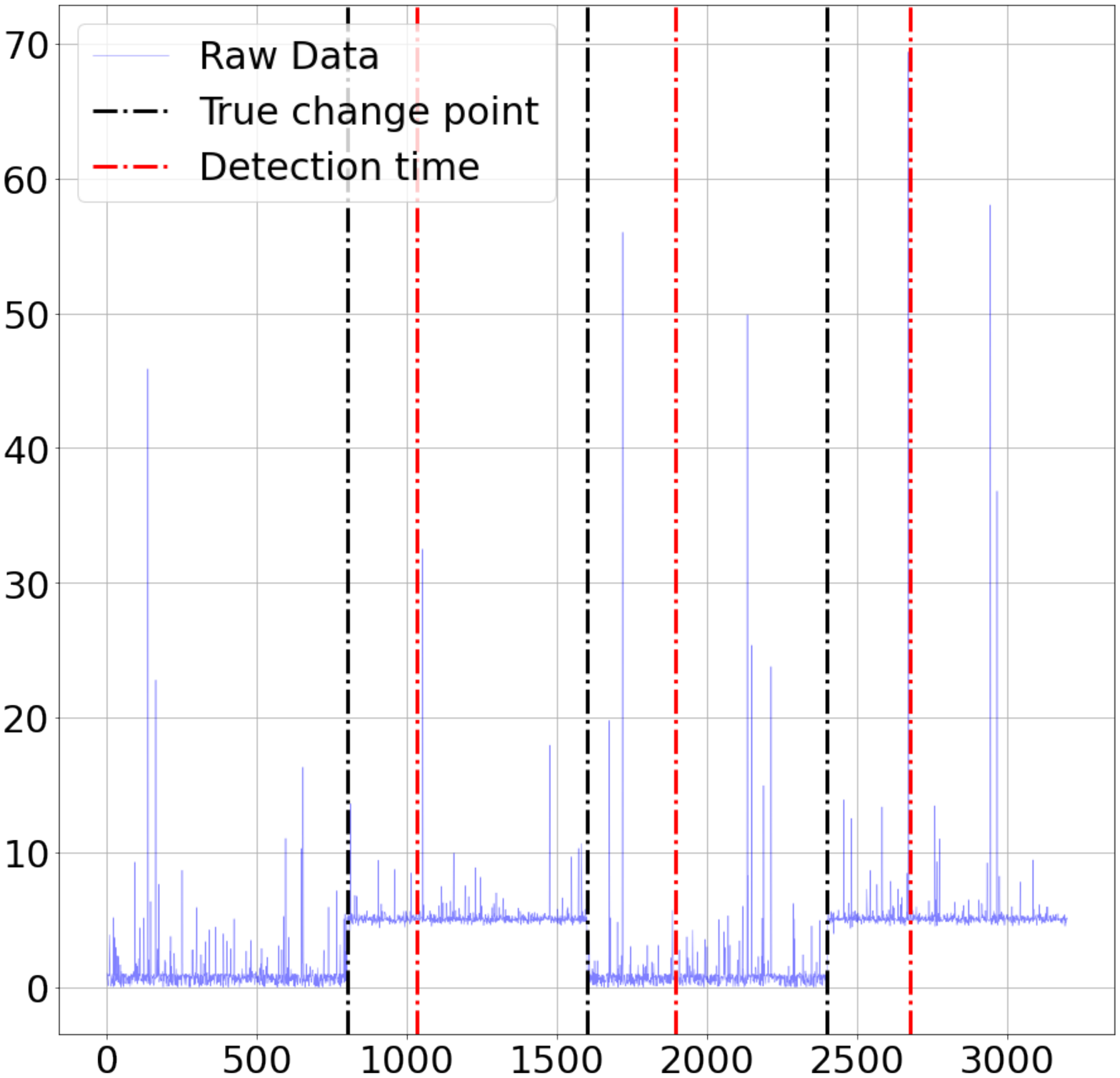}
\caption{Pareto $s=2.01, d=15,\Delta=5$}
\label{fig:f7}
\end{subfigure}
\begin{subfigure}{0.32\linewidth}
\includegraphics[width=0.99\linewidth]{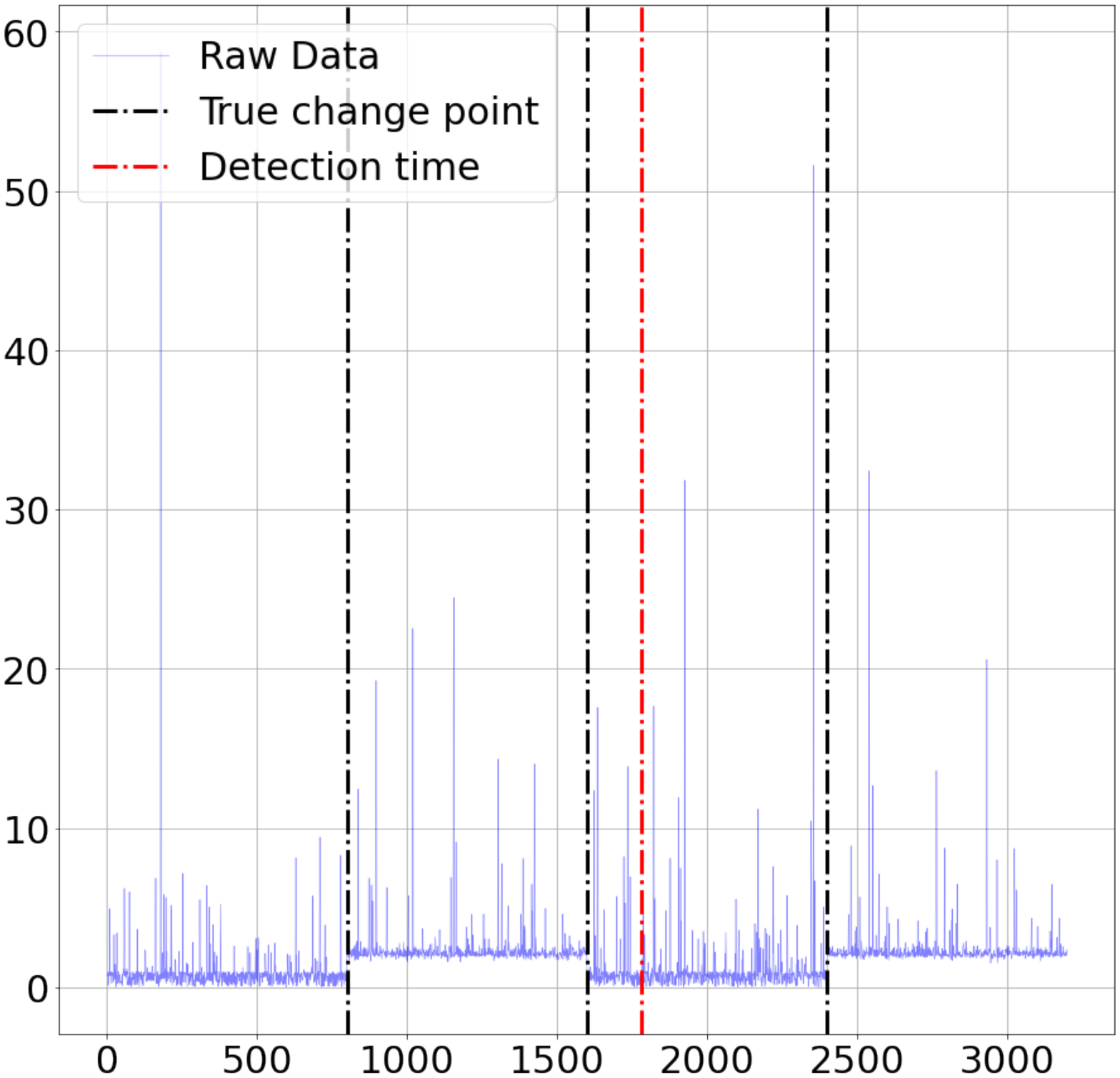}
\caption{Pareto $s=2.01, d=15,\Delta=2$}
\label{fig:f8}
\end{subfigure}
\caption{In all plots, we choose the change-point gap to be $\Delta=0.1$ and $\delta=0.05$ except (g) and (h) where $\Delta=5$ and $2$ respectively. In plots $(g)$ and $(h)$, we plot the norm of the observed random vector and thus the Y-axis is non-negative. We see missed detection in Figures $(e)$ and $(h)$ with the last change-point on the right being missed. We do not observe False-positives in these plots.  }
\label{fig:sample_path}
\end{figure*}

In Figure \ref{fig:sample_path}, we plot a sample path of observed data and mark out the true change-points and the detected time-instants by Algorithm \ref{algo:learn_model}. The plots indicate that although visually identifying the change in the means is hard, our change-point detection algorithm is able to consistently across variety of distribution families. 

\end{document}